\documentclass[letterpaper, 10 pt, journal, twoside]{IEEEtran} 

\IEEEoverridecommandlockouts      

\usepackage{amsmath} 
\usepackage{amssymb}  
\usepackage[nolist]{acronym}
\usepackage{graphicx}
\usepackage{xcolor}
\usepackage{booktabs}
\usepackage[percent]{overpic}
\usepackage{multirow}
\usepackage{newtxtext}
\newcommand{\revison}[1]{#1}

\newcommand{\GG}{\mathcal{G}}
\newcommand{\RR}[1]{\mathbb{R}^{#1}}
\newcommand{\dimG}{n}
\newcommand{\numX}{N}

\newcommand{\SO}[1]{\ensuremath{\mathrm{SO}\!\left(#1\right)}}
\newcommand{\SE}[1]{\ensuremath{\mathrm{SE}\!\left(#1\right)}}
\newcommand{\SEtwo}[1]{\ensuremath{\mathrm{SE}_2\!\left(#1\right)}} 
\newcommand{\SU}[1]{\ensuremath{\mathrm{SU}\!\left(#1\right)}}
\newcommand{\SL}[1]{\ensuremath{\mathrm{SL}\!\left(#1\right)}}
\newcommand{\Sim}[1]{\ensuremath{\mathrm{Sim}\!\left(#1\right)}}
\newcommand{\Gal}[1]{\ensuremath{\mathrm{Gal}\!\left(#1\right)}}

\newcommand{\numZ}{M}
\newcommand{\Zij}{\tilde X_{ij}}
\newcommand{\Hg}{H}      
\newcommand{\Hgk}{H_{i,j}}   
\newcommand{\Hd}{\bar{H}}   
\newcommand{\Hdk}{\bar{H}_{i,j}}  

\newcommand{\vecc}{\operatorname{vec}}

\newcommand{\Exp}{\mathrm{Exp}}
\newcommand{\Log}{\mathrm{Log}}

\newcommand{\X}{X}
\newcommand{\eij}{\epsilon_{ij}}
\newcommand{\FF}[1]{\big\| #1 \big\|_F}

\DeclareMathOperator*{\argmin}{arg\,min}

\newcommand{\fastsync}{\textit{Fast-Sync}}
\newcommand{\sesync}{\textit{SE-Sync}}
\newcommand{\mst}{\textit{MST}}

\newlength{\PanelW}
\newcommand{\PanelBox}[1]{%
  \begin{minipage}[t]{\PanelW}\vspace{0pt}\centering
    #1
  \end{minipage}%
}

\newcommand{\InsetTimeTable}[4]{%
  \setlength{\fboxrule}{0.3pt}%
  \setlength{\fboxsep}{2.5pt}%
  \fcolorbox{black!50}{white}{%
    \begingroup
    \setlength{\tabcolsep}{2.5pt}%
    \renewcommand{\arraystretch}{0.9}%
    \scriptsize
    \begin{tabular}{@{}lcc@{}}
      \textbf{Time} & \textbf{MST} & \textbf{Fast} \\[-1pt]
      \cmidrule(lr){2-3}
      \textbf{orig}  & #1 & #2 \\
      \textbf{noisy} & #3 & #4 \\
    \end{tabular}%
    \endgroup
  }%
}

\begin{document}

\setlength{\textfloatsep}{8pt plus 2pt minus 2pt}
\setlength{\dbltextfloatsep}{8pt plus 2pt minus 2pt}
\setlength{\intextsep}{6pt plus 2pt minus 2pt}

\title{
  FAST-Sync: Fast Group Synchronization for any Matrix Lie Group
}

 \author{Shane Holmes$^{2}$, Yiran Luo$^{1}$, Firat Taxpulat$^{1}$, David M. Rosen$^{2}$
 and Frank Dellaert$^{1}$
   \thanks{Manuscript received: March 5, 2026; Revised May 21, 2026; Accepted June 15, 2026.}%
   \thanks{This paper was recommended for publication by Editor Lucia Pallottino upon evaluation of the Associate Editor and Reviewers' comments.}%
   \thanks{$^{1}$Frank Dellaert, Yiran Luo, and Firat Taxpulat are with the School of
     Interactive Computing, Georgia Institute of Technology, Atlanta, GA, USA
   {\tt\footnotesize  {frank.dellaert@cc.gatech.edu, yluo432@gatech.edu, ftaxpulat3@gatech.edu}}}%
   \thanks{$^{2}$Shane Holmes and David M. Rosen are with Northeastern University, Boston, USA
   {\tt\footnotesize  holmes.sha@northeastern.edu, d.rosen@northeastern.edu}}%
   \thanks{Digital Object Identifier (DOI): 10.1109/LRA.2026.3710327.}
 }

\begin{acronym}
  \acro{SfM}{structure from motion}
  \acro{SLAM}{simultaneous localization and mapping}
  \acro{MLE}{maximum likelihood estimation}
  \acro{SDP}{semidefinite programming}
  \acro{RA}{rotation averaging}
  \acro{PGO}{pose graph optimization}
  \acro{MST}{minimum spanning tree}
  \acro{GS}{group synchronization}
  \acro{SOTA}{state of the art}
  \acro{ND}{Nested Dissection}
  \acro{DAG}{directed acyclic graph}
  \acro{AMD}{approximate minimum degree}
  \acro{WS}{Watts-Strogatz}
\end{acronym}

\markboth{IEEE Robotics and Automation Letters. Preprint Version. Accepted June, 2026}
{Holmes \MakeLowercase{\textit{et al.}}: FAST-Sync}

\makeatletter
\def\@IEEEpubidpullup{34pt}
\makeatother
\IEEEpubid{\parbox[b]{\textwidth}{\fontsize{8}{9}\selectfont
\copyright\ 2026 IEEE. Personal use of this material is permitted. Permission from IEEE must be obtained for all other uses, in any current or future media, including reprinting/republishing this material for advertising or promotional purposes, creating new collective works, for resale or redistribution to servers or lists, or reuse of any copyrighted component of this work in other works.}}

\IEEEaftertitletext{\vspace{-18pt}}
\maketitle

\begin{abstract}

  \emph{Group synchronization} (GS) is the problem of estimating a set of $N$ unknown
  elements $g_1,\dotsc, g_N \in \GG$ in a group $\GG$, given noisy measurements of a
  subset of their pairwise ratios $g_i^{-1} g_j$. {GS} problems lie at the core of many
  state estimation tasks in robotics and computer vision, including 3D vision, robotic
  mapping, inertial navigation, and molecular reconstruction. Unfortunately, {GS}
  problems are typically both high-dimensional and non-convex, and therefore hard to solve
  in general. In this paper, we present \fastsync{}, a fast linear approximation method for
  {GS} that is suitable for initializing local manifold-based optimizers or certifiable
  global methods. Our approach generalizes chordal
  initialization~\cite{Govindu01cvpr_MotionEstimation,Martinec07cvpr_robust} to arbitrary
  matrix Lie groups, and additionally proposes two new key algorithmic enhancements: we
  show how to exploit both the Kronecker-product structure in the problem data matrix and
  the topology of the synchronization graph to improve speed, scalability,
  and accuracy.  Experimental evaluation across several {GS} tasks demonstrates that
  \fastsync{} provides high-quality initializations that enable local optimizers to
  efficiently recover globally optimal GS solutions, achieving high success rates even
  with considerable measurement noise.

\end{abstract}



\section{INTRODUCTION}

\IEEEPARstart{P}{ose} graph optimization (PGO) is a mainstay of robotics, dating back to the seminal paper by Lu
and Milios~\cite{Lu97ar_GloballyConsistent}, and has become one of the dominant
techniques in \ac{SLAM}. Similarly, in computer vision, \ac{RA} dates back to the
pioneering work by
Govindu~\cite{Govindu01cvpr_MotionEstimation,Govindu04cvpr_LieAlgebraic}, and is a key step in
global 3D reconstruction pipelines.
While these problems are in general non-convex, certifiably optimal methods have recently
been proposed~\cite{Rosen19ijrr_SE_Sync, Dellaert20eccv_Shonan}.
Both are instances of the much broader \ac{GS}
problem~\cite{Arrigoni20ijcv_Synchronization}, in which we want to recover a set
of absolute group elements ($\SE{d}$ for \ac{PGO}, $\SO{d}$ for \ac{RA}) from a collection
of noisy \textit{relative} measurements between them.

At the same time, interest in different Lie groups has been growing, e.g., $\Sim{d}$ to
deal with scale drift in monocular \ac{SLAM}~\cite{Strasdat10rss_ScaleDrift},
$\SL{d}$ in 3D reconstruction~\cite{Maggio25arxiv_VGGT_SLAM}, the
\textit{extended pose group} $\SEtwo{3}$ in invariant
filtering~\cite{Barrau16tac_InvariantEKF}, and
the Galilean group to describe the relationship between moving reference
frames~\cite{Kelly23arxiv_Galilean}.
Translation averaging~\cite{Arrigoni20ijcv_Synchronization} is also of interest in \ac{SfM},
although as it is linear, the focus of that work is primarily in outlier
rejection~\cite{Wilson14eccv_1dsfm}.
Outside of robotics, \ac{GS} also arises in cryo-electron microscopy~\cite{Singer11siam_Cryo}.

\begin{figure}[!t]
  \centering
  \includegraphics[width=0.21\textwidth, trim=58 58 58 58, clip]{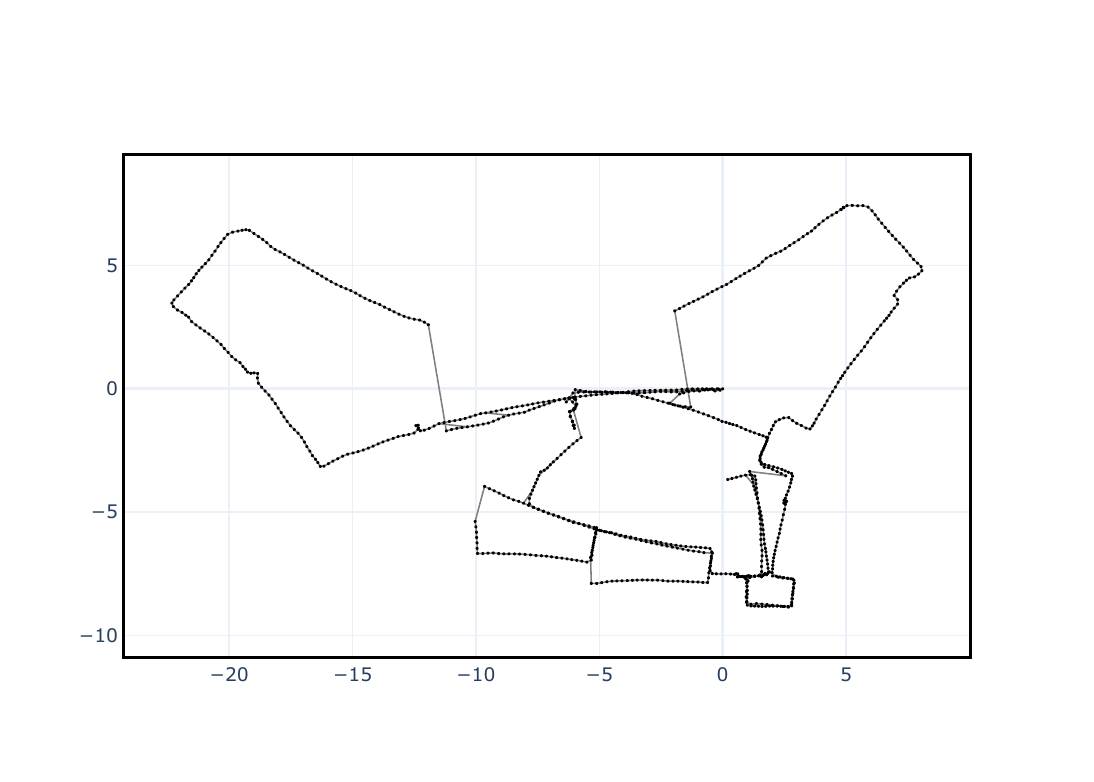} \\
  \includegraphics[width=0.21\textwidth, trim=58 58 58 58, clip]{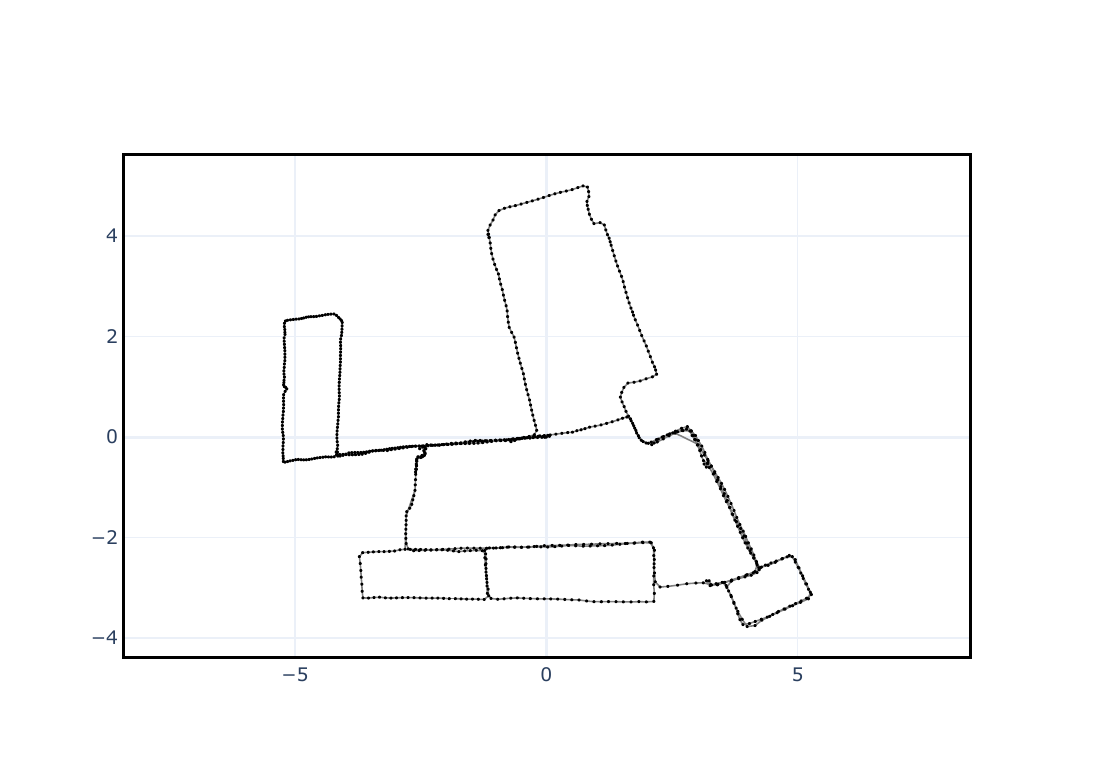} \\
  \includegraphics[width=0.21\textwidth, trim=58 58 58 58, clip]{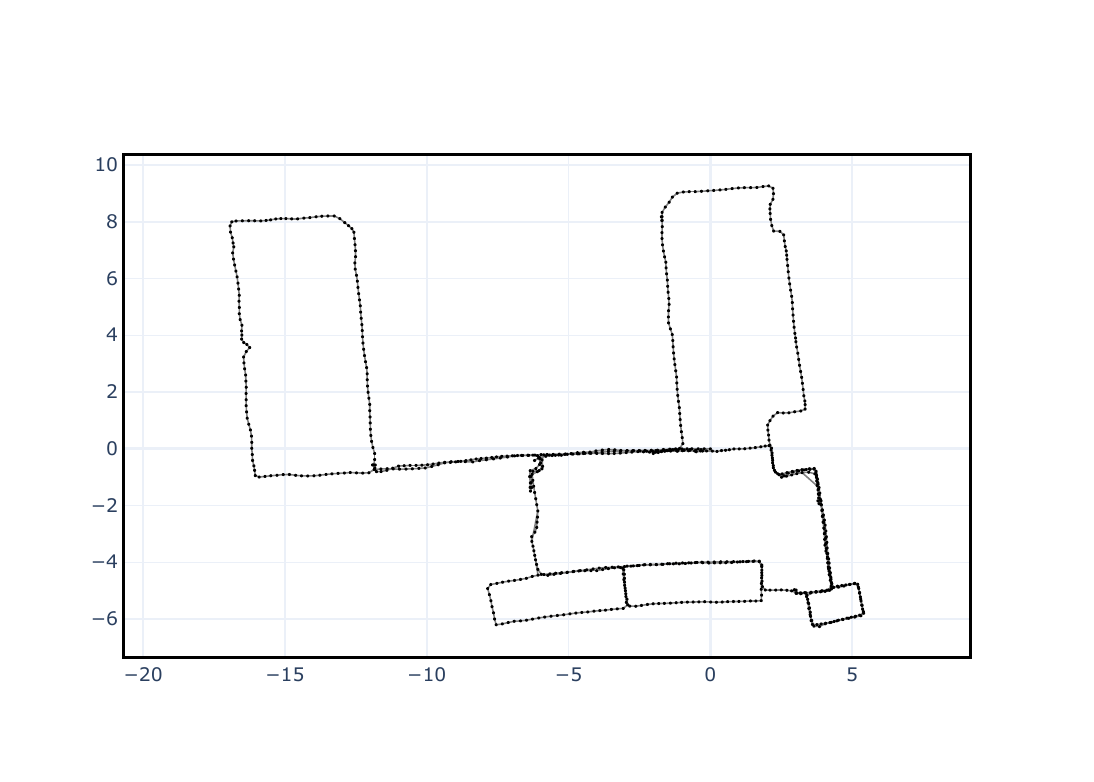}
  \caption{MIT dataset. Top panel shows a spanning tree initialization of an $\SE{2}$
    \ac{PGO} problem, which is not in the basin of attraction of the global optimizer;
    the middle panel shows our \emph{linear} \fastsync{} initialization;
    bottom panel shows the subsequent \emph{local} optimization starting from \fastsync{},
  converging to the global optimum.}
  \label{fig:banner_mit}
\end{figure}

\IEEEpubidadjcol

In this paper we develop a fast and general approximate \ac{GS} method that applies to
\textit{any} matrix Lie group, with a focus on providing high-quality initial estimates.
While local optimization is always an option, it requires a good initialization to avoid
getting trapped in local minima. Globally certified methods, available for special
cases~\cite{Rosen19ijrr_SE_Sync,Dellaert20eccv_Shonan}, also benefit from a fast
initialization. Beyond generalizing chordal initialization, we
introduce two key enhancements: (i) a Kronecker-product-based block reduction and
back-substitution that improves scalability, and (ii) a Nested-Dissection-based
gauge-fixing scheme that substantially improves initialization quality. We validate these
contributions through a small world analysis~\cite{Wilson20cvpr_Distribution}
demonstrating that our initialization very often finds the global basin of attraction, as
well as experimental evaluation on a wide range of simulated and real datasets across a
broad sample of Lie groups.

\section{RELATED WORK}

The dominant approach to \ac{GS} is \ac{MLE} under a
probabilistic generative model for the measurements. This formulation appears
in cryo-electron microscopy
\cite{Singer11acha_AngularSync,Singer11siam_Cryo}, 3D computer vision
\cite{Hartley13ijcv_RA}, and robotic mapping \cite{Lu97ar_GloballyConsistent}, among many
other applications. Unfortunately,
\ac{MLE} formulations of \ac{GS} problems are typically high-dimensional and non-convex,
and thus computationally hard.
Indeed, several important \ac{GS} instances, including Max-Cut
\cite{Goemans95jacm_SDP}, rotation averaging \cite{Bandeira17mp_MaxLikelihood}, and
pose-graph SLAM, are all known to be NP-hard in general,
necessitating the development of practical \emph{approximate} inference methods.

The most common approximation strategy replaces \emph{global} optimization with fast
\emph{local} optimization. Since the groups arising in practical applications are often
Lie groups (and thus smooth manifolds), this admits the efficient application of fast
first- or second-order smooth \emph{local} optimization algorithms
\cite{Boumal14jmlr_Manopt} even to very large-scale problems.
However, due to the underlying problem's nonconvexity, the critical point to which a
local optimizer ultimately converges depends upon
its initialization.  Consequently, this approach crucially relies upon \emph{effective
initialization strategies}.

Previous work on local-search initialization strategies for synchronization problems in
robotics and computer vision can be broadly categorized into three major classes.
\emph{Spanning tree} methods construct a rooted tree for the synchronization graph and
iteratively assign initial estimates by concatenating measurements along the edges of the
tree \cite{Martinec07cvpr_robust, Carlone15icra_initialization}. While simple and
scalable, these methods tend to be inaccurate due to the accumulation of error along long
paths.

\emph{Linear} initialization methods relax nonlinear manifold constraints to
either unit-norm or single-element gauge-fixing constraints, producing linear
least-squares approximations of the \ac{GS} problem that can be solved in closed-form
using standard matrix factorizations.  This approach has proven
effective for sparse 3D spatial estimation problems, being commonly applied to
rotation synchronization using both quaternion \cite{Govindu01cvpr_MotionEstimation}
and matrix representations \cite{Martinec07cvpr_robust}, as well as to special Euclidean
group synchronization
\cite{Govindu01cvpr_MotionEstimation,Martinec07cvpr_robust,Carlone15icra_initialization}.

\emph{Spectral} methods construct estimates from extremal eigenvectors of
a generalized Laplacian associated with the underlying
graph \cite{Arrigoni20ijcv_Synchronization}. This approach was pioneered
by Singer \cite{Singer11acha_AngularSync}, who studied its application to angular
synchronization, and has subsequently been extended to a wide range of other \ac{GS}
applications (cf.\ \cite{Arrigoni20ijcv_Synchronization} and the references therein). Spectral methods are attractive for their generality, as they can be easily applied to \emph{any}
matrix group possessing an efficiently-computable metric
projection (i.e.\ \emph{rounding}) method
(cf.\ \cite[Sec.\ 5]{Arrigoni20ijcv_Synchronization}).  However, scaling spectral
initialization to large problems requires the use of iterative eigenvalue methods whose
convergence rates depend upon the distribution of eigenvalues in the underlying graph's
spectrum.
The practical consequence is that spectral methods can sometimes be slow
to converge, especially on sparse graphs where linear initialization methods employing
sparse factorization excel.

Finally, as an alternative to local search, some recent work has explored the development
of \emph{certifiably correct} estimation methods
that aim to perform tractable direct \emph{global} optimization in certain cases.  These
techniques, which are based upon \emph{convex relaxation}, are provably
capable of recovering \emph{verifiably globally optimal} \ac{GS} solutions under mild
conditions. Originating in Goemans and Williamson's Max-Cut work \cite{Goemans95jacm_SDP}, these
methods have subsequently been extended to various synchronization problems over
rotations and orthogonal matrices
\cite{Singer11siam_Cryo,Dellaert20eccv_Shonan}
and the special Euclidean group \cite{Rosen19ijrr_SE_Sync}.  These methods are
attractive for their exceptionally strong theoretical robustness guarantees and practical
performance.  However, solving the large-scale relaxations underpinning certifiable
estimators remains a formidable computational challenge, making this class of methods
difficult to deploy in practice.

Our proposed \fastsync{} method generalizes the linear initialization schemes of prior work
from specific groups like \SO{d} and \SE{d} to arbitrary matrix Lie groups. We incorporate
novel algorithmic enhancements including Kronecker structure exploitation for improved
efficiency and a novel gauge selection strategy that significantly improves initialization
quality. As such, we provide a simple, general, and efficient approach for
high-quality \ac{GS} initialization, directly extending the philosophy of motion averaging
pioneered by Govindu~\cite{Govindu04cvpr_LieAlgebraic} and the
many \acl{GS} methods surveyed by Arrigoni et al.~\cite{Arrigoni20ijcv_Synchronization}.

\makeatletter
\@startsection{section}{1}{\z@}{3.0ex}{0.7ex}{\normalfont\normalsize\centering\scshape}{GROUP SYNCHRONIZATION}
\makeatother
\begin{figure}[t]
  \centering
  \vspace{-0.2cm}
  \includegraphics[width=0.40\textwidth]{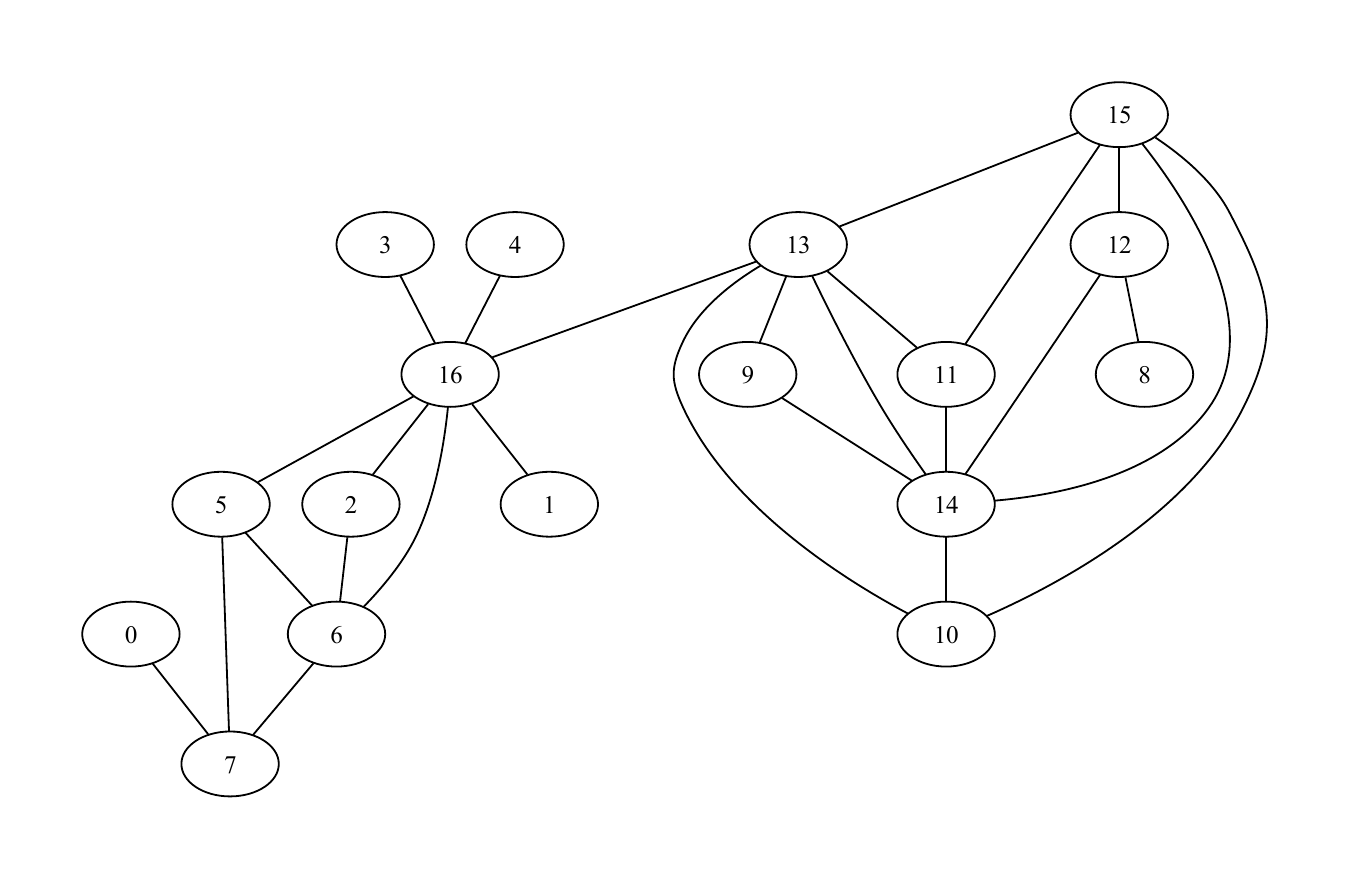}
  \vspace{-0.7cm}
  \caption{Measurement graph for a \ac{SfM} dataset (Statue of Liberty, $\numX=17$).}
  \label{fig:statue_graph}
\end{figure}

\subsubsection*{Preliminaries} Let $\GG$ be a matrix Lie group.
Its Lie algebra $\mathfrak{g}$ is a vector space isomorphic to
$\mathbb{R}^\dimG$ for $\dimG=\dim(\mathfrak{g})$. We denote the isomorphism from the
vector representation to the algebra's matrix representation as the wedge operator,
$[\cdot]^\wedge: \mathbb{R}^\dimG \to \mathfrak{g}$, and its inverse as the vee operator,
$(\cdot)^\vee: \mathfrak{g} \to \mathbb{R}^\dimG$. The group exponential and logarithm
maps, $\Exp: \mathbb{R}^\dimG \to \GG$ and $\Log: \GG \to \mathbb{R}^\dimG$, are defined by
composing the standard matrix exponential/logarithm with these operators: $\Exp(v) \doteq
\exp([v]^\wedge)$ and $\Log(X) \doteq (\log X)^\vee$.

\subsubsection*{The \acf{GS} problem}
In our problem, the unknowns are a collection of $\numX$ states which we write as the state tuple
$\X \doteq (X_1,\dots,X_\numX) \in \GG^\numX$, where each
element of the group is $X_i \in \mathbb{R}^{d \times d}$. We are given a set of relative
measurements $\Zij$
between pairs of states, defining a measurement graph $G$, an example of which is shown
in Figure~\ref{fig:statue_graph}.
We adopt the \emph{isotropic concentrated Gaussian (ICG)} noise
model, which assumes left-invariant noise on the relative measurement. This gives the
measurement model
\begin{equation}
  \label{measurement_model}
  \Zij = X_i^{-1} X_j\,\Exp(\eij),
\end{equation}
where the noise vector $\eij \in \mathbb{R}^{\dimG}$ is drawn from a zero-mean
Gaussian, $\eij\sim\mathcal{N}(0,\sigma_{ij}^2 I_{\dimG})$. The per-edge weight is the
inverse variance, $\kappa_{ij} \doteq \sigma_{ij}^{-2}$.
The maximum-likelihood estimate $\X^*$ is then given by
\begin{align}
  \label{eq:MLE}
  \X^* &= \argmin_X \sum_{(i,j)} \tfrac12 \,\kappa_{ij}\, \big\| \eij \big\|_2^2 \\
  &= \argmin_\X \sum_{(i,j)} \tfrac12 \,\kappa_{ij}\, \big\| \Log \big(X_j^{-1} X_i
  \Zij\big) \big\|_2^2,
\end{align}
where we have assumed that the errors $\eij$ are small enough that $\Log$ and $\Exp$ are
inverses on the support.

\subsubsection*{Frobenius error}
\fastsync{}, introduced below, is based on the \textit{Frobenius norm}, rather than the
Lie-algebraic distance above. This is valid, as for small errors $\eij$,
\begin{equation}
  \Exp(\eij)\approx I+[\eij]^\wedge,
\end{equation}
so minimizing
$\|\eij\|^2$ is (to first order) equivalent to minimizing the matrix-space
surrogate
\begin{equation}
  \phi_{ij}^{\mathrm{F}} \;\doteq\; \kappa_{ij}\, \FF{I-X_j^{-1} X_i \Zij}^2.
  \label{eq:original_error}
\end{equation}
Left-multiplying the argument in \eqref{eq:original_error} by the (invertible) matrix
$X_j$ produces the following \emph{quadratic approximation}:
\begin{equation}
  \phi_{ij}^{\mathrm{F}} \approx \kappa_{ij}\, \FF{X_j-X_i \Zij}^2.
  \label{eq:frobenius_error}
\end{equation}
Note that \eqref{eq:frobenius_error} is actually equal to \eqref{eq:original_error} for many
groups common in robotics (e.g., $\SO{d}$, $\SE{d}$, $\SU{2}$), and serves as a practical
heuristic surrogate for more general matrix Lie groups; we evaluate this approximation
empirically below.

The cost function derived from employing the quadratic surrogate loss
\eqref{eq:frobenius_error} is thus given by:
\begin{equation}
  \label{eq:J_F}
  J_F(\X) = \sum_{(i,j)} \tfrac12 \,\kappa_{ij}\, \FF{X_j-X_i \Zij}^2,
\end{equation}
with the constraint that $\X\in\GG^\numX$.

\section{FAST-SYNC}

Since the cost function \eqref{eq:MLE} is non-convex, we can, given an initial estimate
$\X^0$, optimize locally to the nearest minimum. This can be done using a manifold
optimization framework such as Manopt~\cite{Boumal14jmlr_Manopt} or
GTSAM~\cite{Dellaert12techreport_GTSAM}, or by
employing motion averaging~\cite{Govindu04cvpr_LieAlgebraic}.
However, a good initial estimate is crucial to ensure convergence to the global minimum.

We now show how to use the surrogate Frobenius cost \eqref{eq:J_F} to build a fast, linear
initialization method, building upon the work by
Govindu~\cite{Govindu01cvpr_MotionEstimation} and
Martinec and Pajdla~\cite{Martinec07cvpr_robust}.

\subsection{The Measurement Matrix}
As a first step, note that we can express the cost in \eqref{eq:J_F} in a compact matricized form:
\begin{equation}
  \label{eq:Hd_x}
  J_F(x) =  \|\Hd\, x\|^2,
\end{equation}
obtained by writing $\X\in \GG^\numX$ in vectorized form:
\begin{equation}
  x =
  \begin{pmatrix}
    \vecc(X_1) \\
    \vdots \\
    \vecc(X_\numX)
  \end{pmatrix} \in \RR{d^2 \numX}.
\end{equation}
To see this, note that for a single edge $(i,j)$, we have
\begin{equation}
  \|X_j - X_i\,\Zij\|_F^2 = \|\vecc(X_j) - \vecc(X_i\,\Zij)\|^2.
\end{equation}
Using the identity
$\vecc${\footnotesize$(X_i\,\Zij)$}$
\;=\;${\footnotesize$ (\Zij^T \otimes I_d)$}$\,\vecc${\footnotesize$(X_i),$}
the contribution to the objective can be written as
\begin{equation}
  \|\vecc(X_j) - (\Zij^T \otimes I_d)\,\vecc(X_i)\|^2 \;=\; \|\Hdk x\|^2,
\end{equation}
where $\Hdk = \Hgk \otimes I_d$, and $\otimes$ denotes the Kronecker product.
The $d\times \numX d$ block row $\Hgk$ has $d\times d$ block columns indexed by
$\ell=1,\dots,\numX$. Its $\ell$-th block column is
\begin{equation}
  (\Hgk)_\ell =
  \begin{cases}
    -\Zij^T,       & \text{if } \ell = i,\\
    I_d,           & \text{if } \ell = j,\\
    0_{d\times d}, & \text{otherwise.}
  \end{cases}
\end{equation}
By stacking these block rows for all edges, we obtain the graph-derived matrix
$\Hg \in \RR{\numZ d \times \numX d}$, and its \textit{Kronecker extension}
$  \Hd = \Hg \otimes I_d
$
of size $d^2 \numZ \times d^2 \numX$.
\subsection{Fast Approximate Synchronization}

Minimizing \eqref{eq:Hd_x} over $X \in \GG^N$ is still non-convex, as each $X_i \in \GG$
derived from $x$ is constrained to be in the group $\GG$. Furthermore, the homogeneous
least-squares form admits a trivial all-zeros solution unless additional constraints are
imposed, so we introduce a gauge fixing step to resolve this ambiguity. This approach is
akin to chordal initialization in \ac{RA}~\cite{Martinec07cvpr_robust,
Carlone15icra_initialization} and often yields a good approximation of the optimum once the
blocks are projected back onto $\GG$.
How exactly we impose the gauge and solve the
system is crucial for both performance and initialization quality.

\textbf{Step 1: Reduced Problem via Kronecker Structure}
First, we observe that we can solve a much smaller problem. The key computational bottleneck is the
(sparse) factorization of the matrix $\Hd$. Not surprisingly,
the Kronecker structure $\Hd = \Hg \otimes I_d$ allows us to factor the $d$-times smaller
matrix $\Hg$ instead, as each row of the linear
system $X_j = X_i\,\Zij$ has the common coefficient matrix $\Zij$.

\begin{figure*}
  \centering
  \begin{tabular}{cc}
    \includegraphics[width=0.25\textwidth]{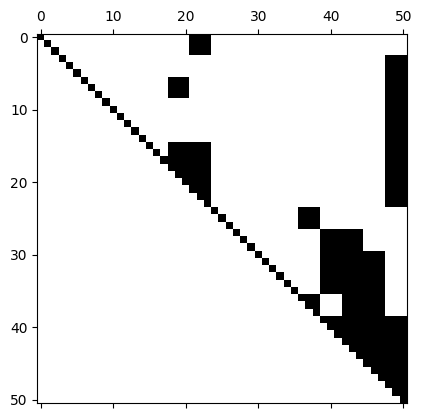} &
    \includegraphics[width=0.50\textwidth]{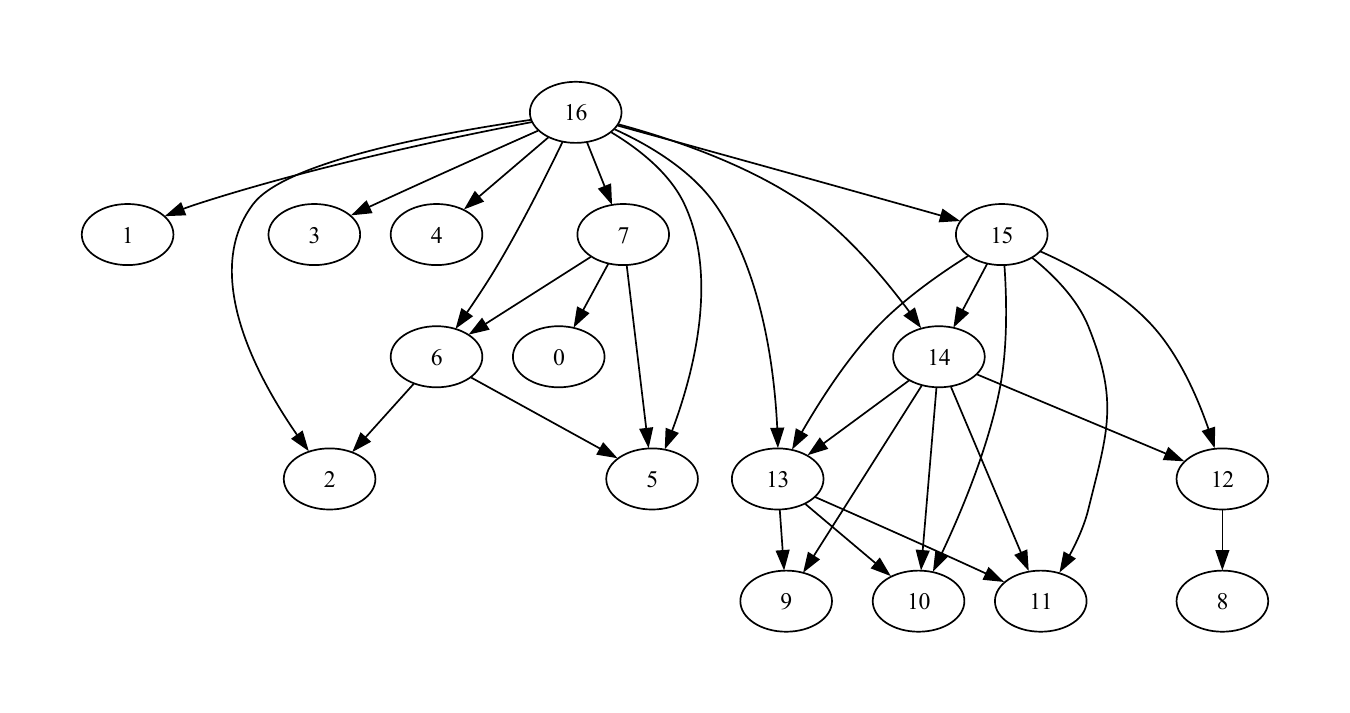} \\
    (a) $R$ Block structure & (b) Elimination \acs{DAG}
  \end{tabular}
  \caption{Statue of Liberty dataset: the $3\numX \times 3\numX$ Cholesky factor $R$ (left)
    corresponding to the back-substitution \ac{DAG} (right) obtained using a \acf{ND}
    ordering heuristic.
    Compare with Figure \ref{fig:statue_graph}. There and in the \ac{DAG} above we
    numbered the vertices and the block columns of $R$ according to the \ac{ND} ordering.
    The root (16) will be assigned the group identity.
    The vertices directly connected to the root will lie exactly on the group $\GG$, whereas
    deeper nodes may move off the manifold in the ambient-space back-substitution process of
  equation~\eqref{eq:suggest}.}
  \label{fig:dag_R}
\end{figure*}

\textbf{Step 2: Gauge Fixing and Augmented System}
The system 
is rank-deficient because of the gauge freedom inherent in \ac{GS}, which we fix
by adding a $d\times \numX d$ block-row to $\Hg$ with an identity matrix
$I_d$ in the last block. The augmented matrix is denoted as $A$. For each column-wise solve, we construct a RHS vector $b$ included in $\RR{(\numZ+1)d}$ that is zero except
for the last $d$ entries, which are set to the corresponding column of $I_d$. This yields the least-squares objective
\begin{equation}
  \label{eq:Abg}
  \hat y = \argmin_y \|A y - b\|^2,
\end{equation}
where $y$ is a stand-in for one column.

\textit{The quality of the obtained solution can vary dramatically
depending on which group element is set to $I_d$}. Intuitively, the linear
solve does not guarantee that the recovered estimates $\hat X_i$ will lie in the group
$\GG$, except for the noiseless case. However, the chosen gauge is always in $\GG$,
by construction, and elements ``close to" the gauge will be closer to the group
manifold. Below we make these notions more precise, and show how to choose a gauge that
is ``central".

We improve both performance \textit{and} the quality of the solution by using a clever
column ordering scheme for the sparse solver.
The cost of sparse QR/Cholesky strongly depends on the column ordering; finding
a fill-minimizing ordering is NP-complete~\cite{Yannakakis81siam_NPComplete}, so practical
solvers rely on heuristics, such as \acl{AMD}~\cite{Davis08toms_CHOLMOD}.

Our strategy uses two heuristics:
\textit{Block preservation:} We never split the $d\times d$ unknown blocks
$\{X_i\}$ but only order the $\numX$ block columns. Let $P$ be the resulting
block-permutation; we solve $\min_{\mathcal X}\|H P^\top \, \mathcal Y\|_F$
with $\mathcal Y = P \mathcal X$, which yields identical optima but far less fill-in.
\textit{\Acf{ND}:} We recursively split the
measurement graph $G$ with small vertex
separators~\cite{George73siam_NestedDissection,Lipton79siam_GeneralizedNestedDissection},
which yields a block permutation $P$ that preserves $d\times d$ structure while
minimizing separator sizes, significantly lowering fill-in.

\textit{The \ac{ND} ordering is key to achieving good results.}
$A$ is re-ordered so the last vertex in the ordering corresponds
to the last block-column, and hence assigned the group identity $I_d$ during
back-substitution.
This is illustrated in Fig.~\ref{fig:dag_R}, where we show the Cholesky factor $R$ and the
block-elimination \ac{DAG} for the dataset from Figure~\ref{fig:statue_graph}.
The \ac{ND} ordering tends to minimize the distance from any
group element to the root (the gauge), effectively limiting the ``group distortion''
introduced during synchronization.

\textbf{Step 3: QR Factorization of the Reduced System}
After fixing the gauge,  we can solve the system $(A,b)$ in \eqref{eq:Abg} efficiently
and in a stable way using the (economy) QR factorization of the reduced, gauge-augmented matrix $A$:
\begin{equation}
  A = Q R,\qquad Q\in\mathbb{R}^{(\numZ+1)d\times \numX
  d},\ \ R\in\mathbb{R}^{\numX d\times \numX d},
\end{equation}
with $Q^{\top}Q=I_{\numX d}$ column-orthonormal.
Using this factorization, solving the least-squares problem reduces to the triangular solve,
$
R\,\hat y = Q^{\top} b.
$
However, this is the reduced size-problem. The final solution $\hat x$
is obtained by the recursive back-substitution procedure described next.

\textbf{Step 4: Efficient Back-Substitution}
\newcommand{\nzj}{\mathop{nz}_j}
Fixing the gauge implies that we can set $\hat X_\numX = I_d$, and the remaining unknowns
satisfy the following equality:
\begin{equation}
  \hat X_j R_{jj}^T + \sum_{k=j+1}^{\numX} \hat X_k R_{jk}^T = 0, \quad j=1,\dots,\numX-1,
\end{equation}
where each $R_{ij}$ is a $d \times d$ block of the reduced factor $R$, and where
the diagonal
blocks $R_{jj}$ are invertible. Many if not most of these blocks will be zero.
Thus, for $j = \numX-1, \numX-2, \ldots, 1$, the estimates $\hat X_j$ are recovered recursively:
\begin{equation}
  \label{eq:backsub}
  \vspace{-0.2cm}
  \hat X_j = - \big( \sum_{k=j+1}^{\numX} \hat X_k R_{jk}^T \big) R_{jj}^{-T}.
\end{equation}
Equivalently, define
\[
  nz_j \doteq \{k>j : R_{jk}\neq 0_{d\times d}\},
  \qquad
  \hat{T}_{kj} \doteq -R_{jk}^T R_{jj}^{-T}.
\]
Then \eqref{eq:backsub} can be written as the linear combination
\begin{equation}
  \label{eq:suggest}
  \vspace{-0.3cm}
  \hat X_j = \sum_{k\in\nzj} \hat X_k \hat{T}_{kj}.
  \vspace{0.2cm}
\end{equation}
The matrices $\hat{T}_{kj}$ are matrix-valued coefficients, not scalar weights.
Thus, \eqref{eq:suggest} should not be interpreted as a group-theoretic averaging
operation on $\GG$; it is a linear combination in the ambient space produced by
triangular back-substitution. When non-unit edge weights are included, they enter this
relation through the factor $R$ of the corresponding weighted least-squares system,
rather than as scalar averaging weights in \eqref{eq:suggest}. In the noiseless case,
this recursion is consistent with the group-valued solution. In case of noise and
$|\nzj|>1$, the ambient linear combination may leave the group $\GG$, so we need a
rounding step:

\textbf{Step 5: Rounding}
Finally, we project each estimated matrix $\hat X_j$ onto $\GG$. For $\SO{d}$, we use the polar
decomposition:
\begin{equation}
  \hat R_j = U_j \,\Sigma_j \,V_j^T \quad \text{then} \quad \hat R_j \leftarrow U_j V_j^T,
\end{equation}
but any group $\GG$ can be equipped with a projection
$\Pi_\GG(\cdot)$.

\textbf{Step 6: Local Optimization}
At this stage, we can refine by running a local optimization method, e.g.,
a Riemannian gradient-descent style algorithm using
libraries such as Manopt~\cite{Boumal14jmlr_Manopt}, which directly operate on the
product manifold $\GG^\numX$, a Gauss-Newton or Levenberg-Marquardt solver as
implemented in toolkits such as GTSAM~\cite{Dellaert12techreport_GTSAM}, which employ
an accurate first-order linearization of the objective \eqref{eq:MLE}, or a
Govindu-style iteration in Lie algebra coordinates~\cite{Govindu04cvpr_LieAlgebraic}.

\begin{figure*}
  \centering
  \vspace*{4pt}
  \setlength{\tabcolsep}{4pt}
  \resizebox{\textwidth}{!}{%
    \begin{tabular}{cccccc}
  & \textbf{Local Success rate, rtol=0.1\%} & \textbf{Performance Profile, tau=1.00} &\textbf{Initialization time} & \textbf{Local Optimization time} & \textbf{Total time} \\

  \textbf{MST} &
  \includegraphics[width=0.4\textwidth]{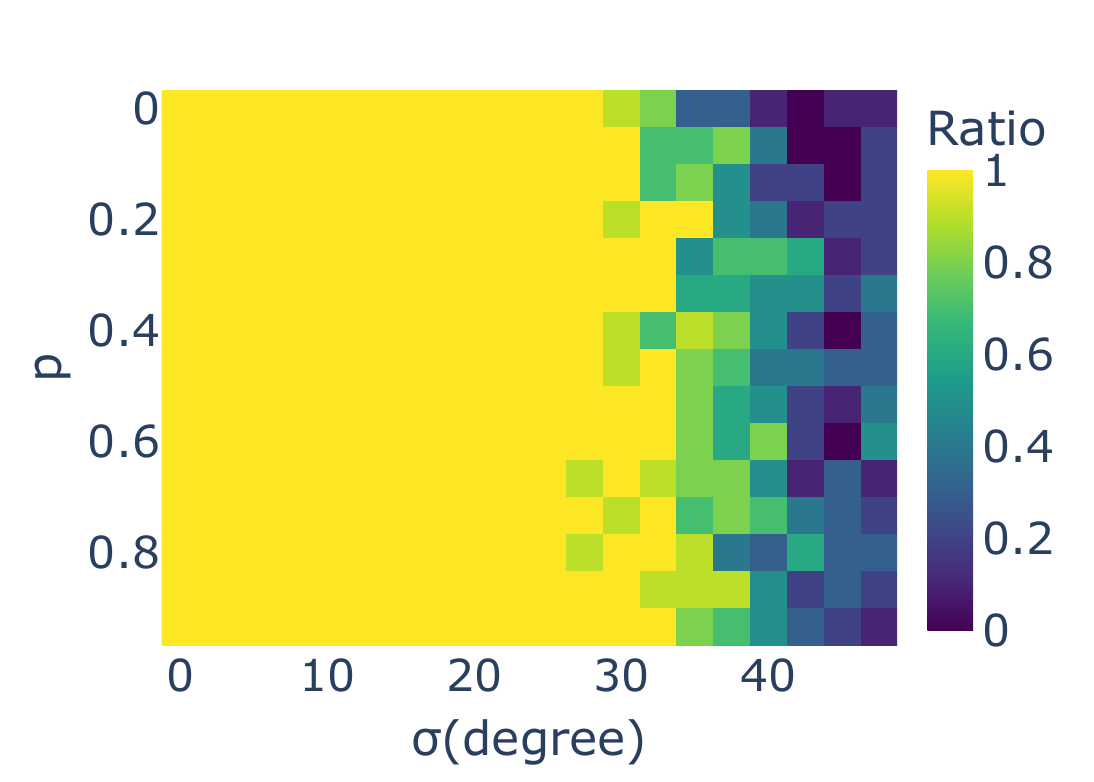} &
  \includegraphics[width=0.4\textwidth]{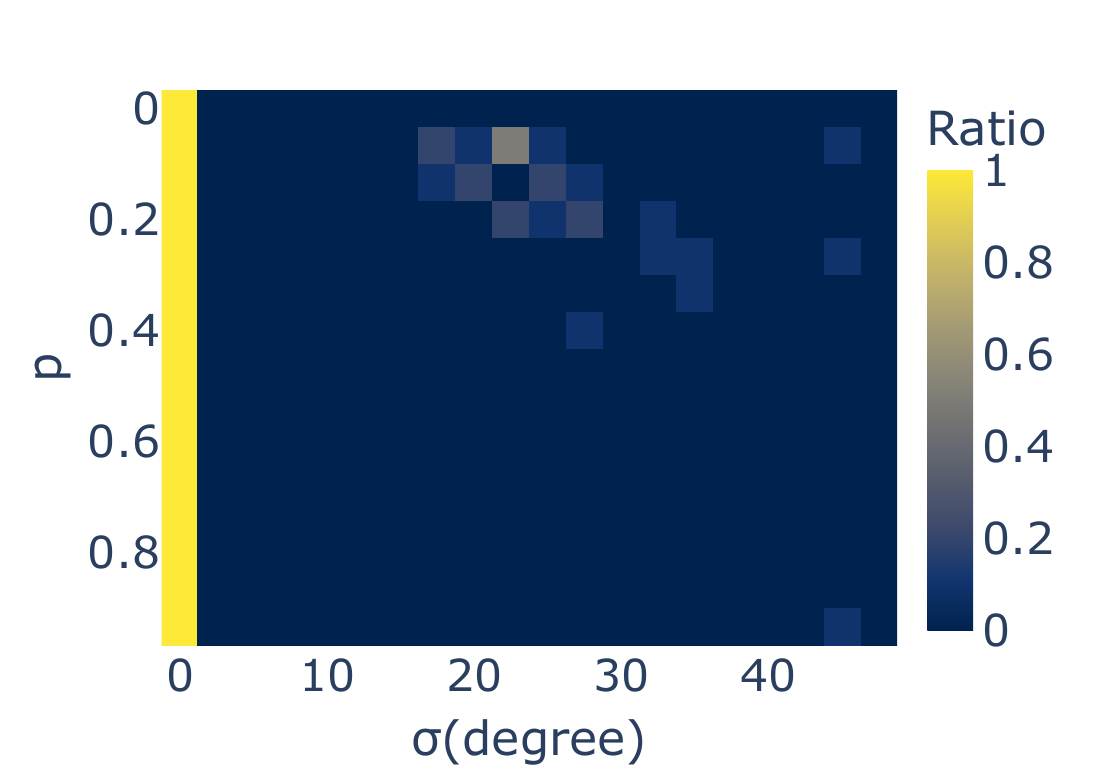} &
  \includegraphics[width=0.4\textwidth]{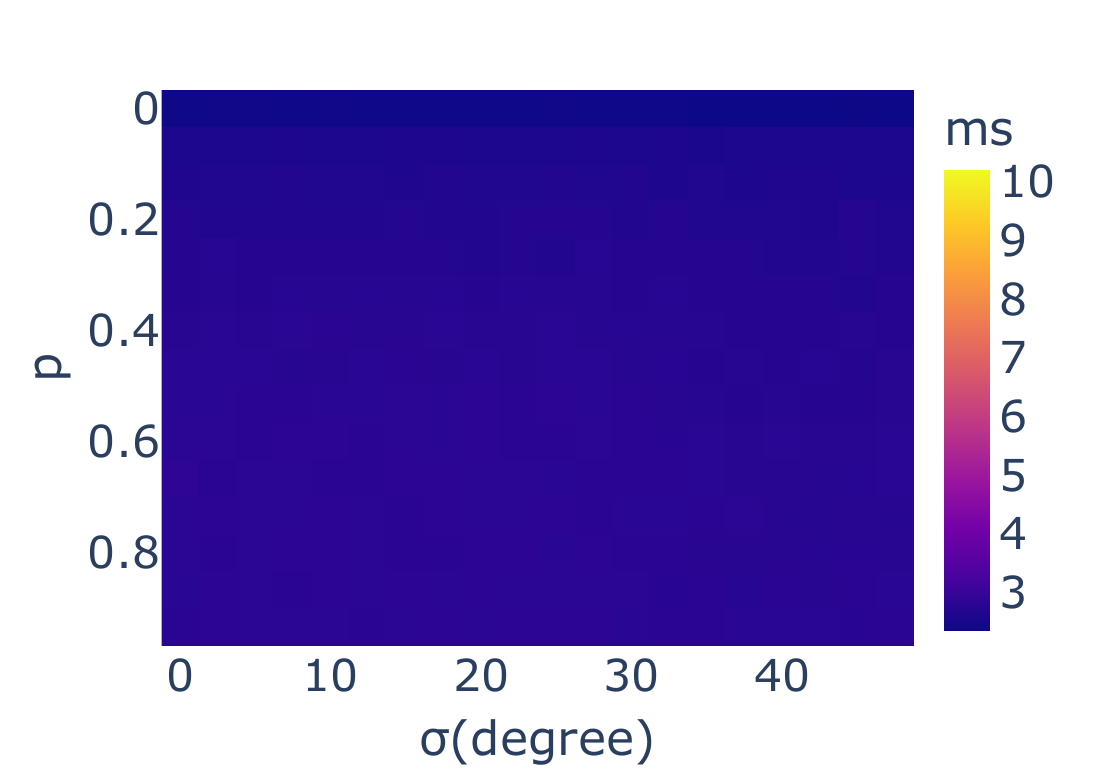} &
  \includegraphics[width=0.4\textwidth]{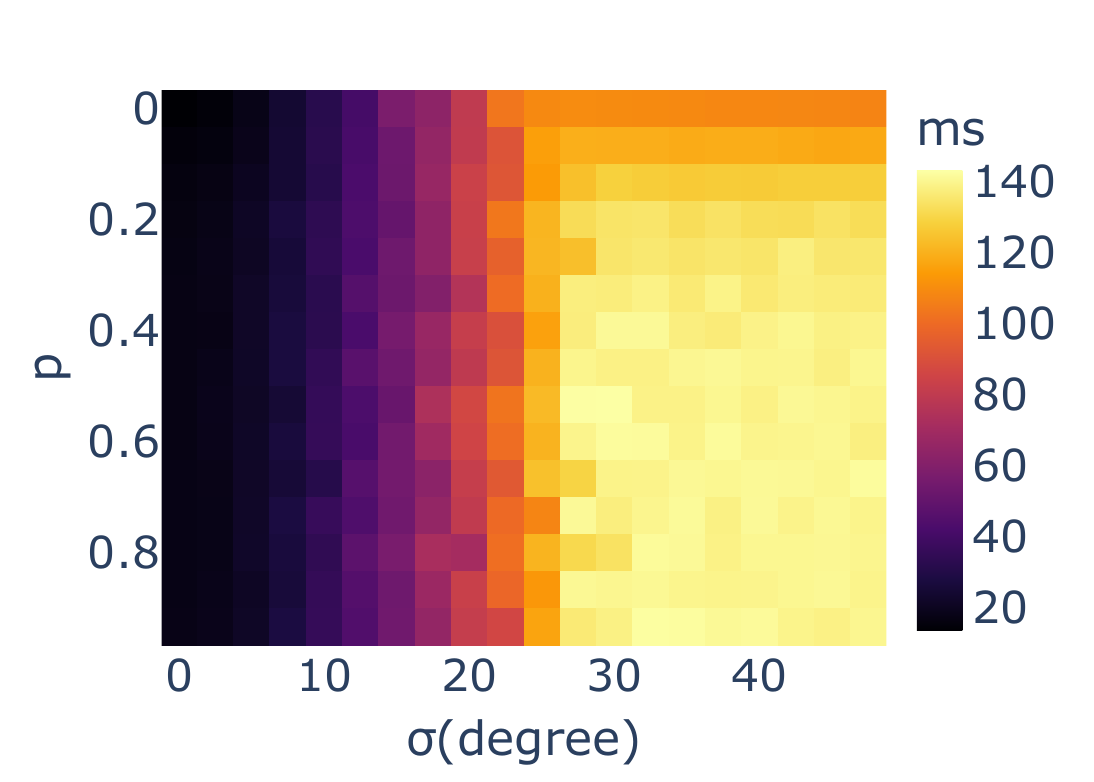} &
  \includegraphics[width=0.4\textwidth]{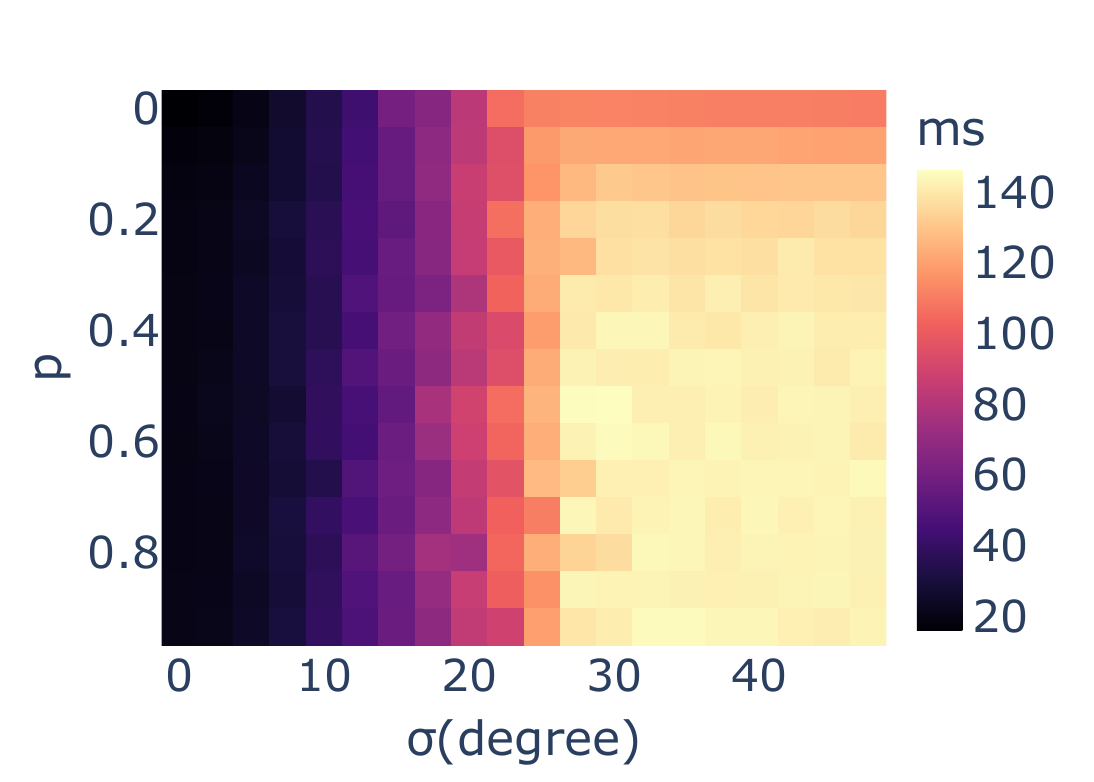} \\

  \textbf{Fast-Sync} &
  \includegraphics[width=0.4\textwidth]{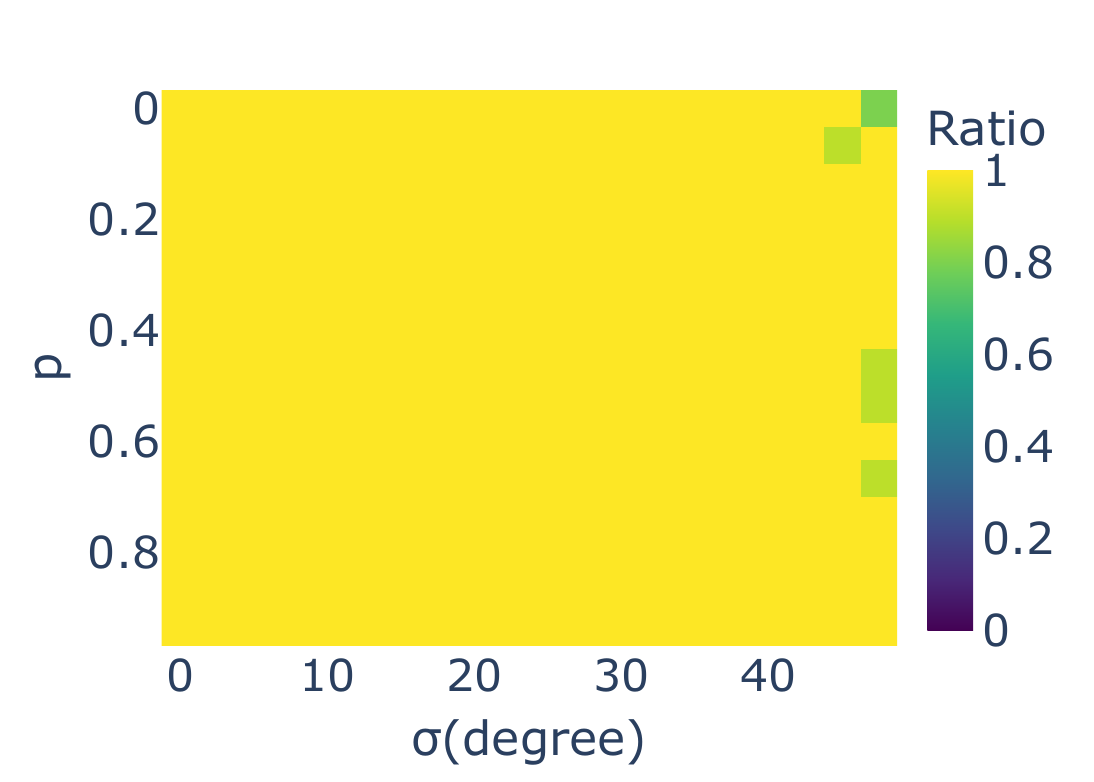} &
  \includegraphics[width=0.4\textwidth]{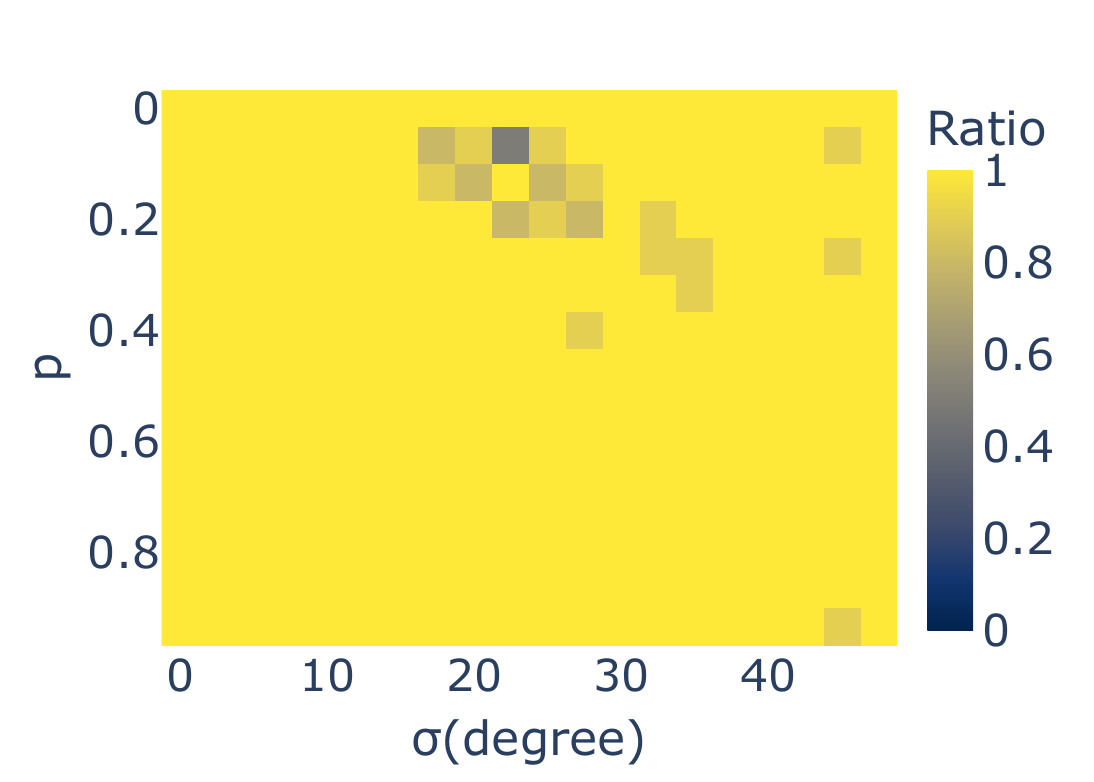} &
  \includegraphics[width=0.4\textwidth]{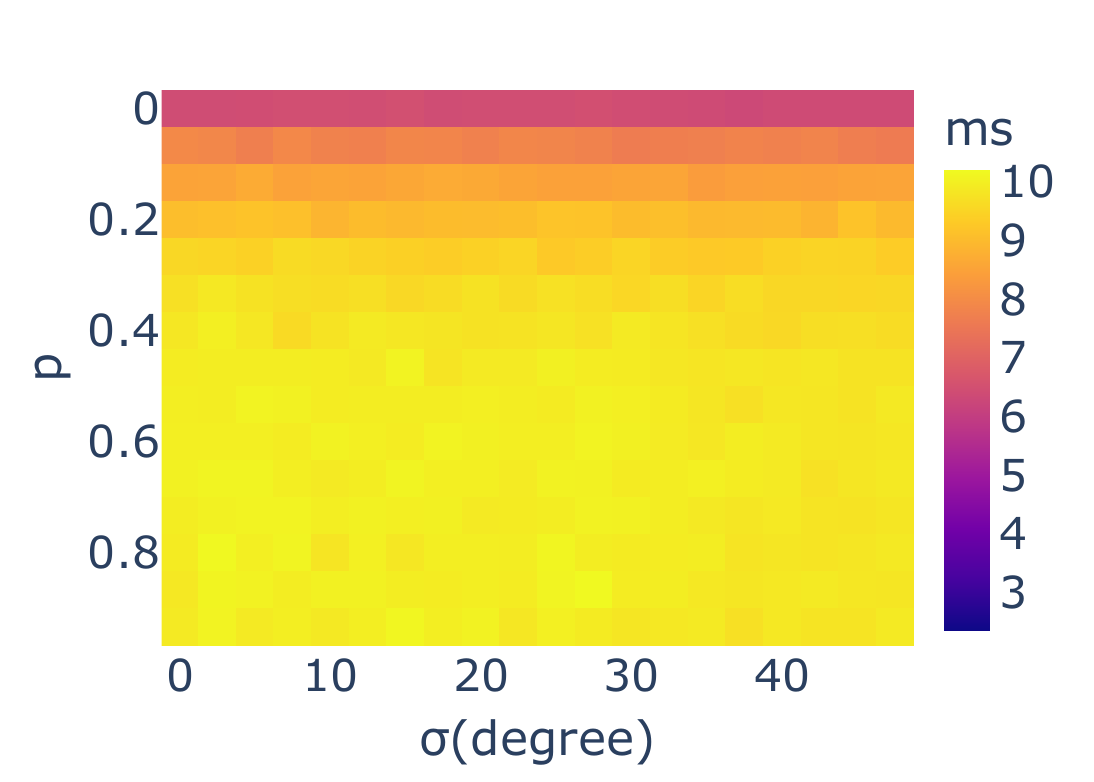} &
  \includegraphics[width=0.4\textwidth]{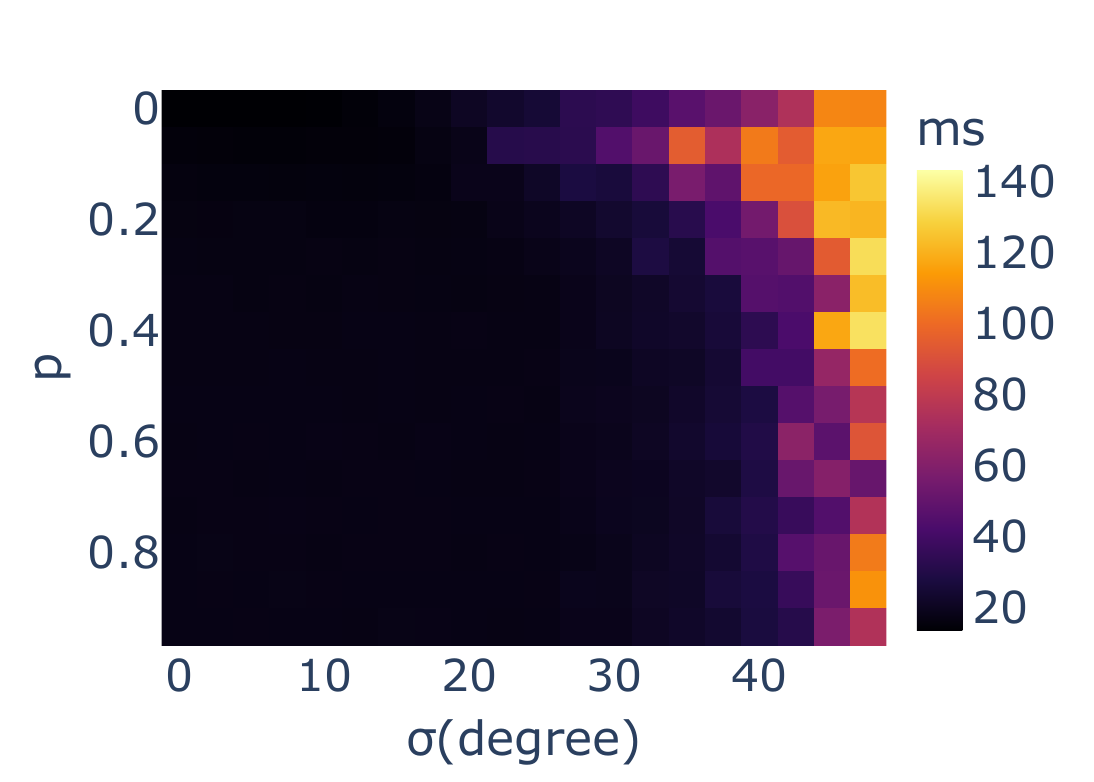} &
  \includegraphics[width=0.4\textwidth]{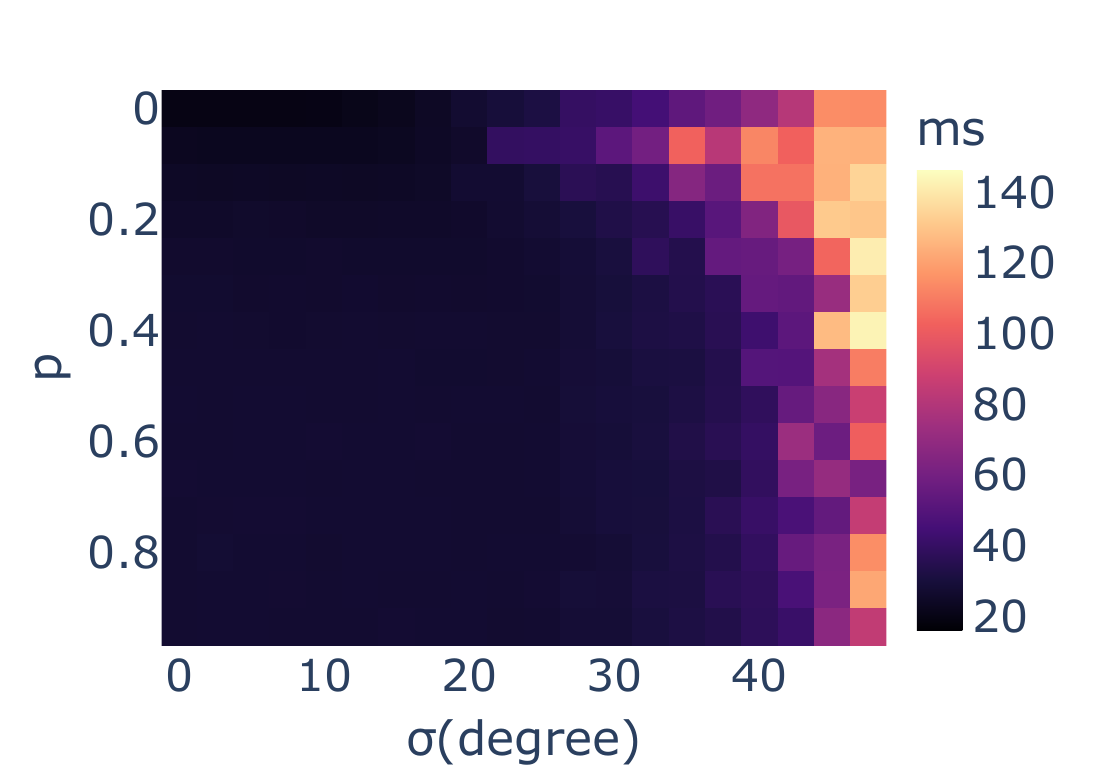} \\

\end{tabular}

  }
  \caption{Heatmap grid analysis of \SO{3}.}
  \label{fig:heatmap_so3}
\end{figure*}

\begin{figure*}
  \centering
  \vspace*{4pt}
  \setlength{\tabcolsep}{4pt}
  \resizebox{\textwidth}{!}{%
    \begin{tabular}{cccccc}
  & \textbf{Local Success rate, rtol=0.1\%} & \textbf{Performance Profile, tau=1.00} &\textbf{Initialization time} & \textbf{Local Optimization time} & \textbf{Total time} \\

  \textbf{MST} &
  \includegraphics[width=0.4\textwidth]{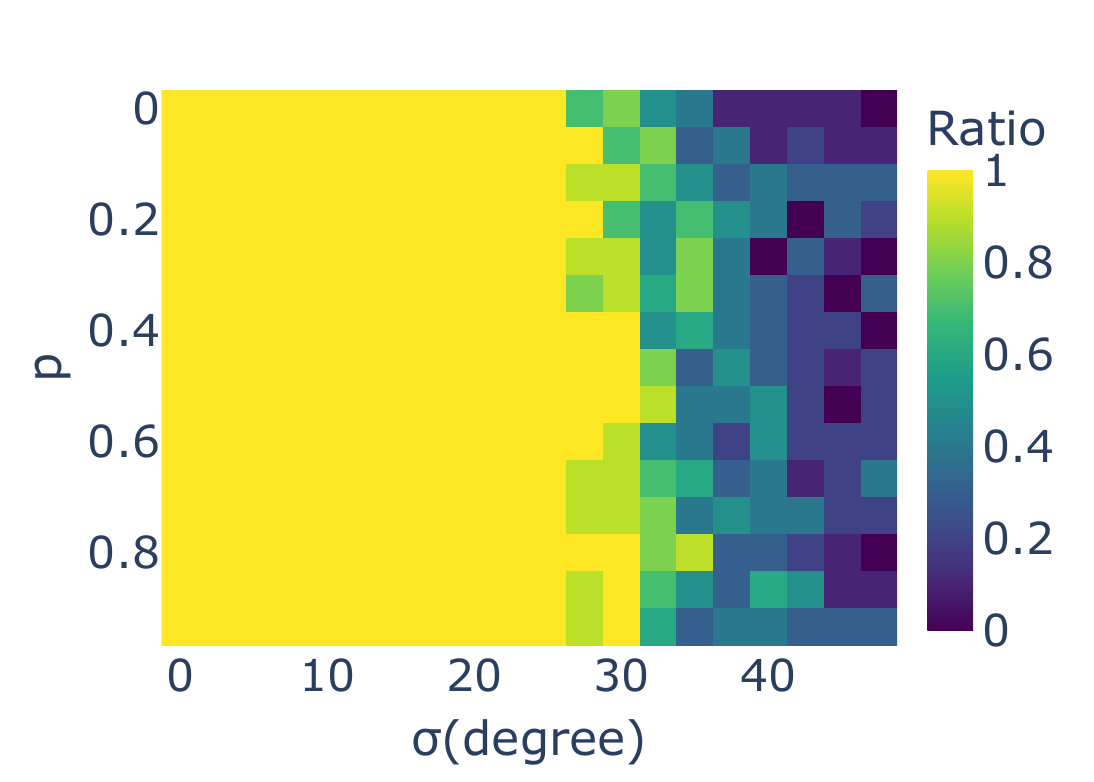} &
  \includegraphics[width=0.4\textwidth]{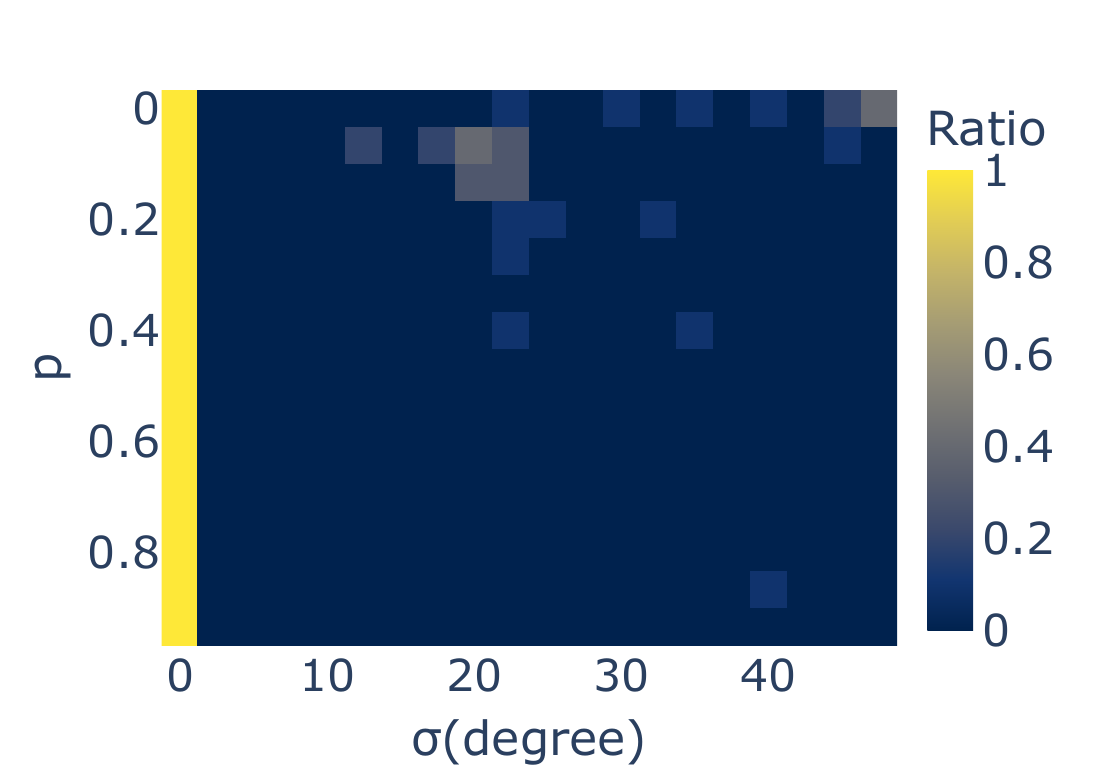} &
  \includegraphics[width=0.4\textwidth]{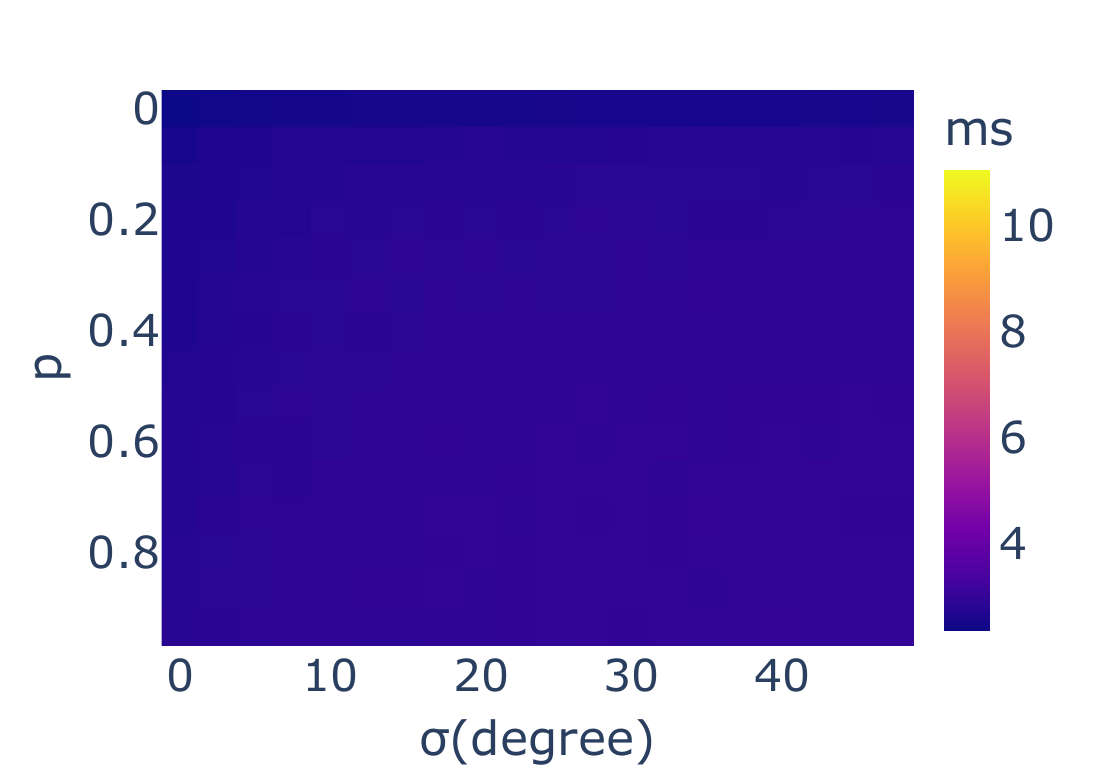} &
  \includegraphics[width=0.4\textwidth]{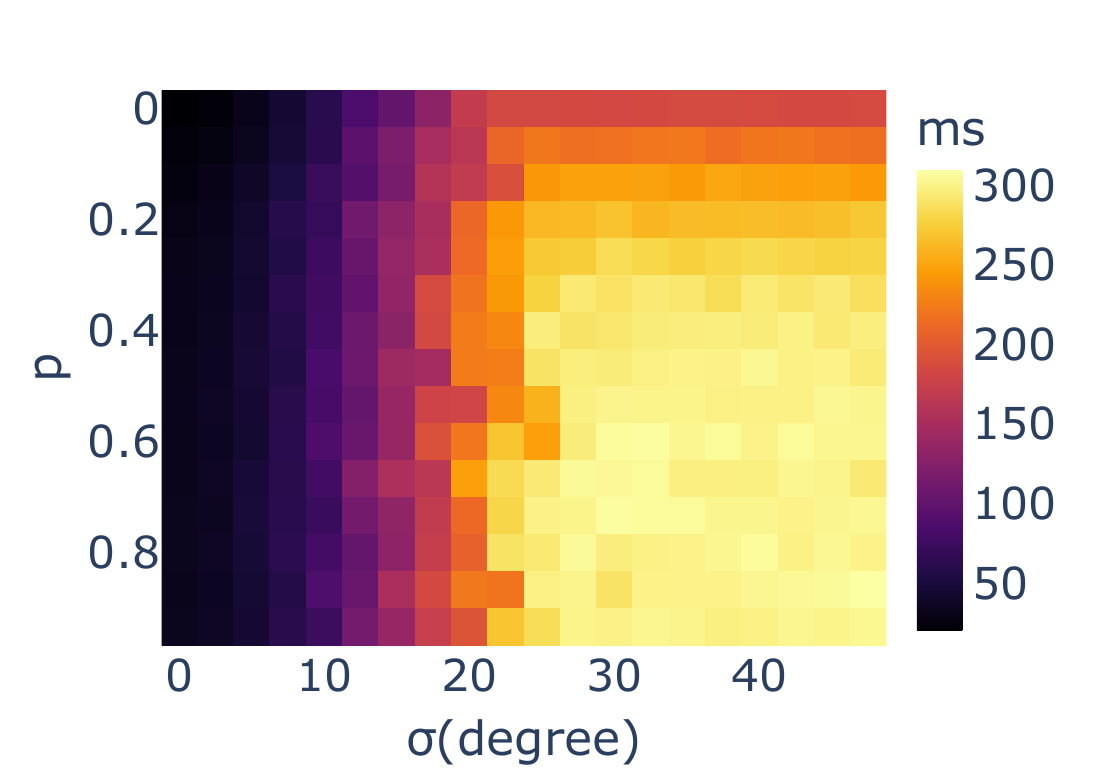} &
  \includegraphics[width=0.4\textwidth]{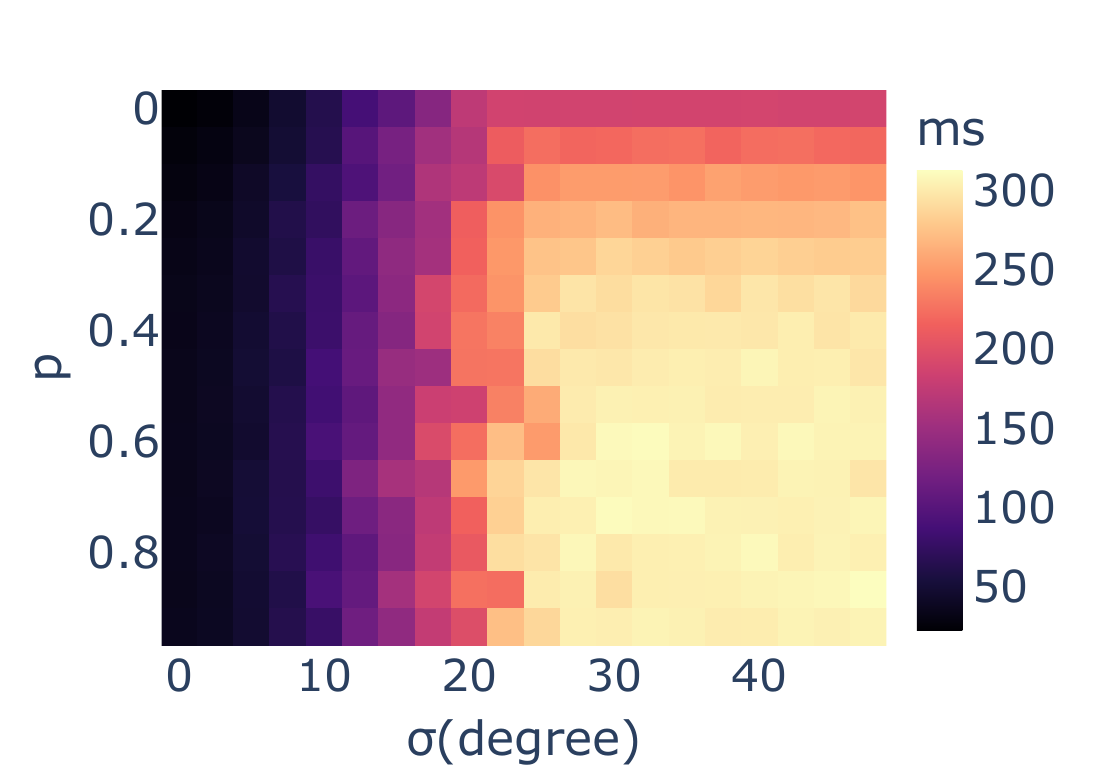} \\

  \textbf{Fast-Sync} &
  \includegraphics[width=0.4\textwidth]{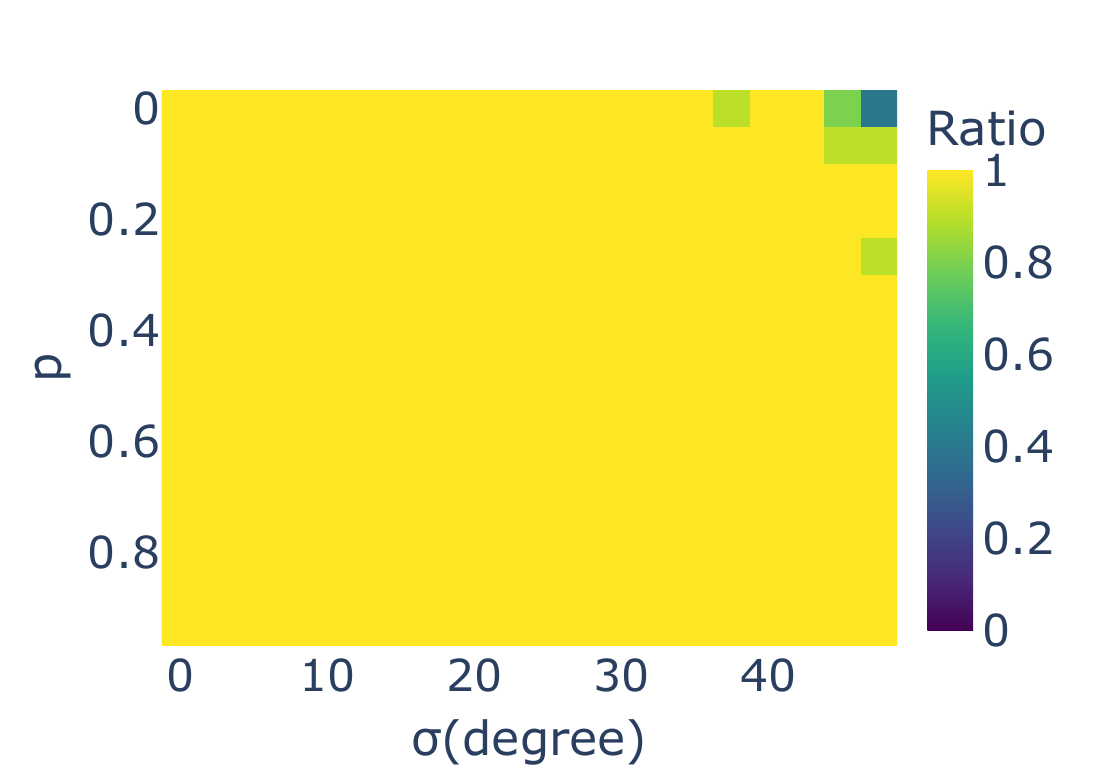} &
  \includegraphics[width=0.4\textwidth]{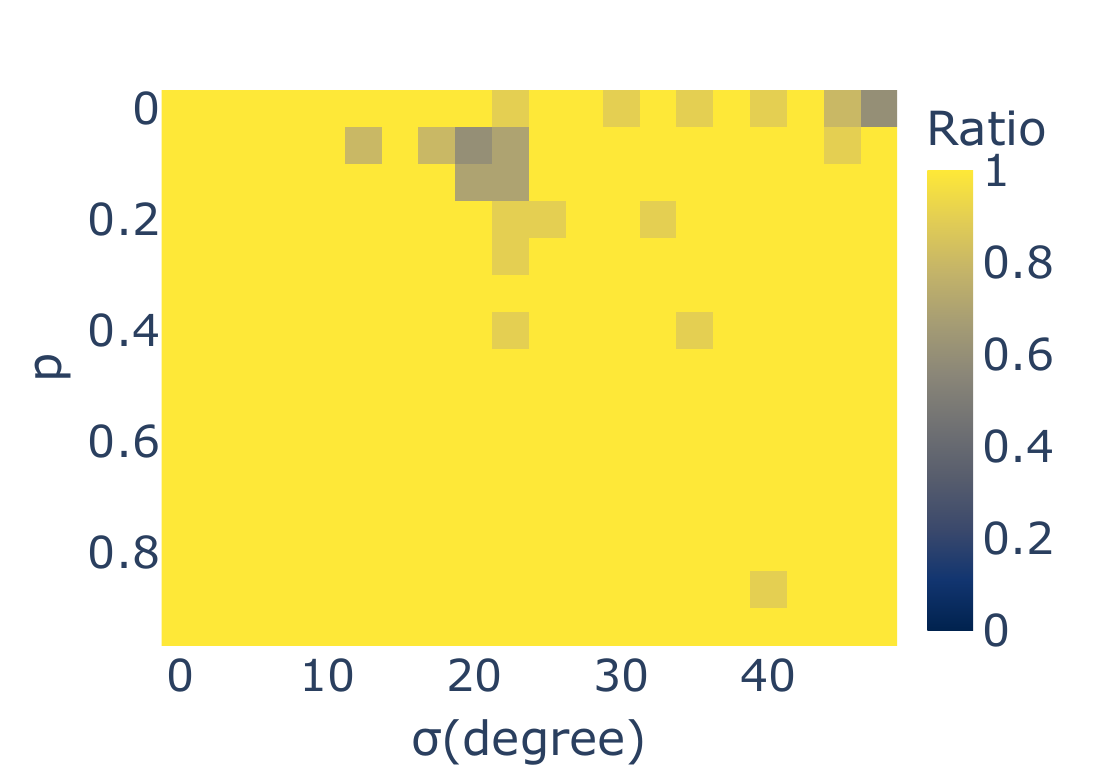} &
  \includegraphics[width=0.4\textwidth]{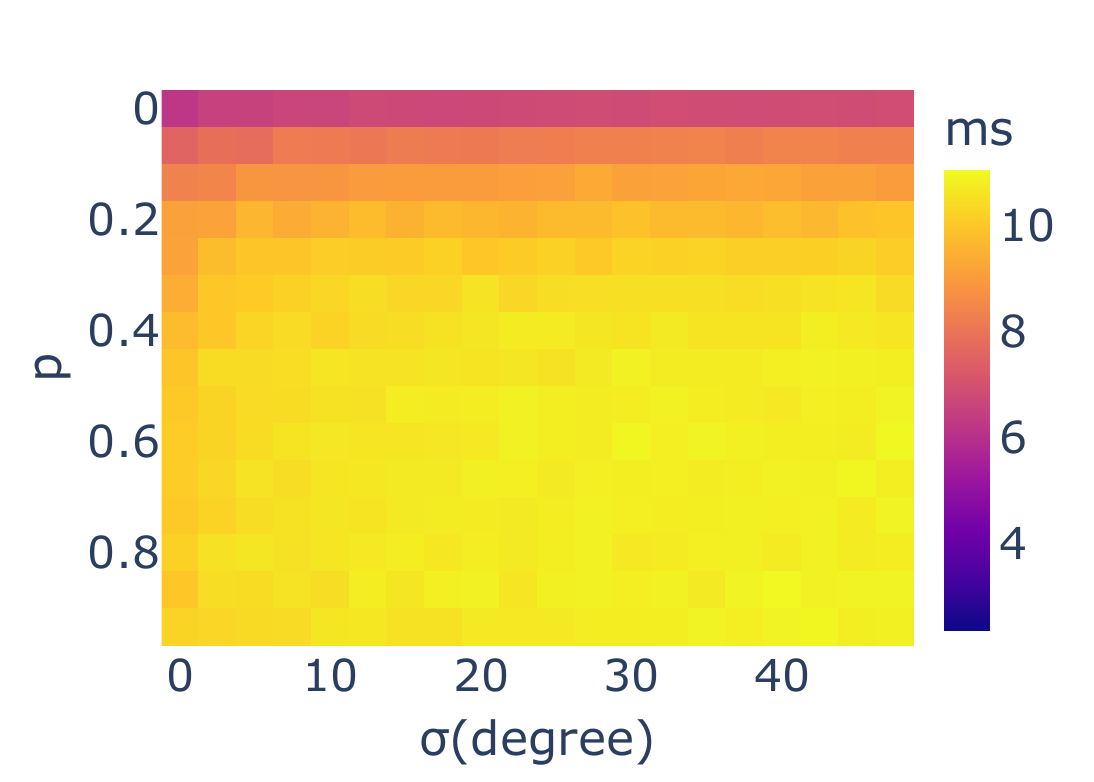} &
  \includegraphics[width=0.4\textwidth]{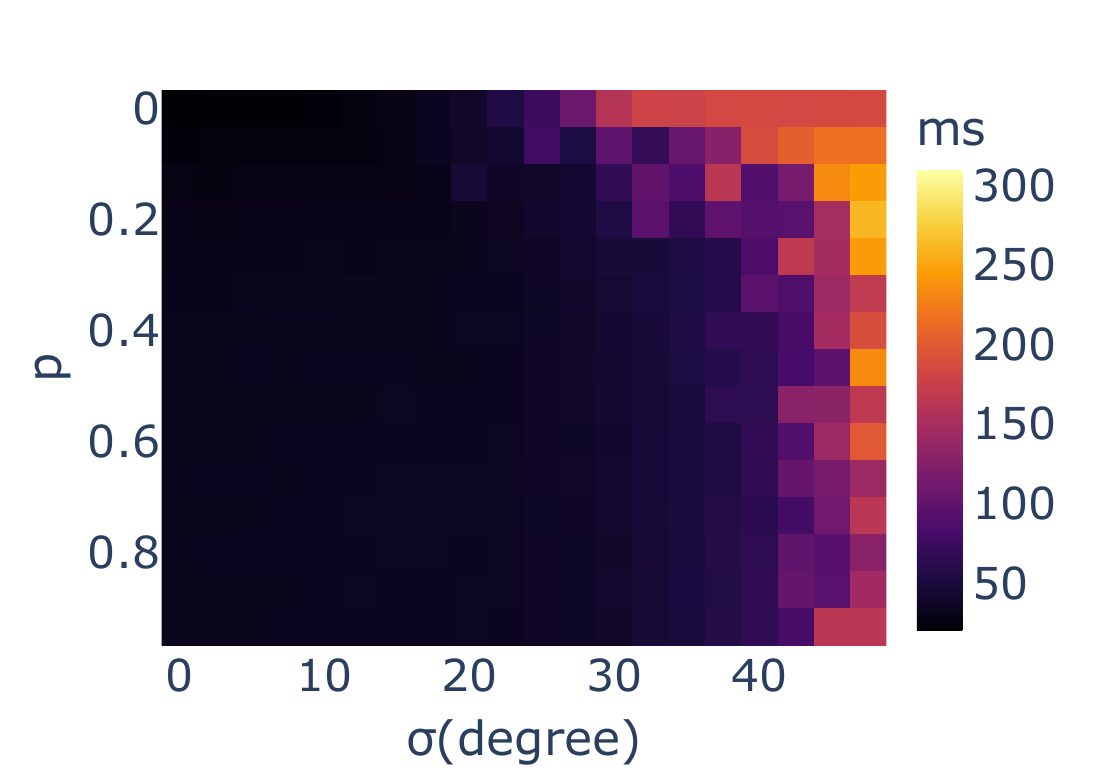} &
  \includegraphics[width=0.4\textwidth]{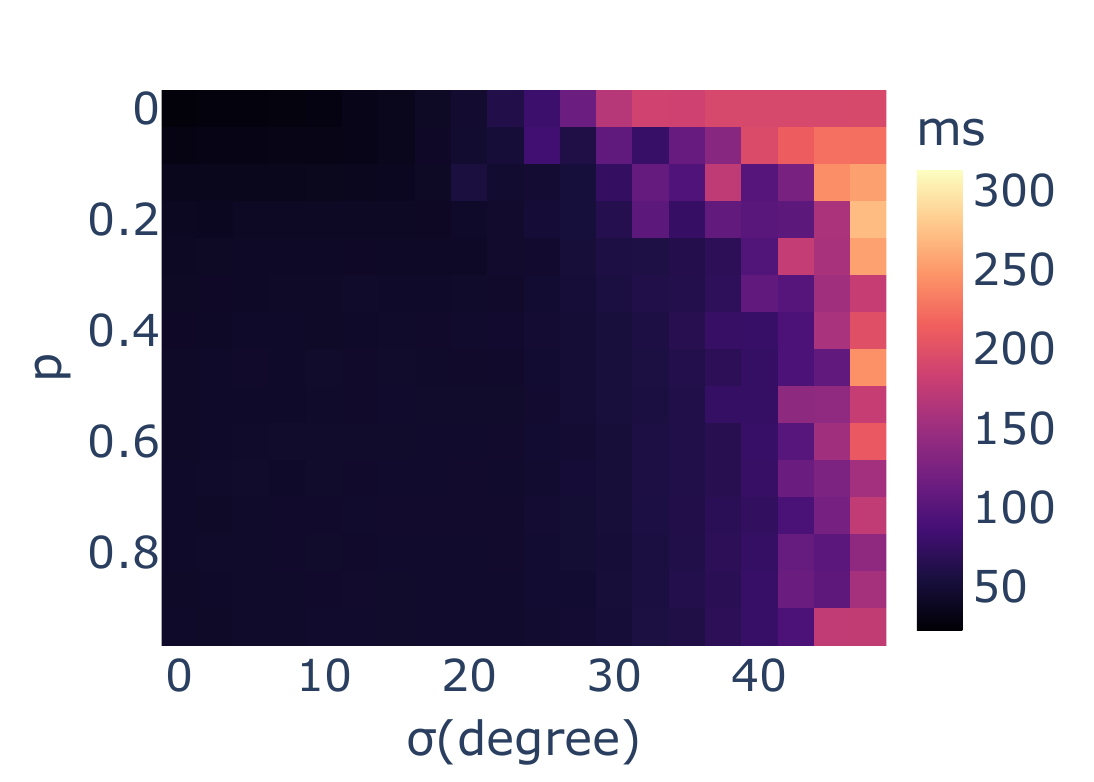} \\

\end{tabular}

  }
  \caption{Heatmap grid analysis of \SE{3}.}
  \label{fig:heatmap_se3}
\end{figure*}

\begin{figure*}
  \centering
  \vspace*{4pt}
  \setlength{\tabcolsep}{4pt}
  \resizebox{\textwidth}{!}{%
    \begin{tabular}{cccccc}
  & \textbf{Local Success rate, rtol=0.1\%} & \textbf{Performance Profile, tau=1.00} &\textbf{Initialization time} & \textbf{Local Optimization time} & \textbf{Total time} \\

  \textbf{MST} &
  \includegraphics[width=0.4\textwidth]{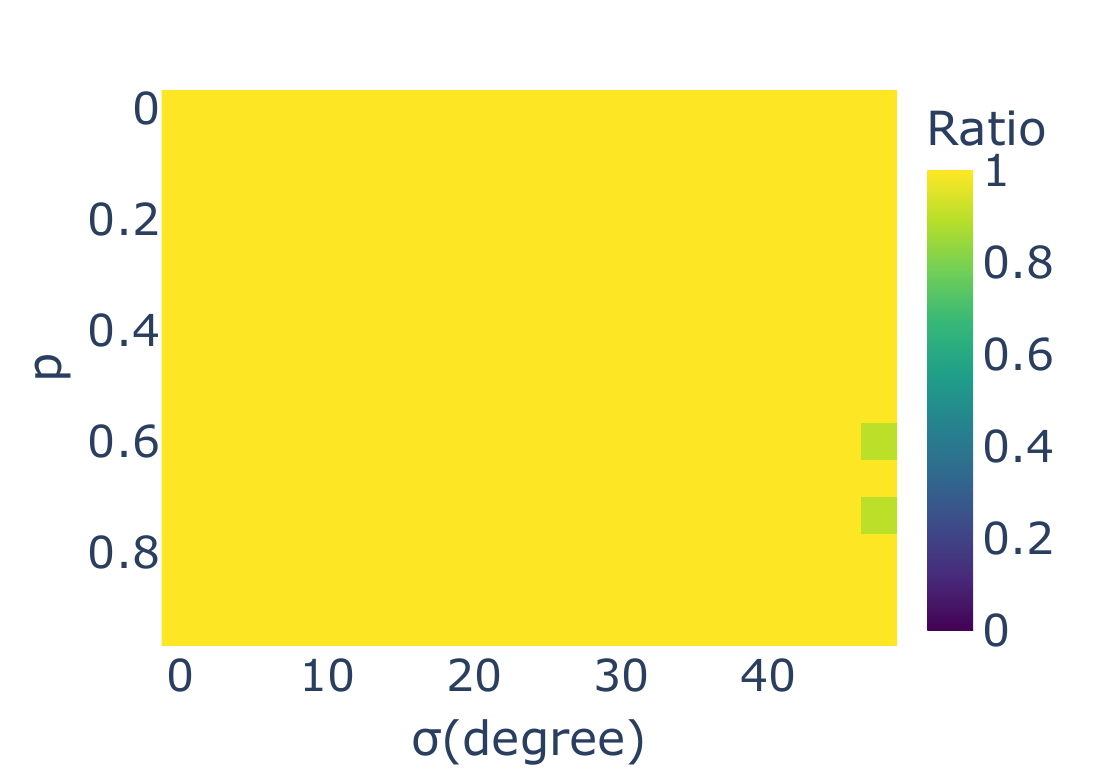} &
  \includegraphics[width=0.4\textwidth]{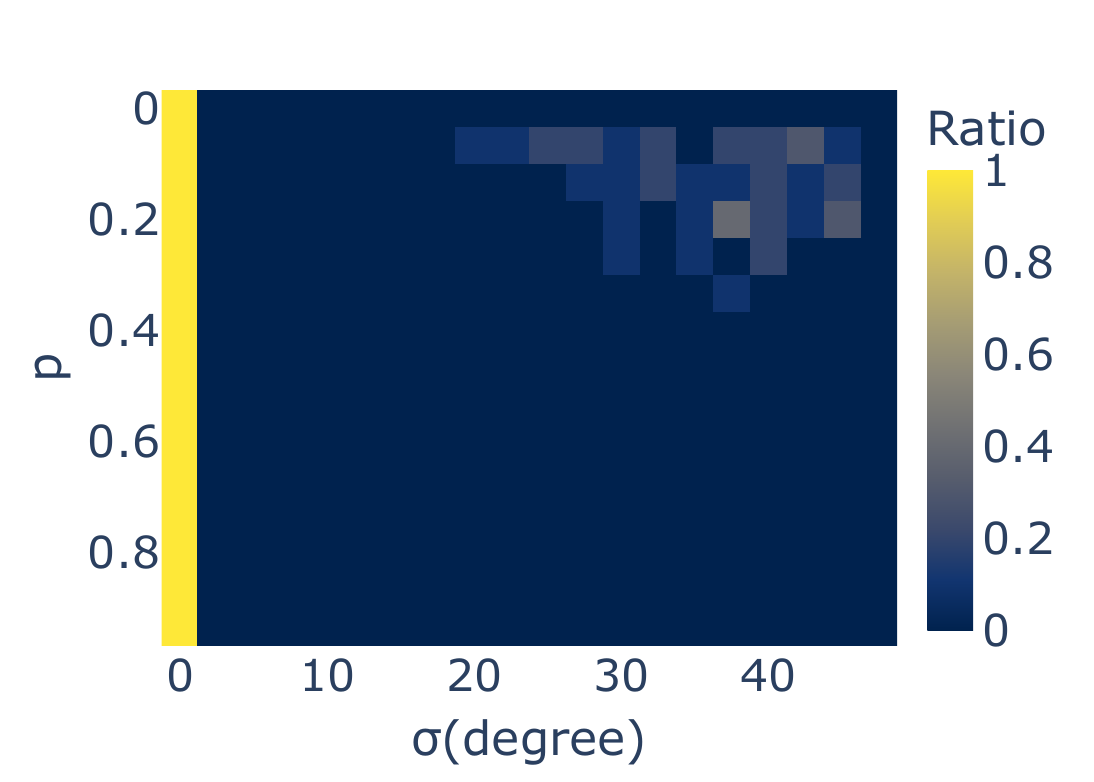} &
  \includegraphics[width=0.4\textwidth]{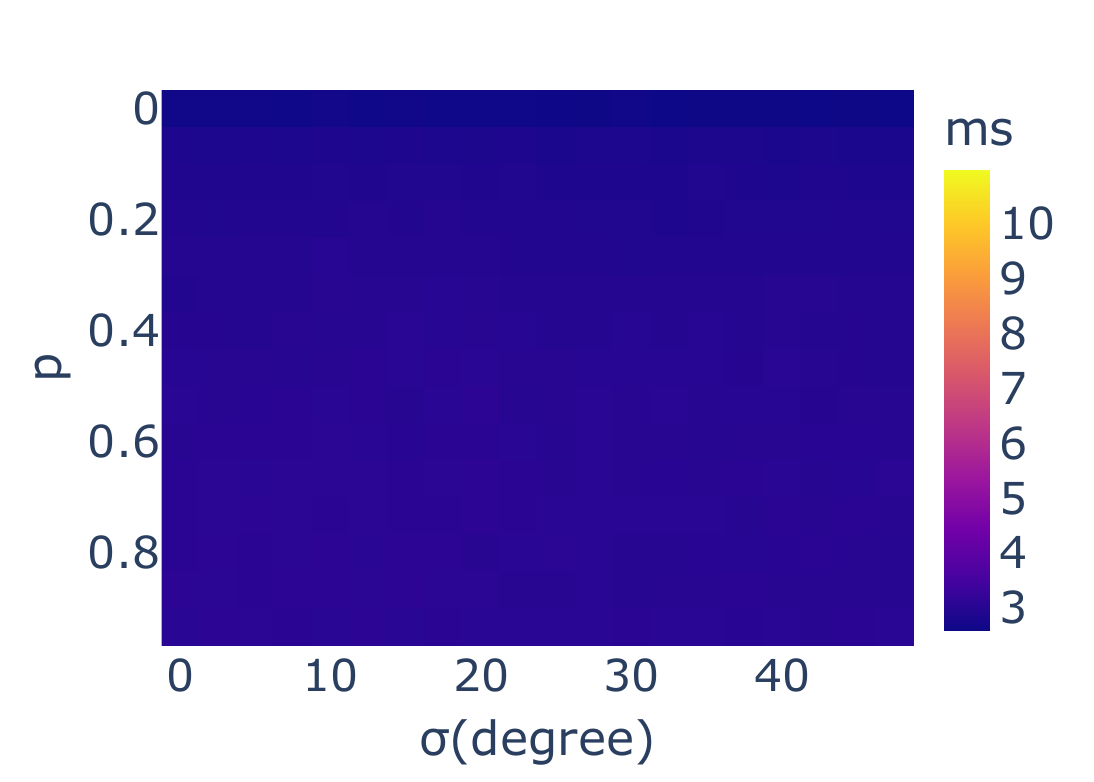} &
  \includegraphics[width=0.4\textwidth]{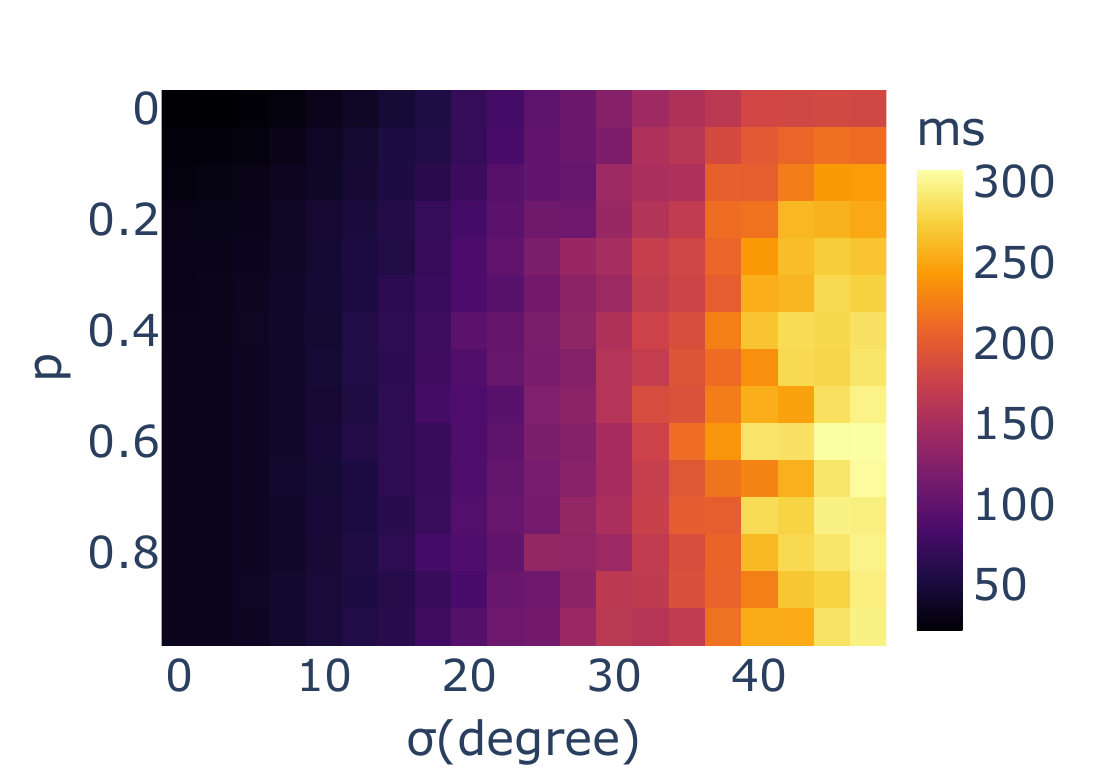} &
  \includegraphics[width=0.4\textwidth]{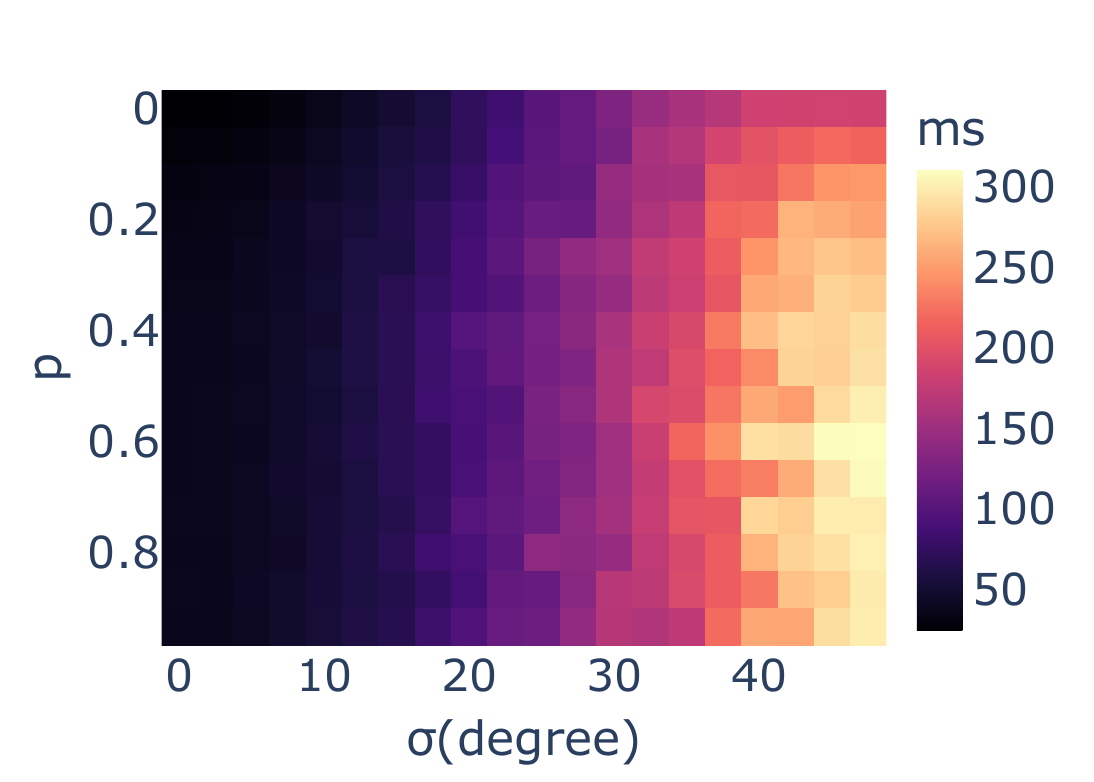} \\

  \textbf{Fast-Sync} &
  \includegraphics[width=0.4\textwidth]{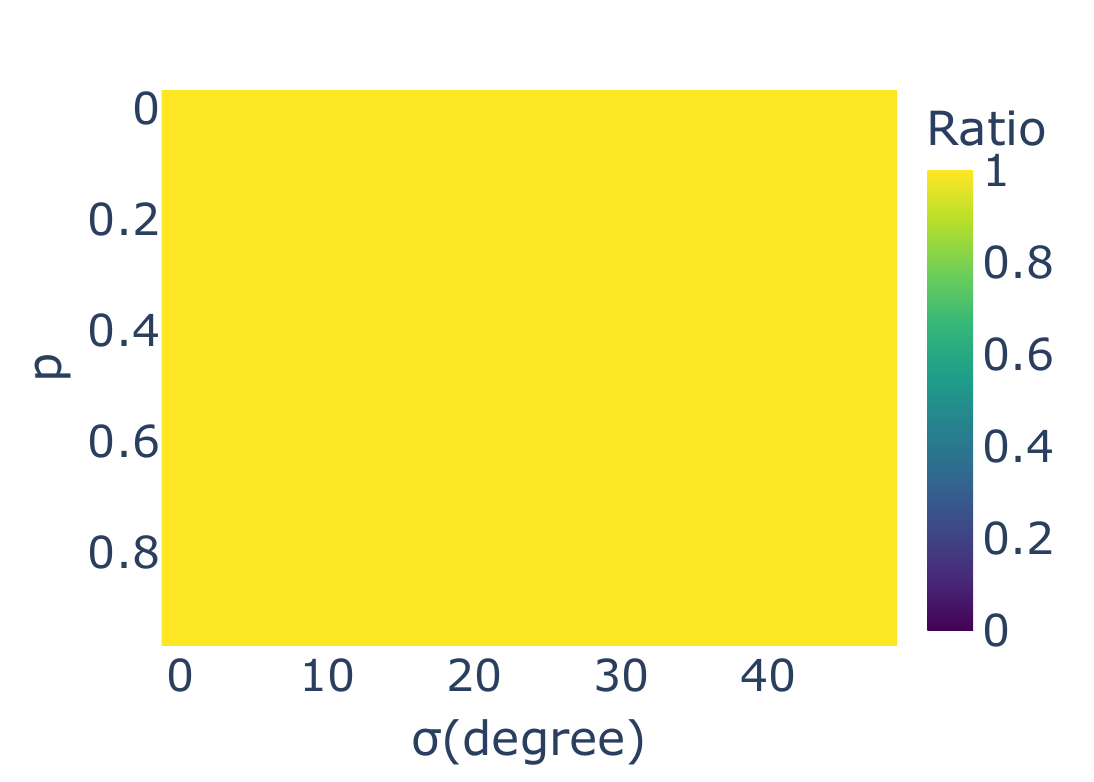} &
  \includegraphics[width=0.4\textwidth]{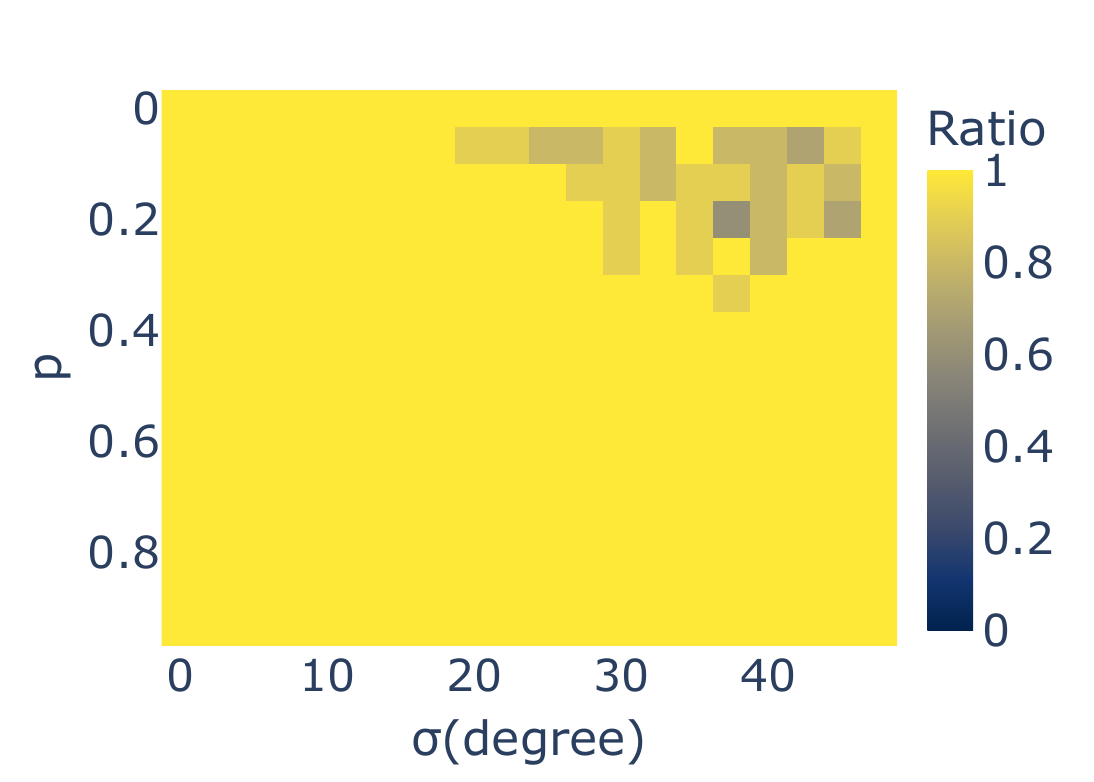} &
  \includegraphics[width=0.4\textwidth]{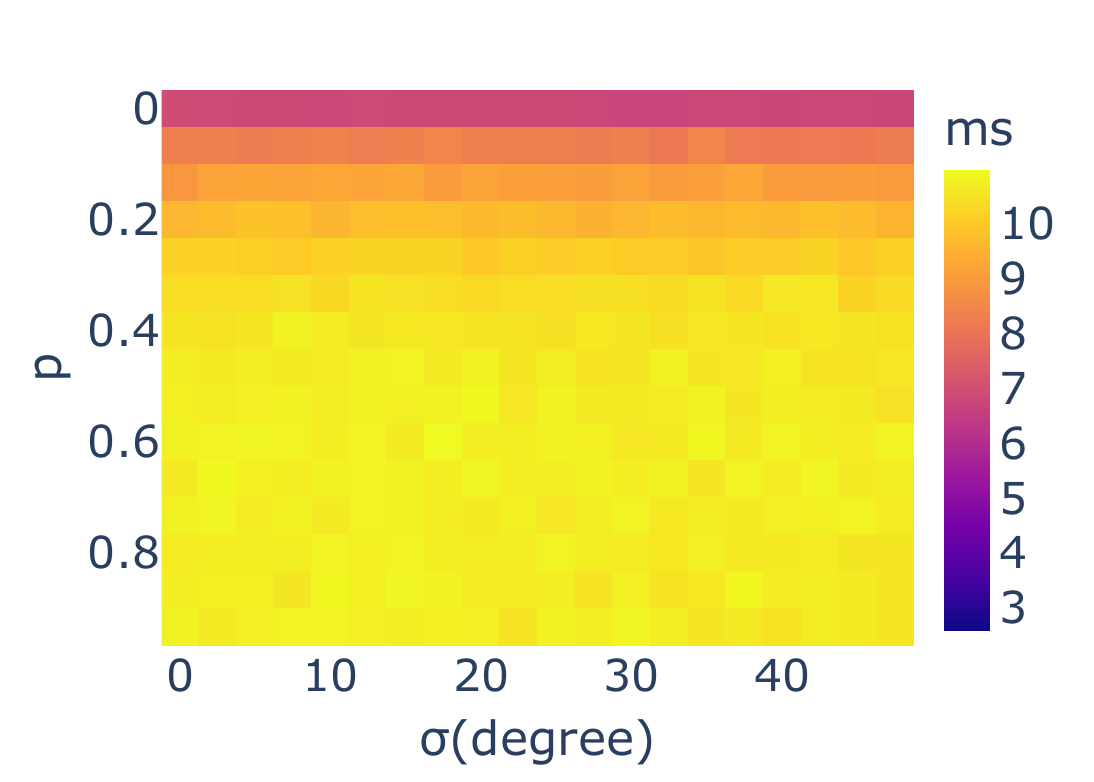} &
  \includegraphics[width=0.4\textwidth]{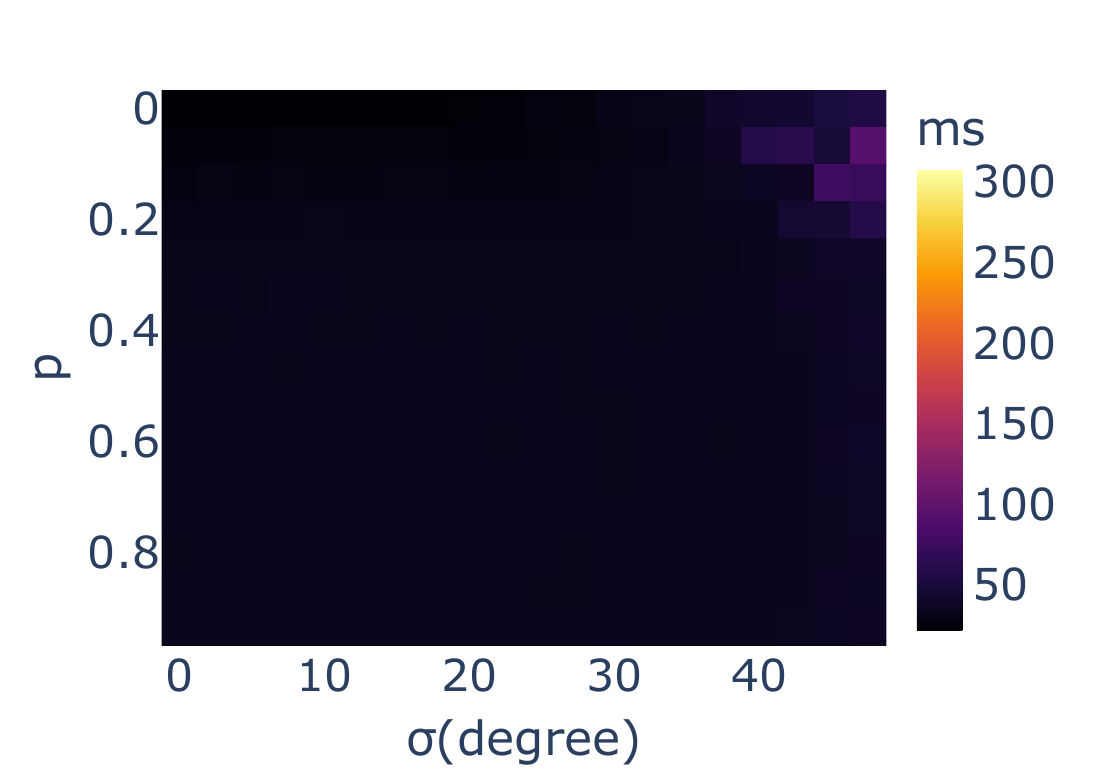} &
  \includegraphics[width=0.4\textwidth]{small_world/heatmaps/SU2_row7_LocalSCG_100N_16k.pdf} \\
\end{tabular}

  }
  \caption{Heatmap grid analysis of \SU{2}.}
  \label{fig:heatmap_su2}
\end{figure*}

\section{SYNTHETIC ANALYSIS}
\label{sec:results}

\begin{figure*}
  \centering
  \vspace*{4pt}
  \setlength{\tabcolsep}{4pt}
  \resizebox{\textwidth}{!}{%
    \begin{tabular}{cccccc}
& \textbf{Local Success rate, rtol=0.1\%} & \textbf{Performance Profile, tau=1.00} &\textbf{Initialization time} & \textbf{Local Optimization time} & \textbf{Total time} \\

  \textbf{MST} &
  \includegraphics[width=0.4\textwidth]{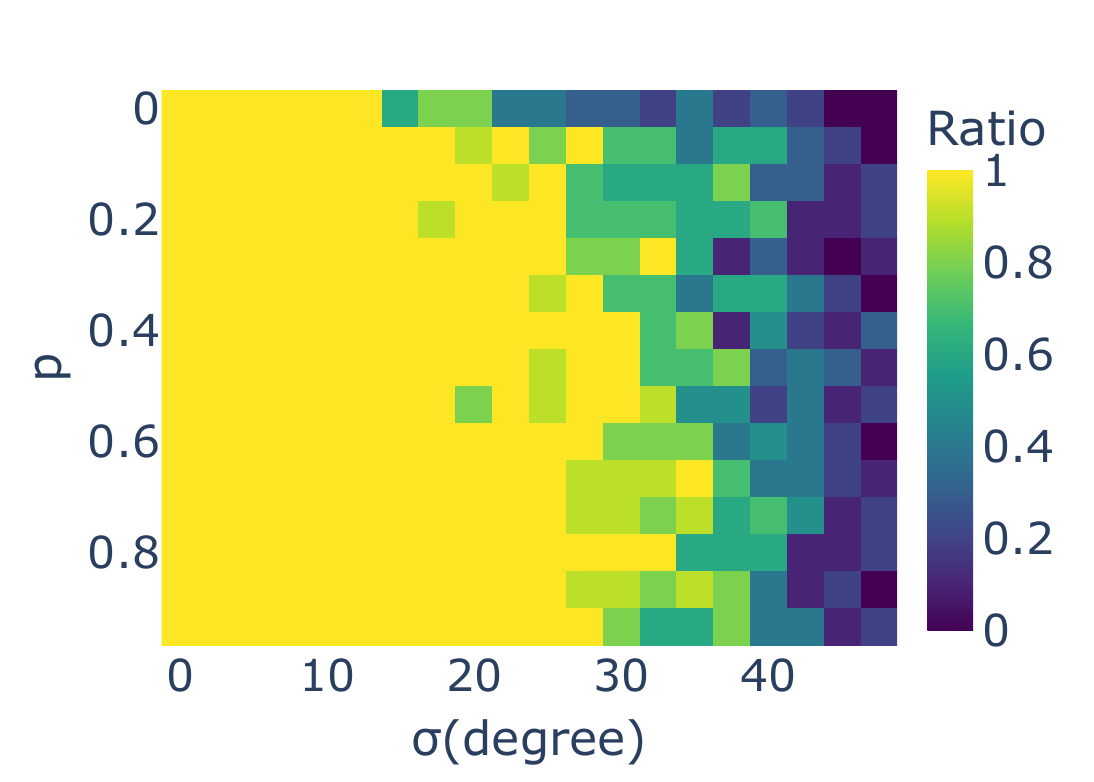} &
  \includegraphics[width=0.4\textwidth]{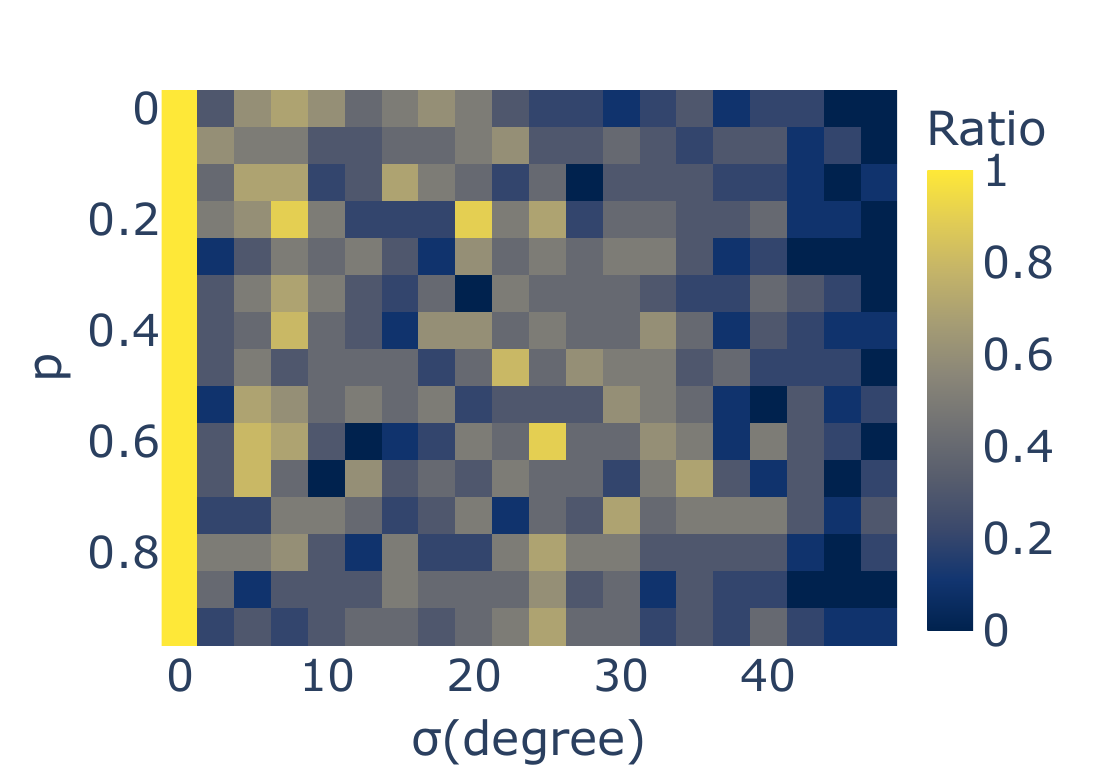} &
  \includegraphics[width=0.4\textwidth]{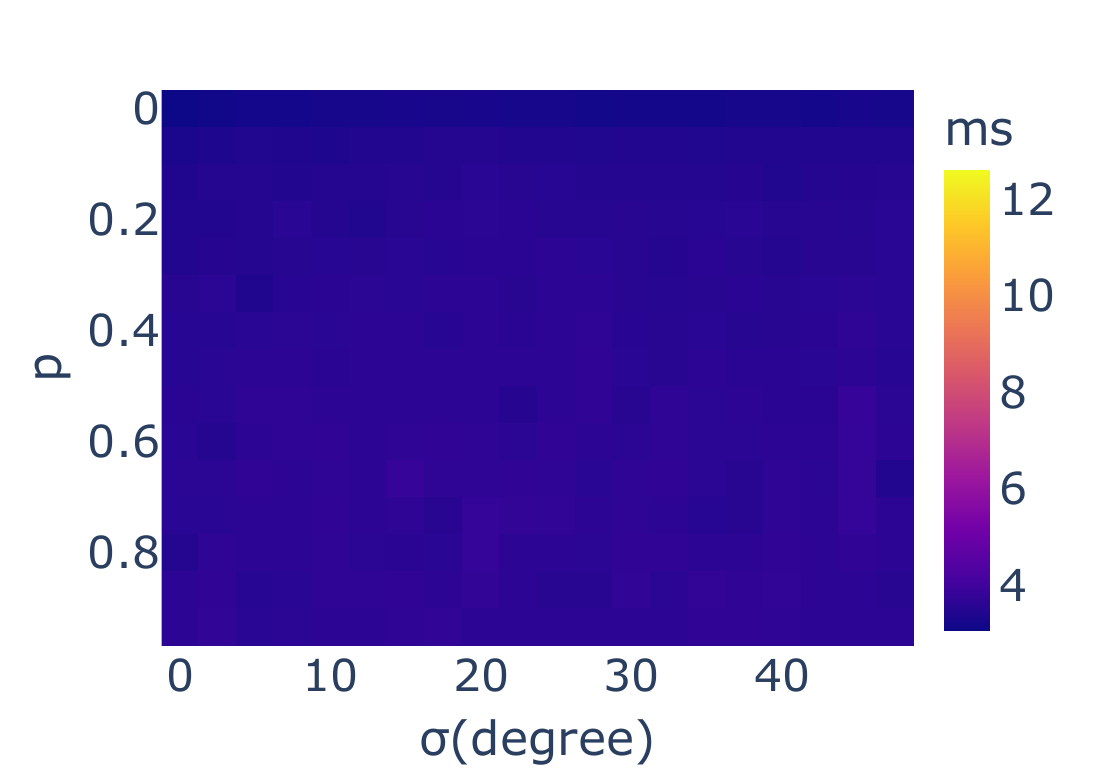} &
  \includegraphics[width=0.4\textwidth]{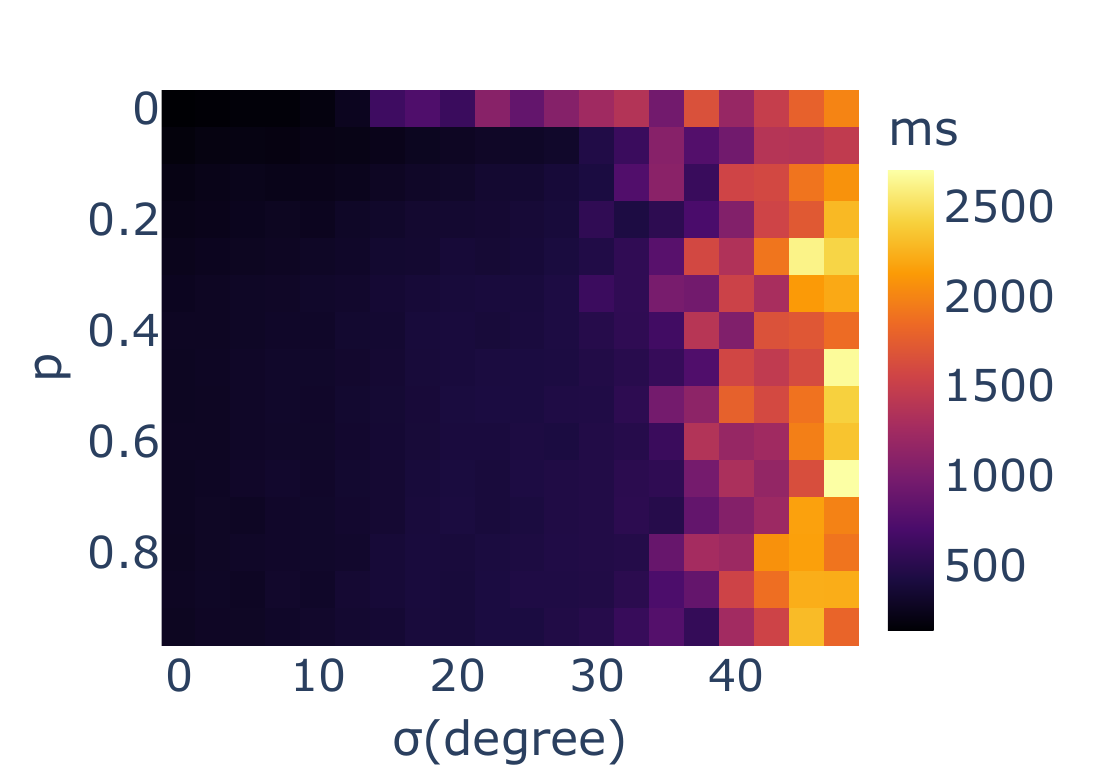} &
  \includegraphics[width=0.4\textwidth]{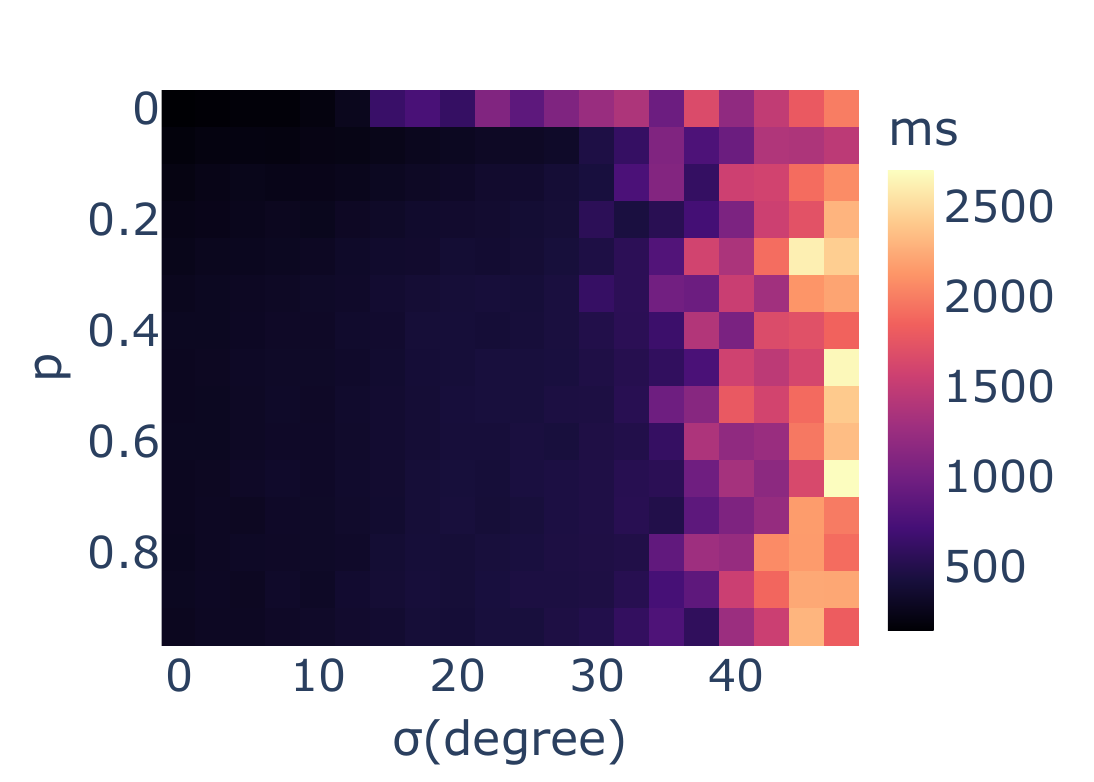} \\

  \textbf{Fast-Sync} &
  \includegraphics[width=0.4\textwidth]{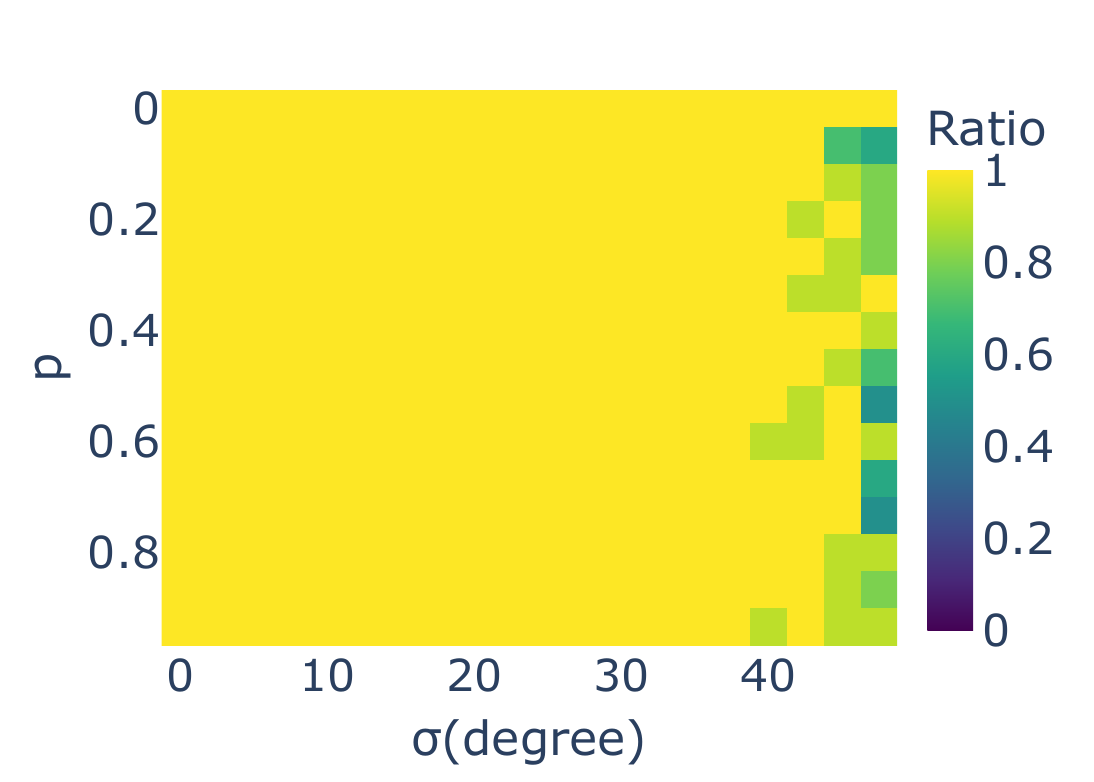} &
  \includegraphics[width=0.4\textwidth]{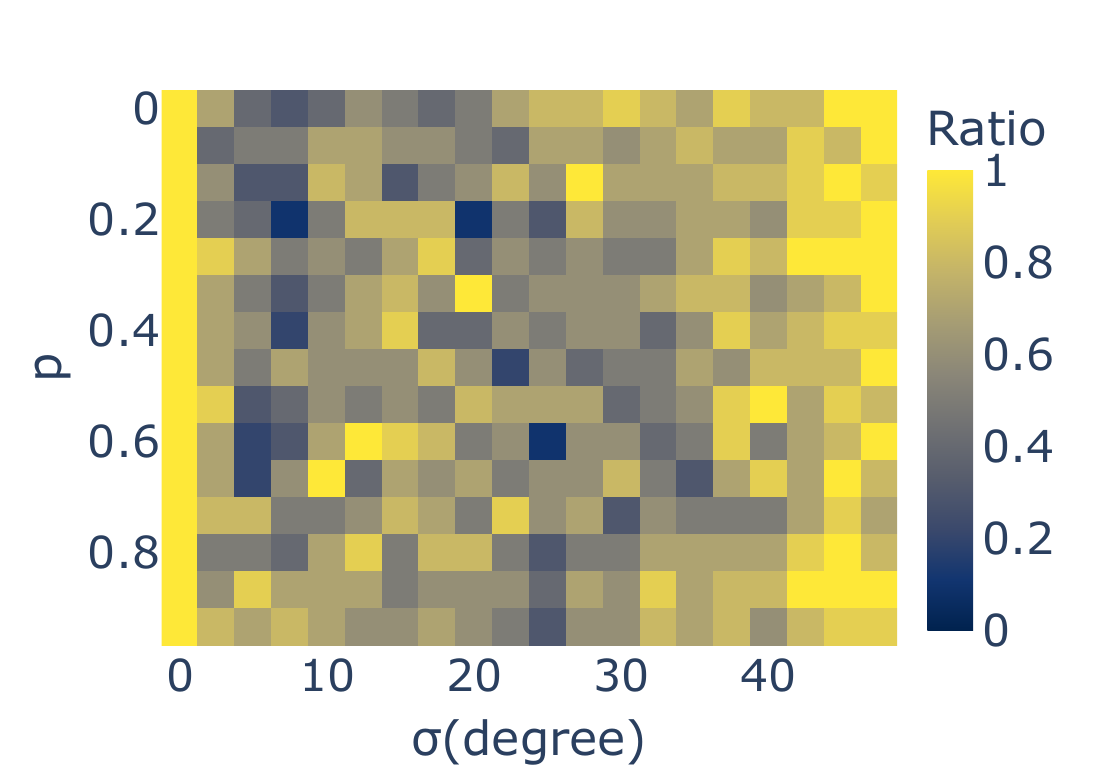} &
  \includegraphics[width=0.4\textwidth]{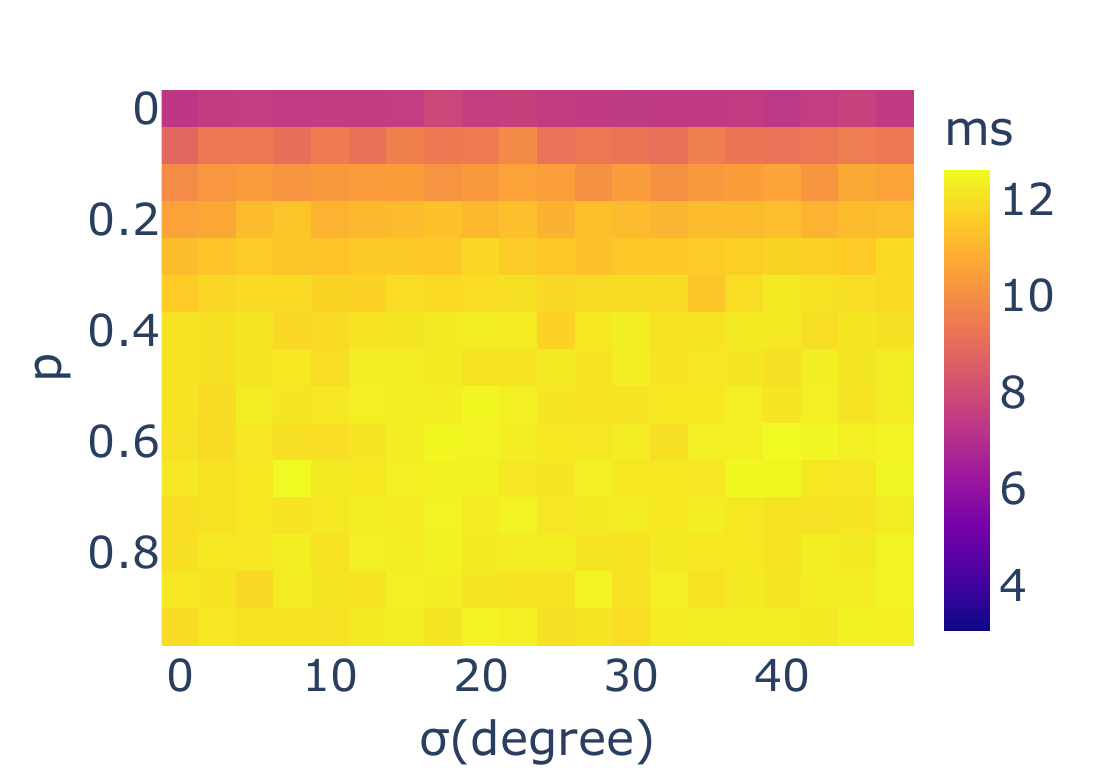} &
  \includegraphics[width=0.4\textwidth]{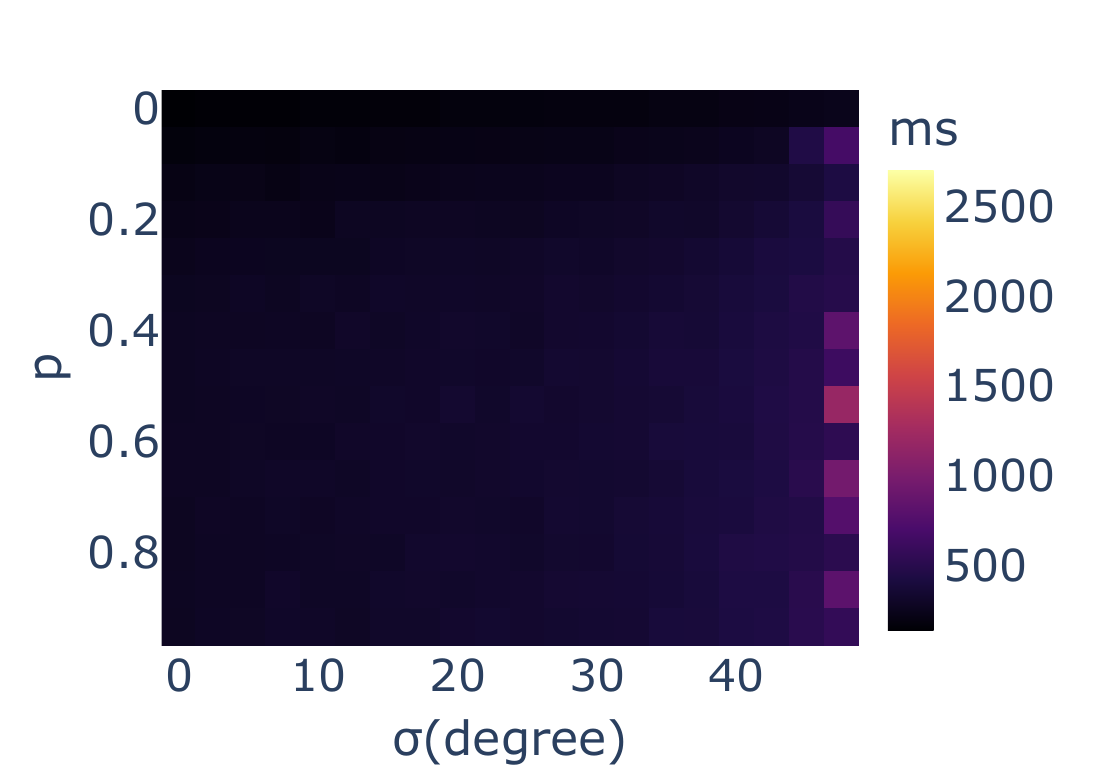} &
  \includegraphics[width=0.4\textwidth]{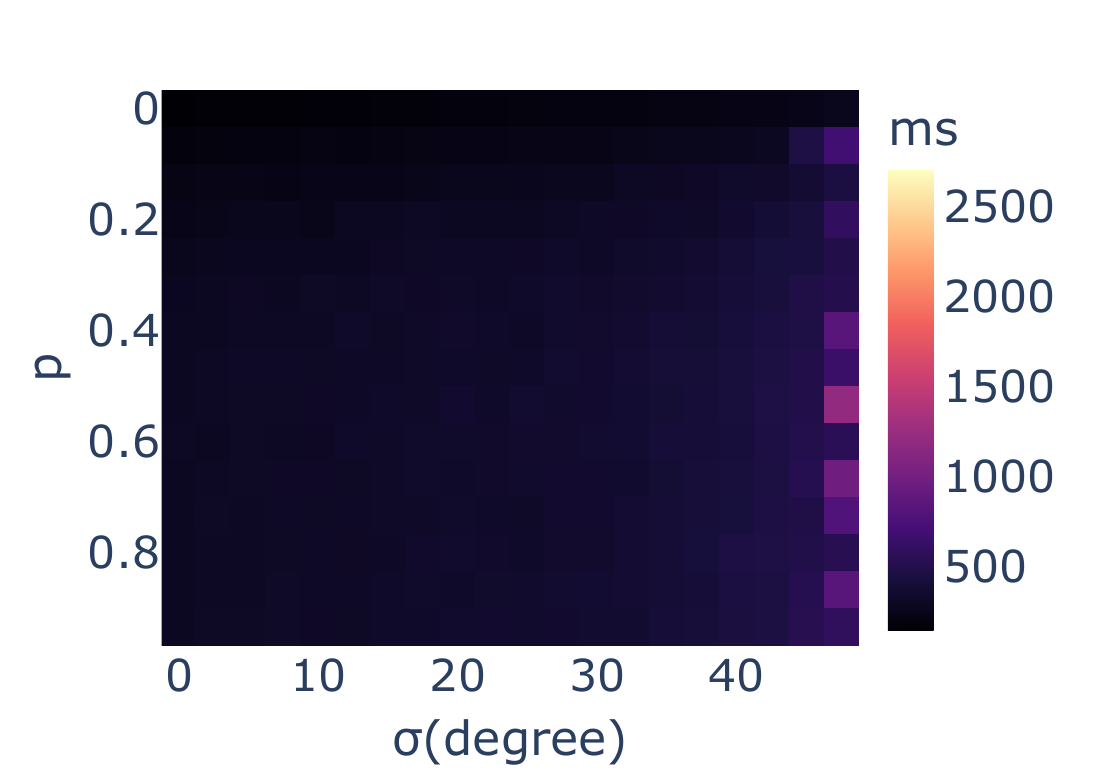} \\
\end{tabular}

  }
  \caption{Heatmap grid analysis of \SL{4}.}
  \label{fig:heatmap_sl4}
\end{figure*}

\begin{figure*}
  \centering
  \vspace*{4pt}
  \setlength{\tabcolsep}{4pt}
  \resizebox{\textwidth}{!}{%
    \begin{tabular}{cccccc}
& \textbf{Local Success rate, rtol=0.1\%} & \textbf{Performance Profile, tau=1.00} &\textbf{Initialization time} & \textbf{Local Optimization time} & \textbf{Total time} \\

  \textbf{MST} &
  \includegraphics[width=0.4\textwidth]{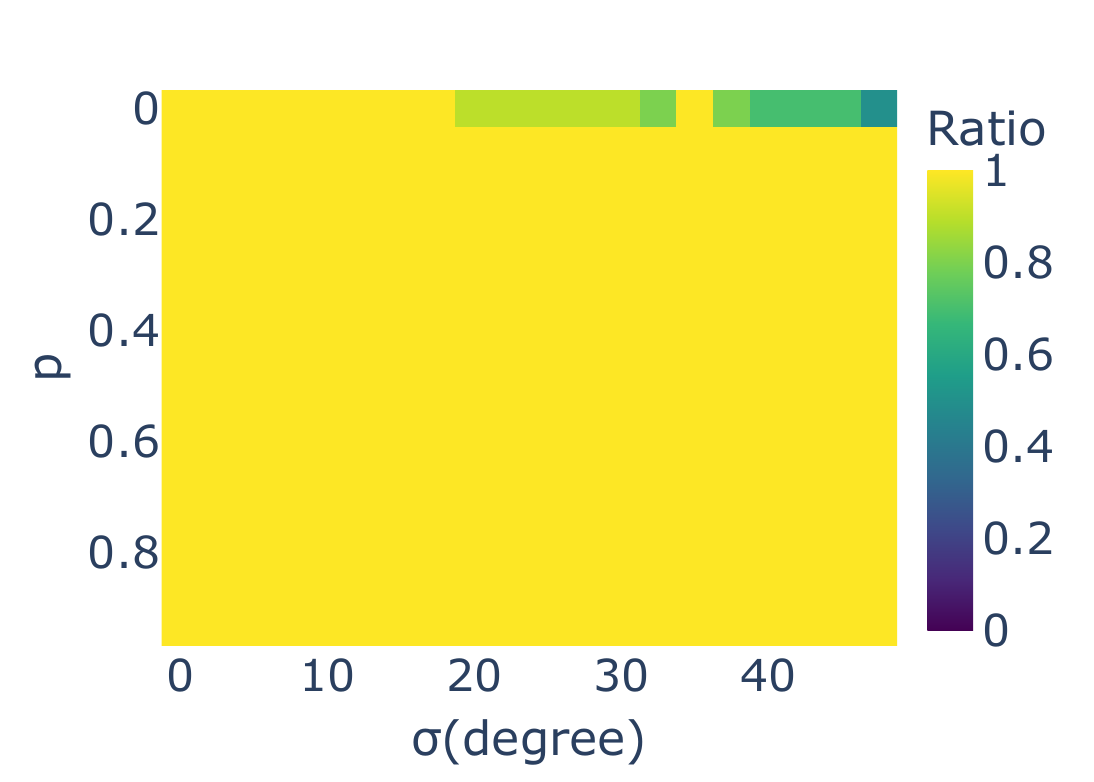} &
  \includegraphics[width=0.4\textwidth]{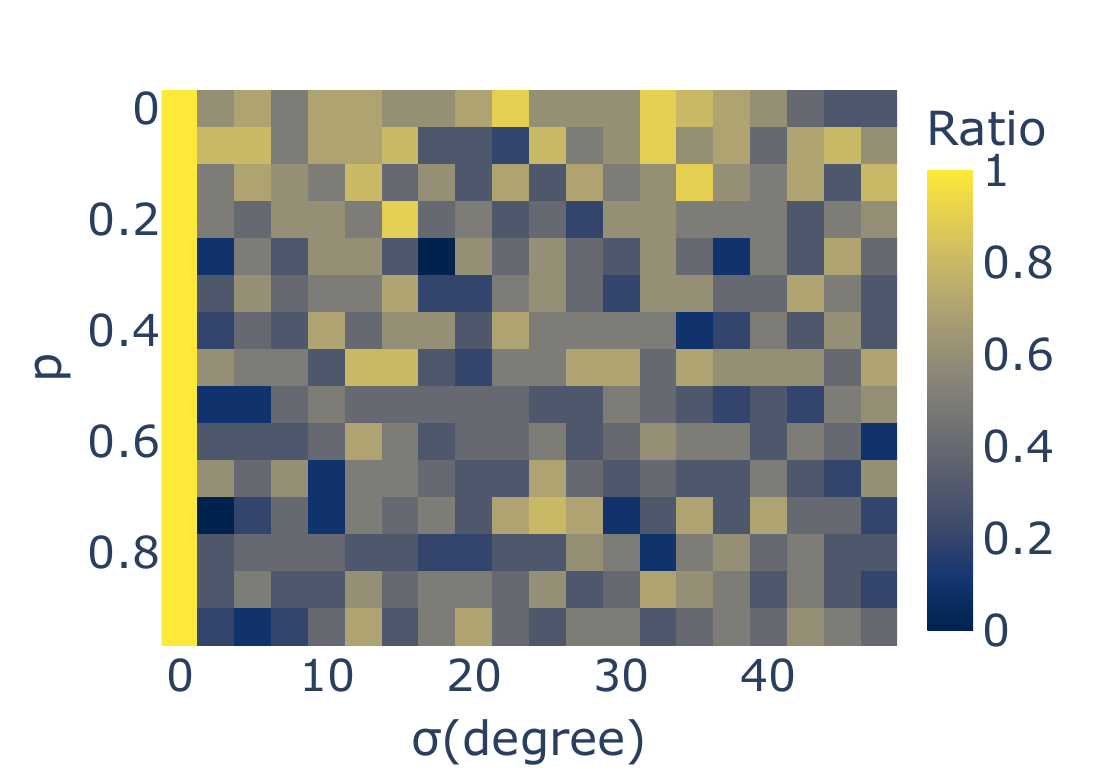} &
  \includegraphics[width=0.4\textwidth]{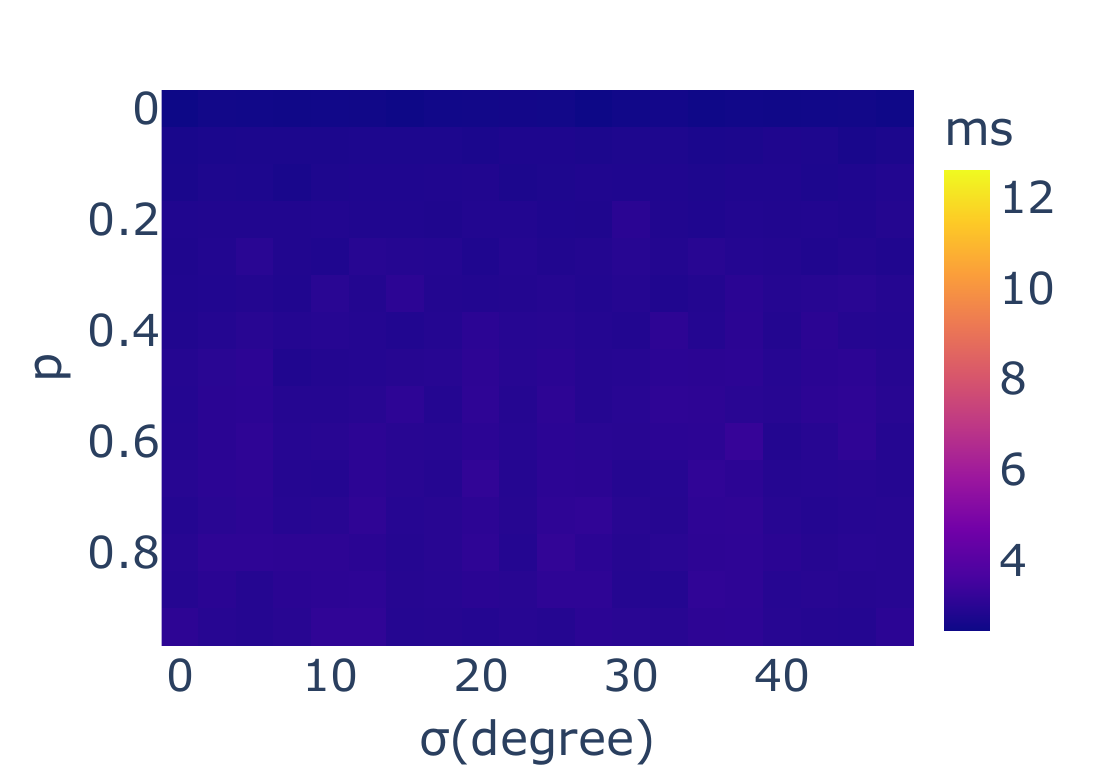} &
  \includegraphics[width=0.4\textwidth]{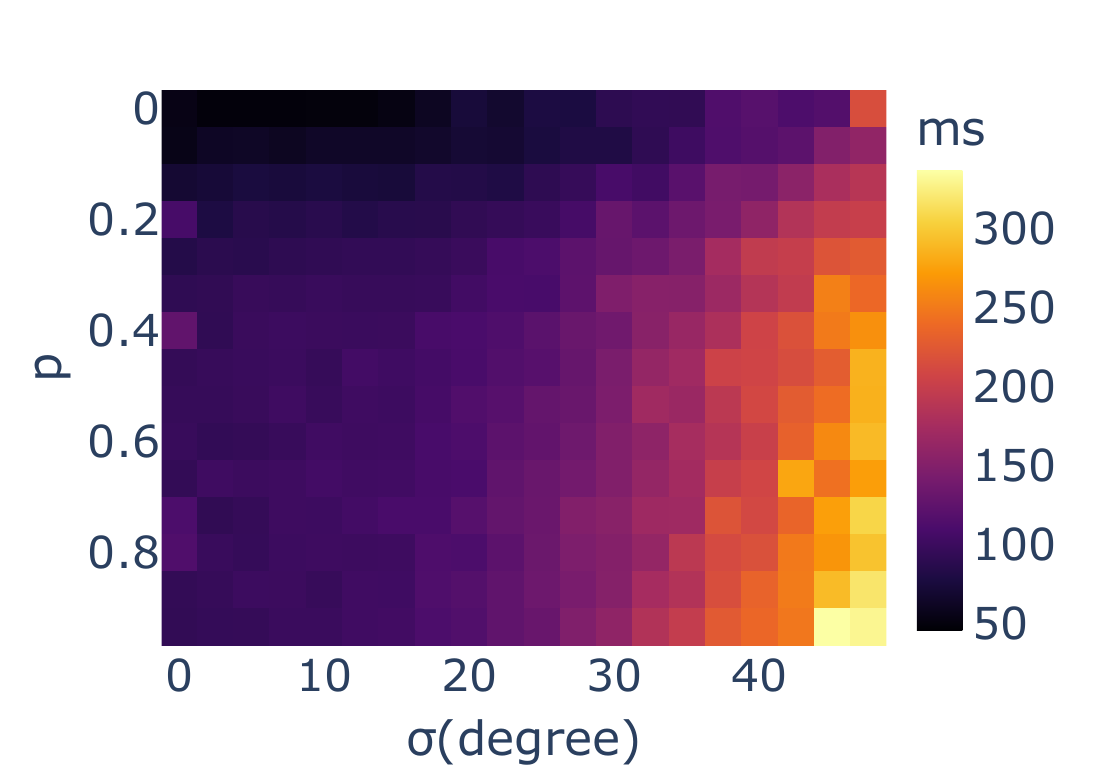} &
  \includegraphics[width=0.4\textwidth]{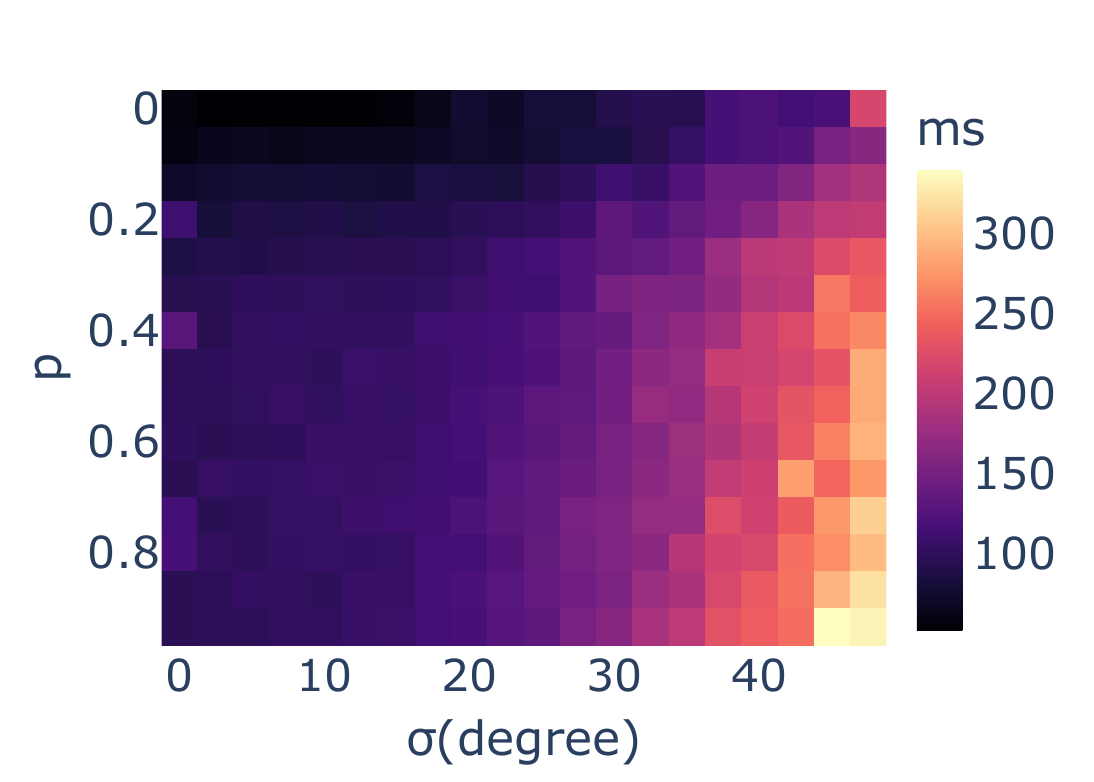} \\

  \textbf{Fast-Sync} &
  \includegraphics[width=0.4\textwidth]{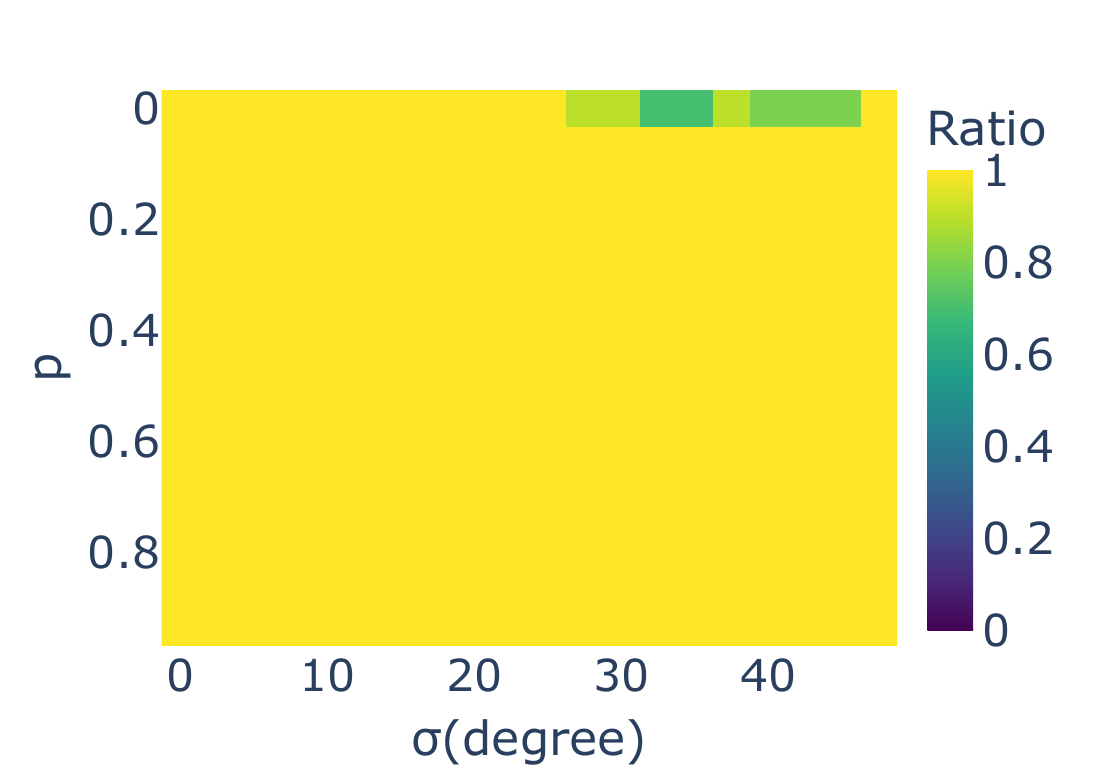} &
  \includegraphics[width=0.4\textwidth]{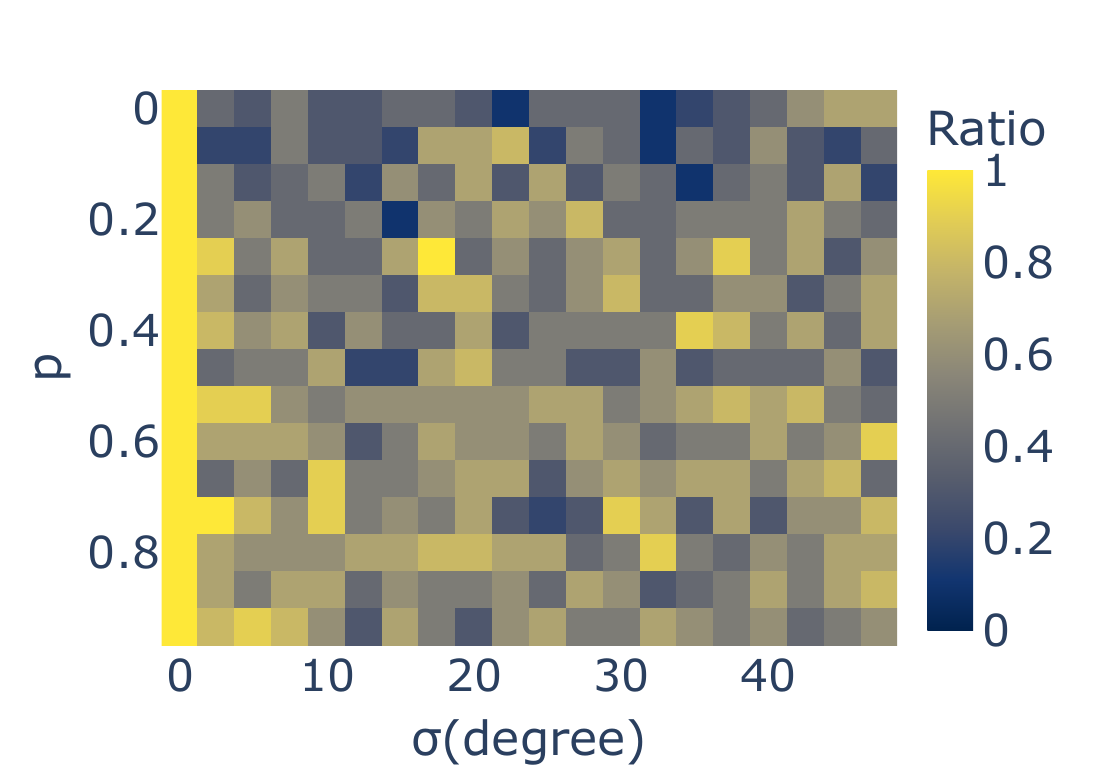} &
  \includegraphics[width=0.4\textwidth]{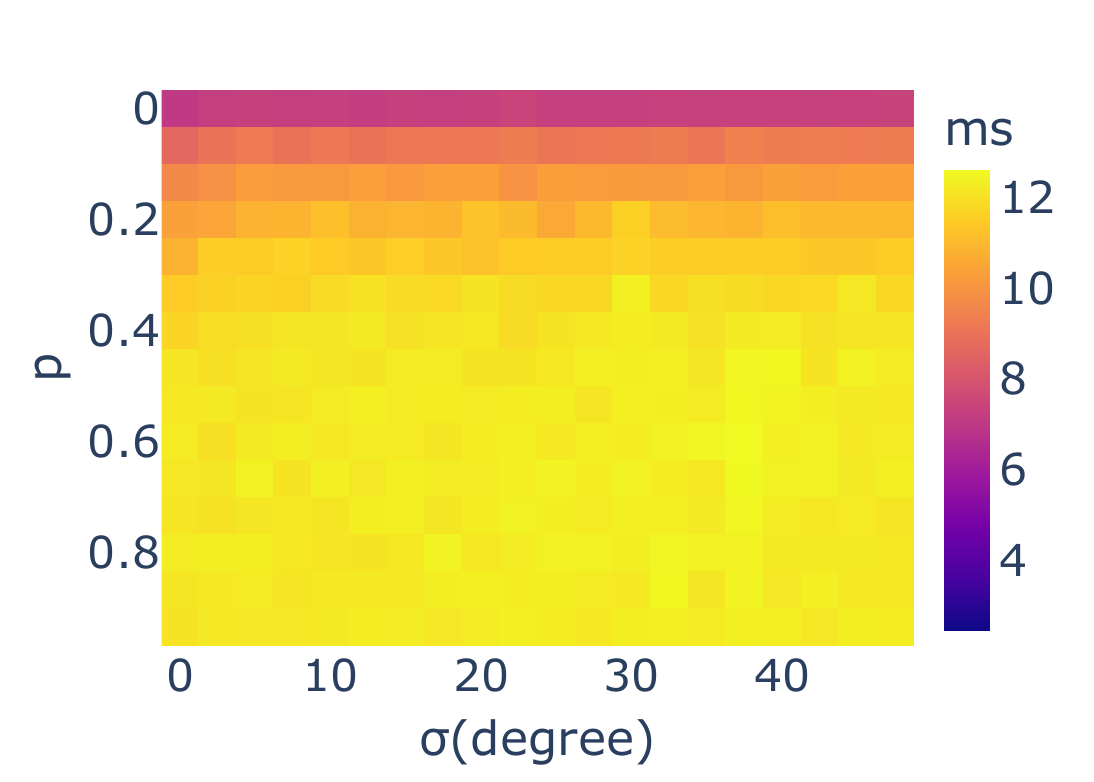} &
  \includegraphics[width=0.4\textwidth]{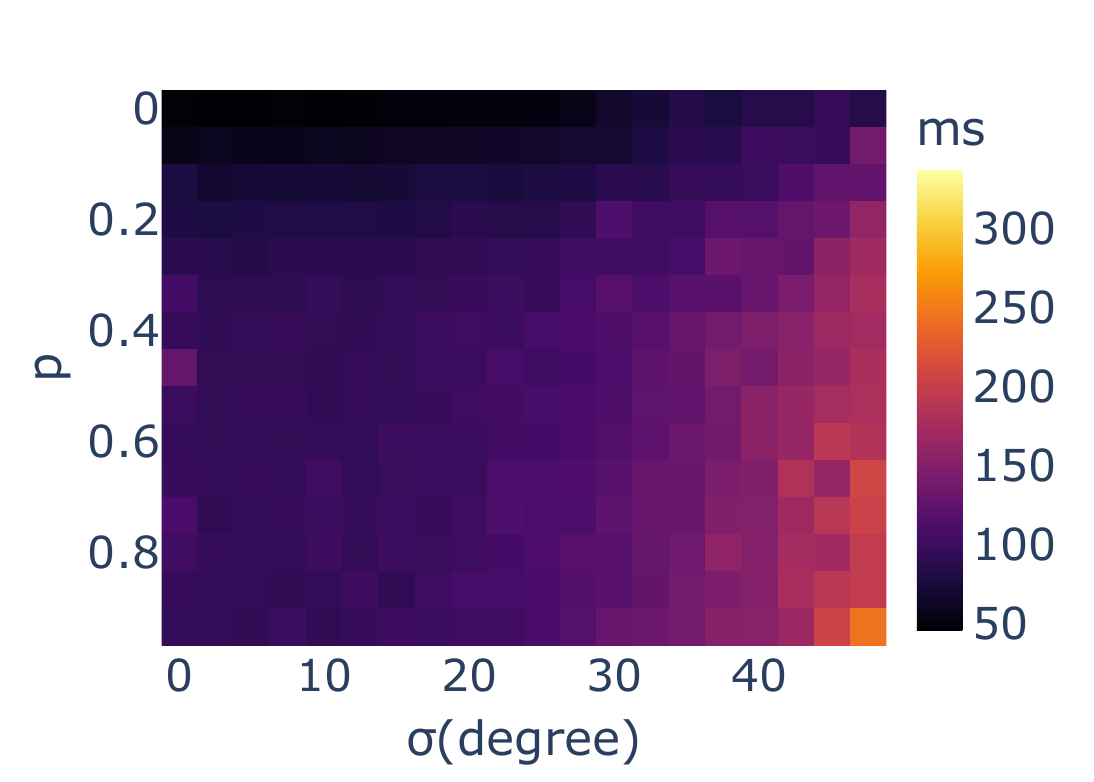} &
  \includegraphics[width=0.4\textwidth]{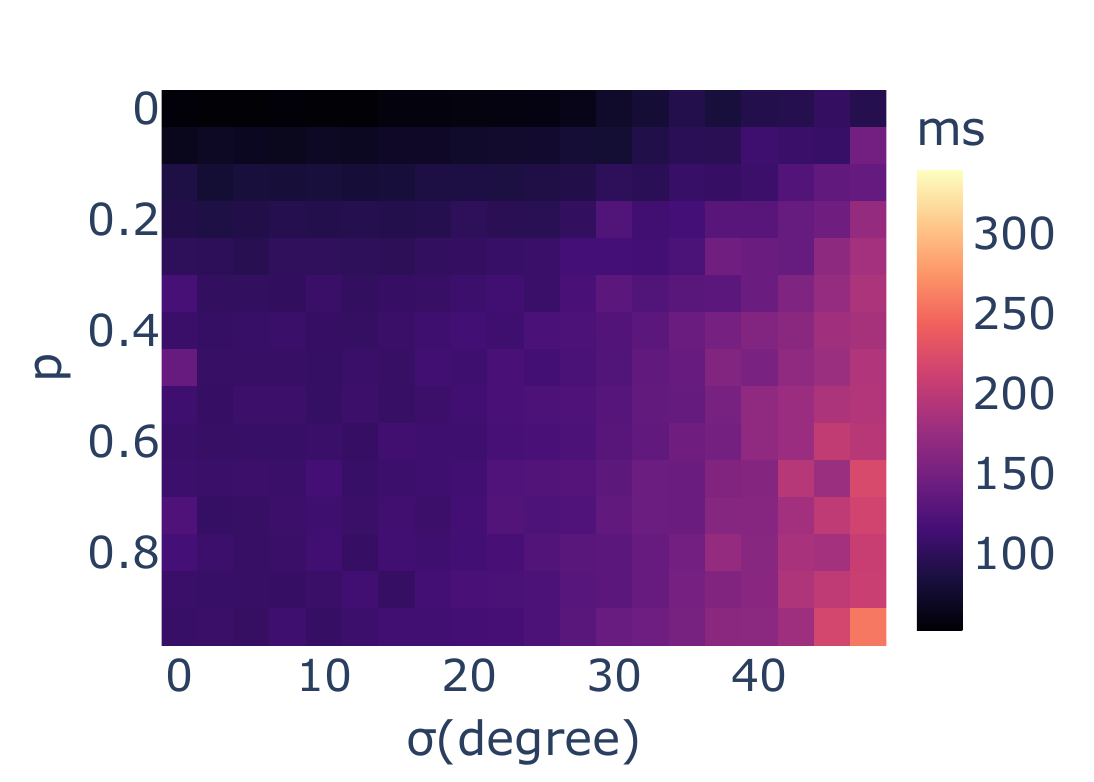} \\
\end{tabular}

  }
  \caption{Heatmap grid analysis of \Gal{3}.}
  \label{fig:heatmap_gal3}
\end{figure*}

\begin{figure*}
  \centering
  \vspace*{4pt}
  \setlength{\tabcolsep}{4pt}
  \resizebox{\textwidth}{!}{%
    \begin{tabular}{cccccc}
& \textbf{Local Success rate, rtol=0.1\%} & \textbf{Performance Profile, tau=1.00} &\textbf{Initialization time} & \textbf{Local Optimization time} & \textbf{Total time} \\

  \textbf{MST} &
  \includegraphics[width=0.4\textwidth]{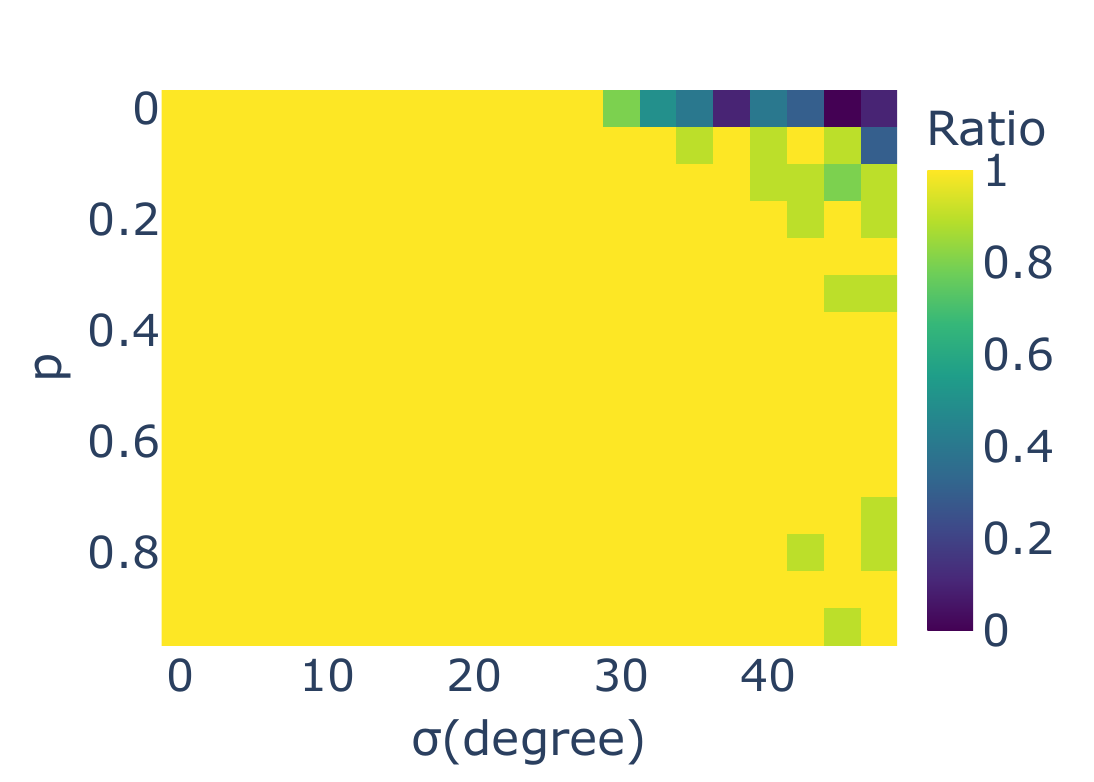} &
  \includegraphics[width=0.4\textwidth]{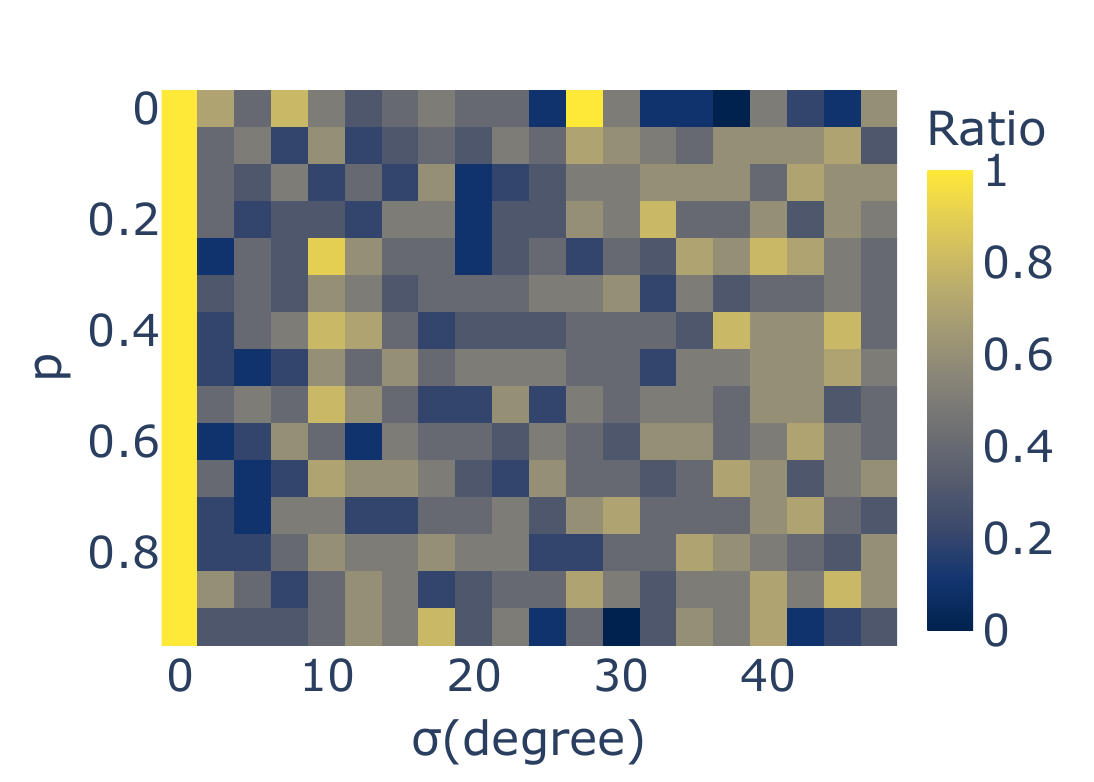} &
  \includegraphics[width=0.4\textwidth]{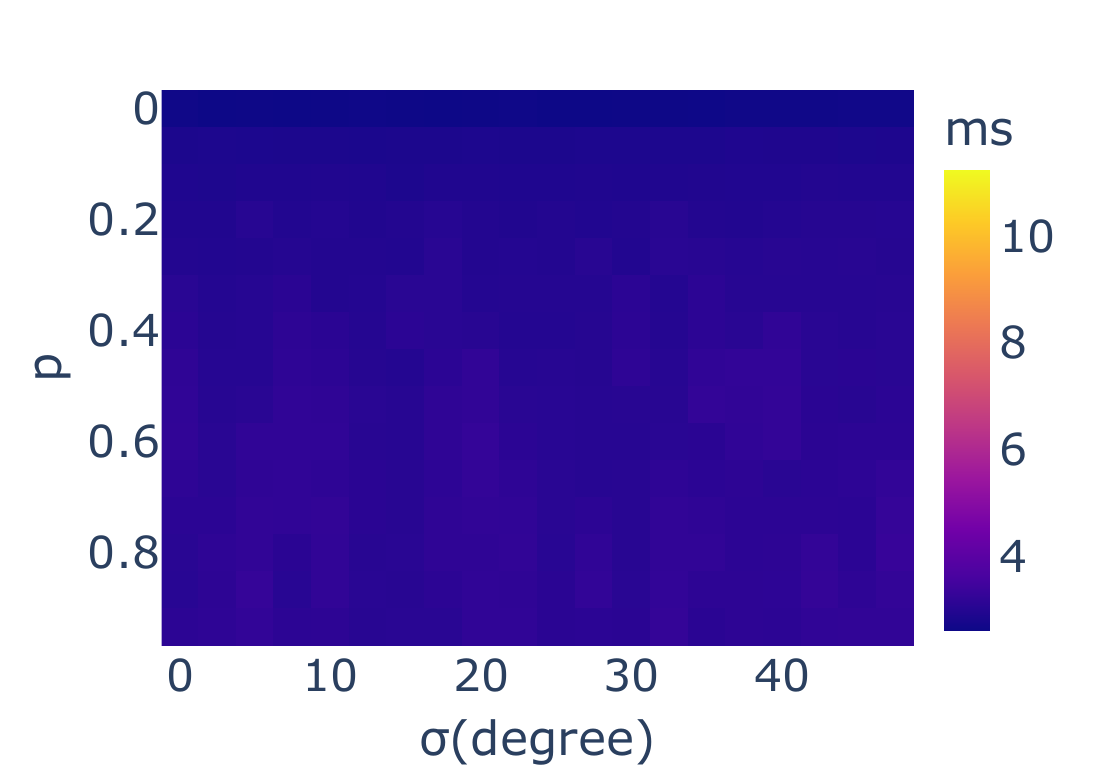} &
  \includegraphics[width=0.4\textwidth]{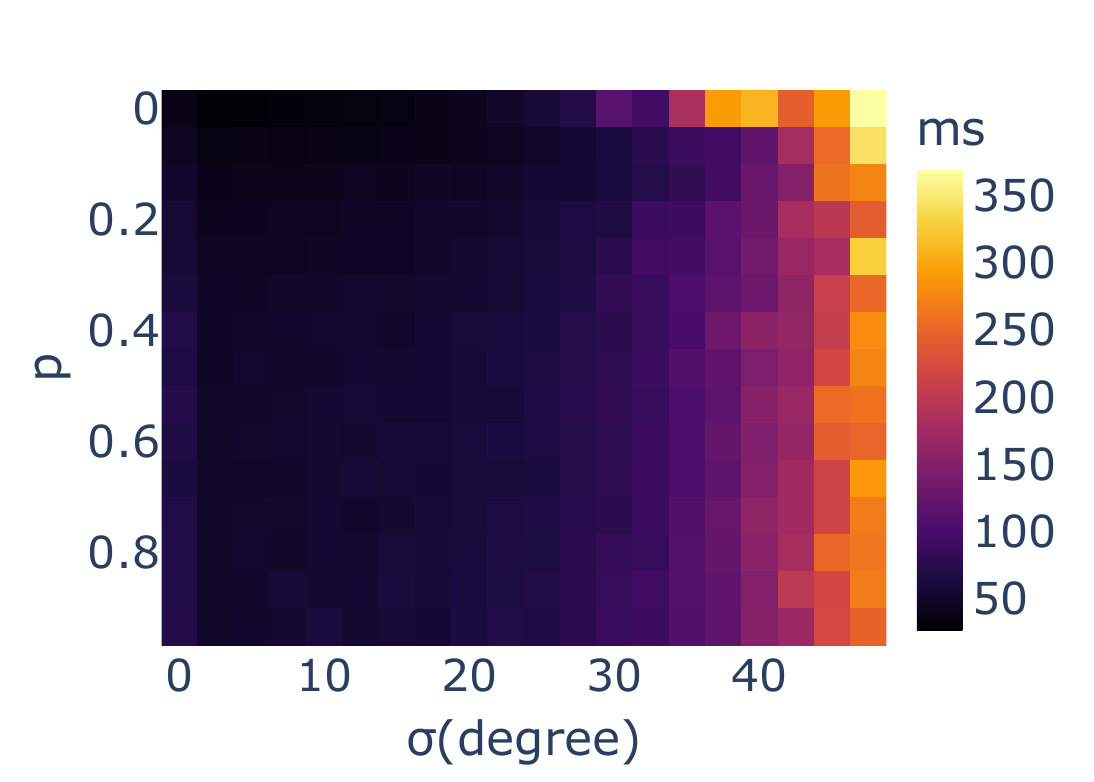} &
  \includegraphics[width=0.4\textwidth]{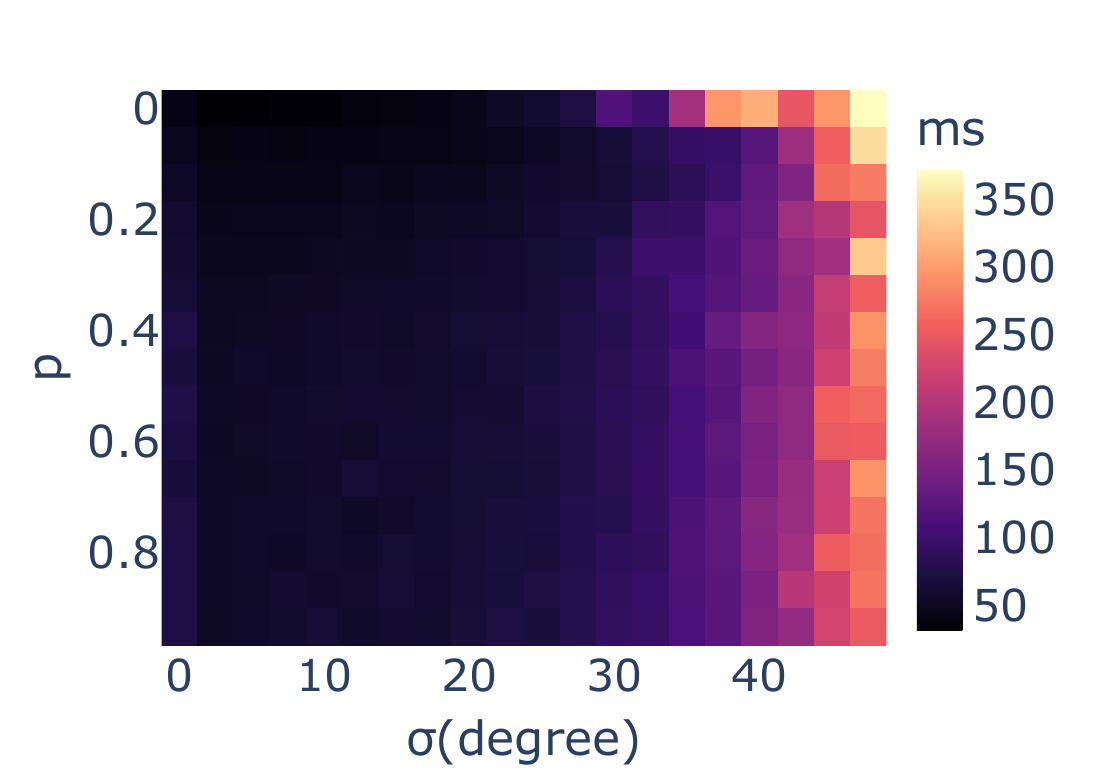} \\

  \textbf{Fast-Sync} &
  \includegraphics[width=0.4\textwidth]{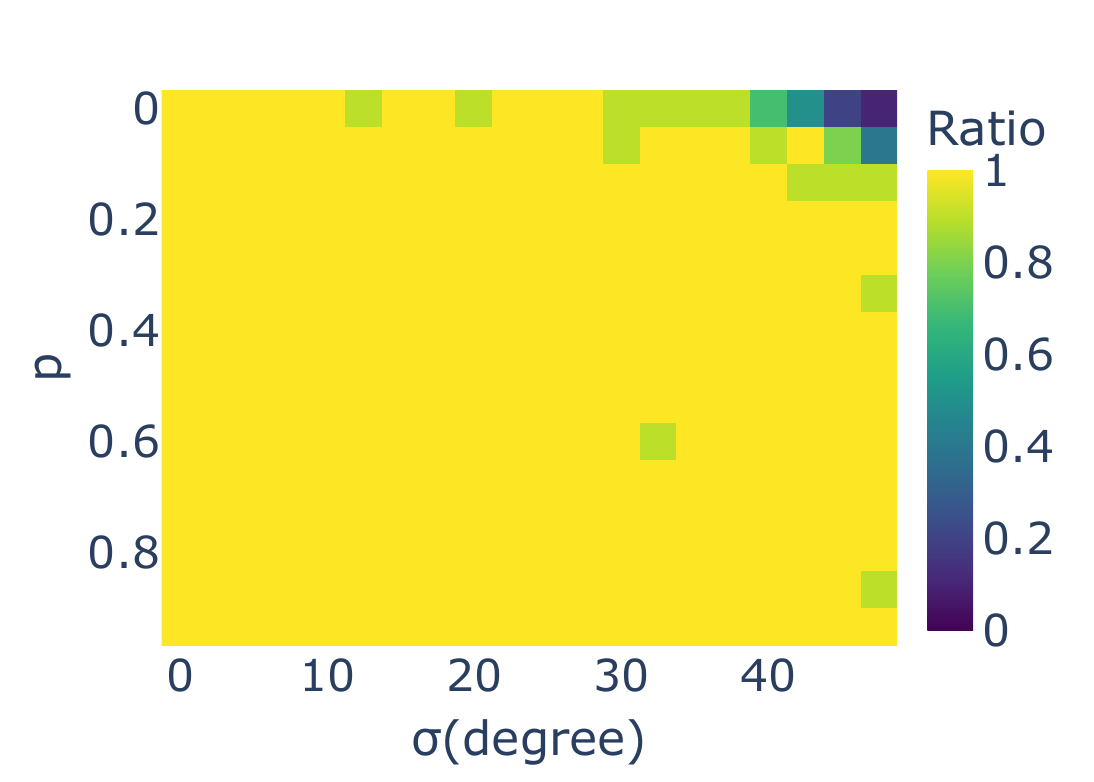} &
  \includegraphics[width=0.4\textwidth]{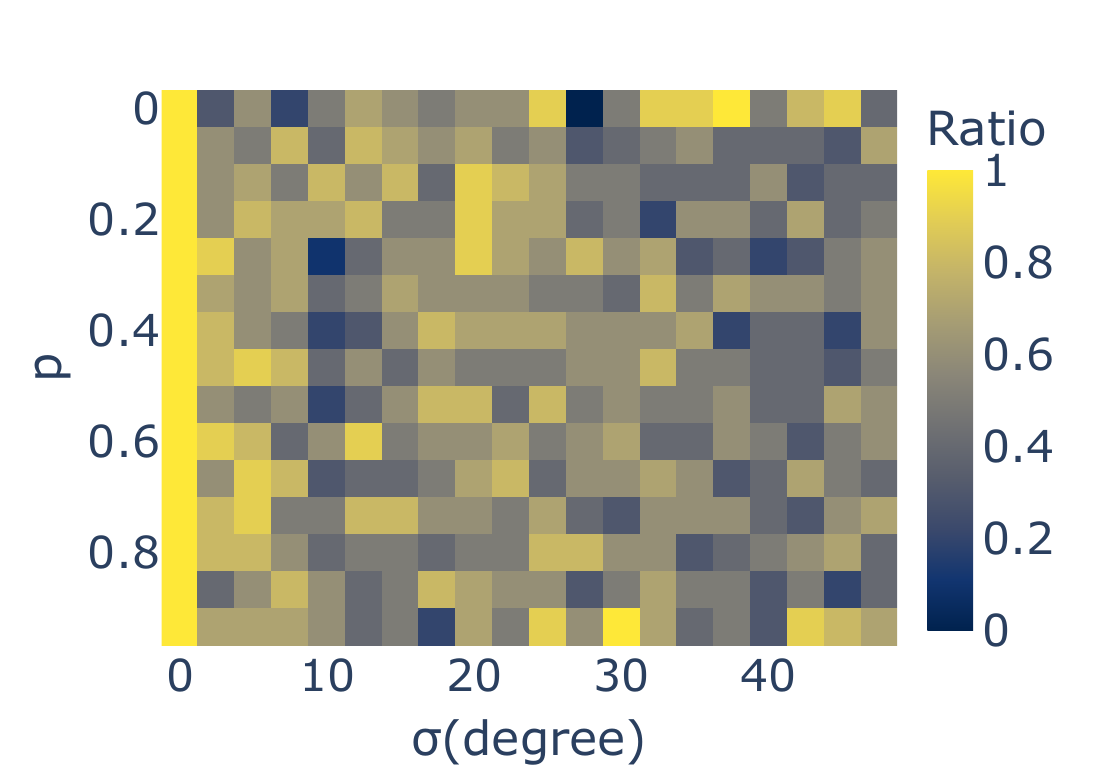} &
  \includegraphics[width=0.4\textwidth]{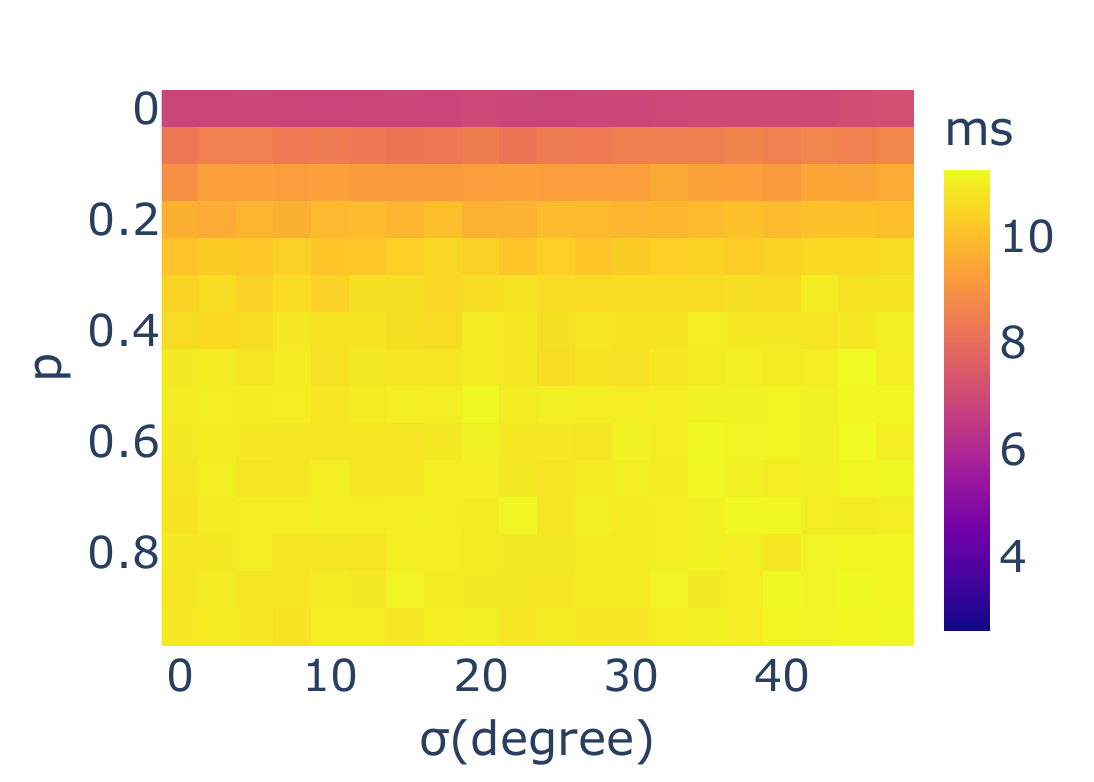} &
  \includegraphics[width=0.4\textwidth]{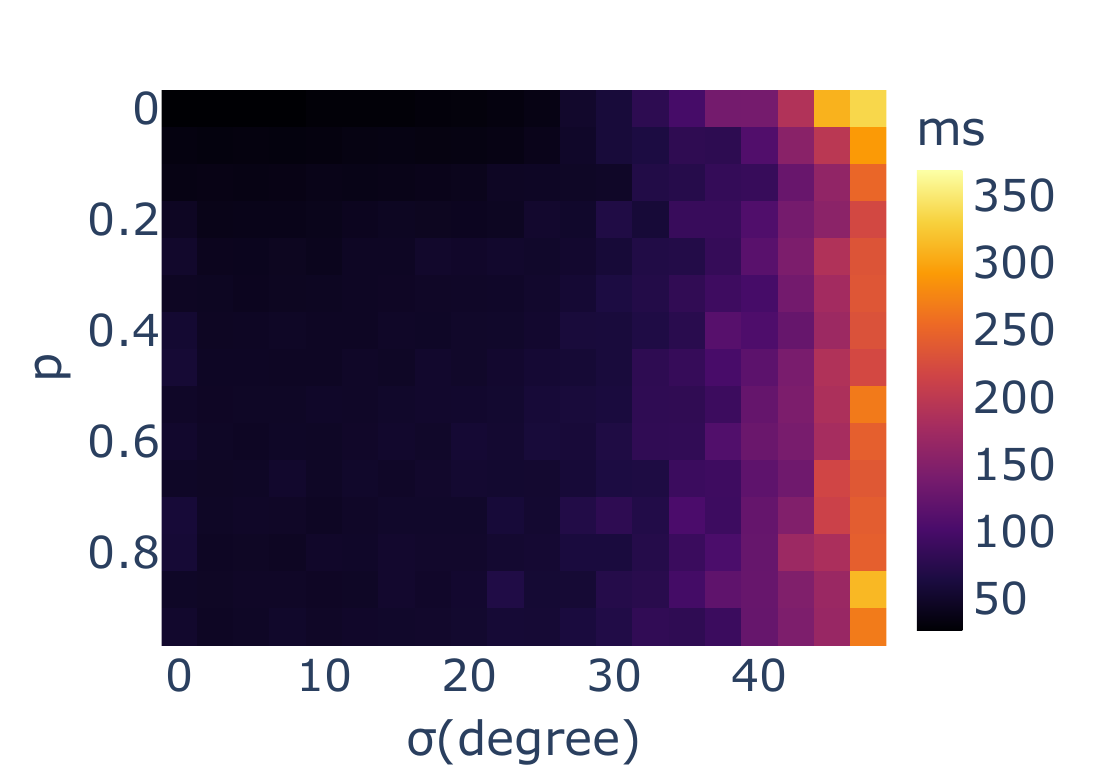} &
  \includegraphics[width=0.4\textwidth]{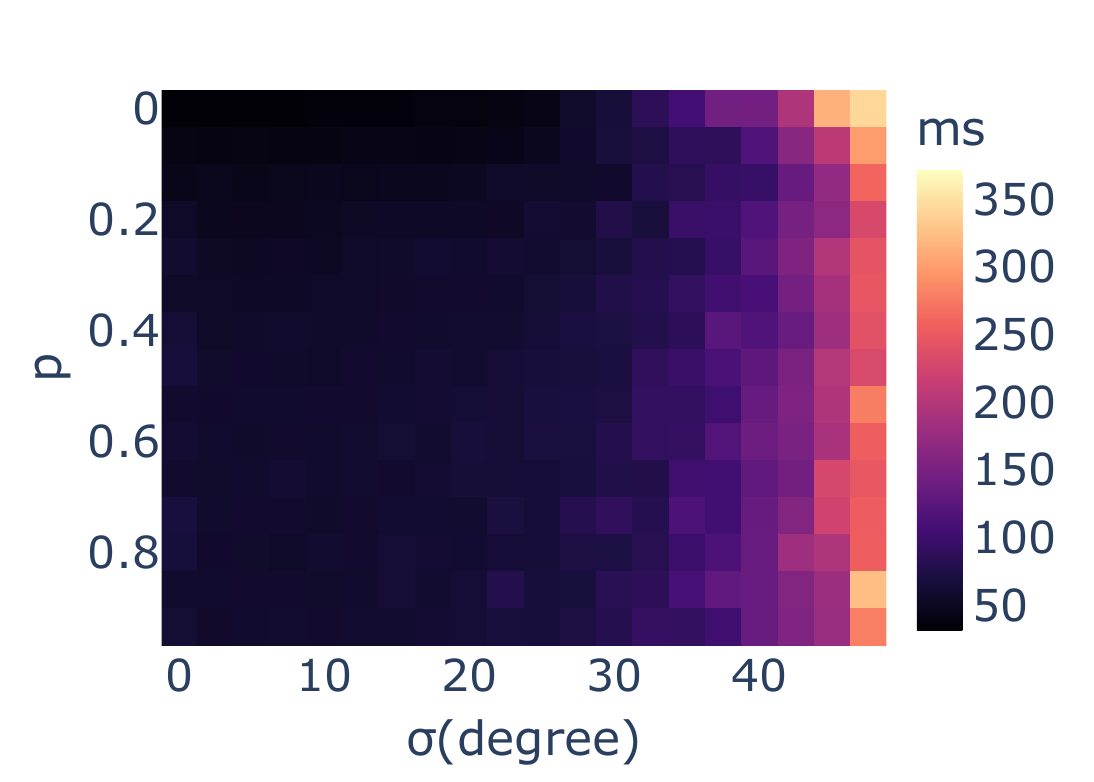} \\
\end{tabular}

  }
  \caption{Heatmap grid analysis of \Sim{3}.}
  \label{fig:heatmap_sim3}
\end{figure*}

To systematically evaluate the robustness of our algorithms, we adopt the \ac{WS}
small-world experimental framework introduced by Wilson~\cite{Wilson20cvpr_Distribution}
in order to generate synthetic \ac{GS} test problems of varying difficulty.
We first construct a measurement network $G$ by
drawing a random sample from the \ac{WS} model, a distribution
over random graphs characterized by three parameters: the number of nodes $N$,
a parameter $k$ that controls the size of a ``local" neighborhood around each node, and
the ``rewiring probability'' $p \in [0, 1]$ that controls the random
generation of long range connections that serve to enhance its \emph{global}
connectivity properties. Below we use $N=100$ and $k=16$.

\begin{figure*}[ht]
  \centering
  \vspace*{4pt}
  \setlength{\tabcolsep}{6pt}
  \renewcommand{\arraystretch}{1.0}
  
  \resizebox{0.90\textwidth}{!}{%
  \begin{tabular}{ccc}
    
    %
  \PanelBox{%
    \centering
    {\footnotesize\bfseries \SE{2} Intel}\quad{\scriptsize $\numX\!=\!1228,\;\numZ\!=\!1483$}\\[2pt]
    \begin{overpic}[width=\linewidth]{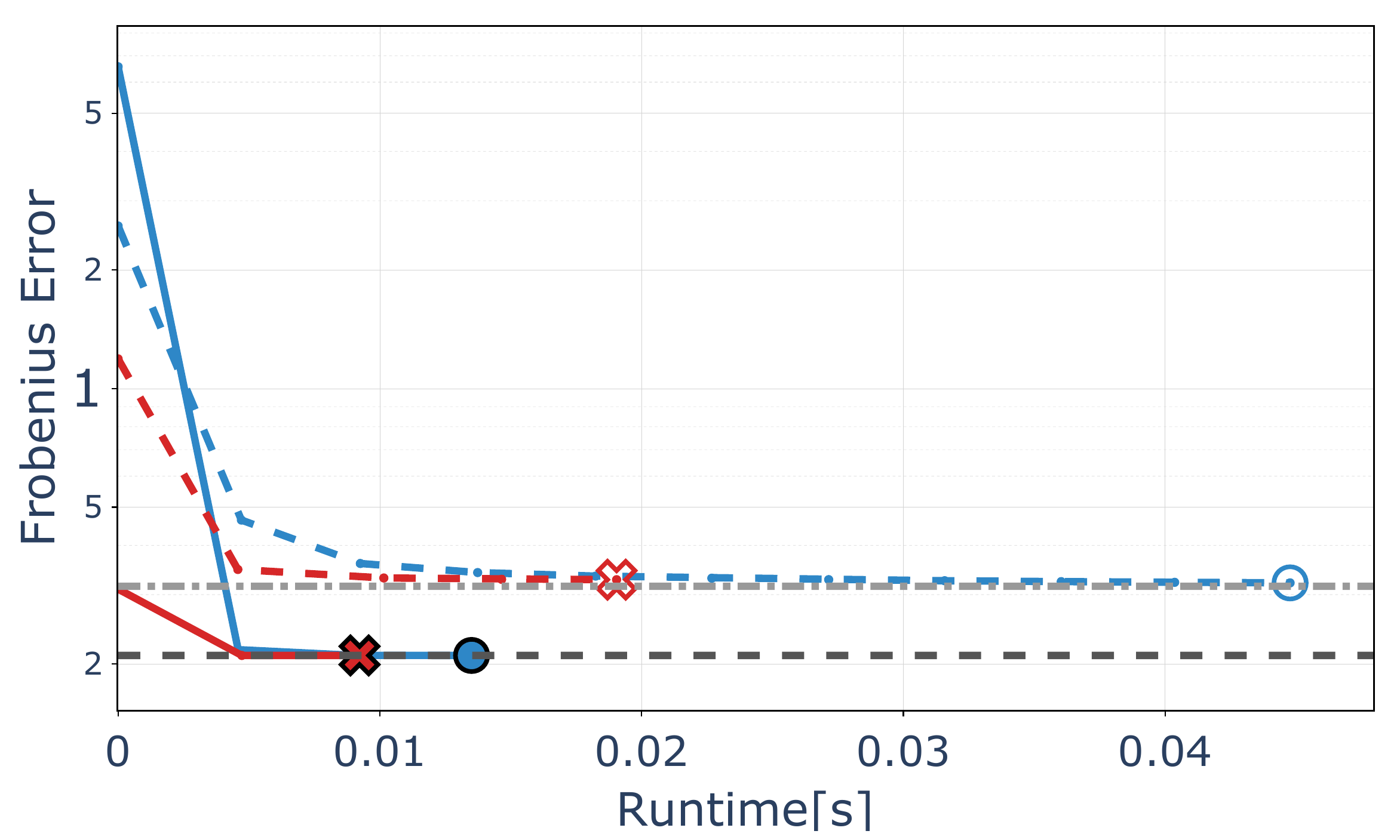}%
      \put(58,47){\InsetTimeTable{0.013}{\textbf{0.009}}{0.044}{\textbf{0.019}}}%
    \end{overpic}%
  }%

    &
  \PanelBox{%
    \centering
    {\footnotesize\bfseries \SE{3} Rim}\quad{\scriptsize $\numX\!=\!10195,\;\numZ\!=\!29743$}\\[2pt]
    \begin{overpic}[width=\linewidth]{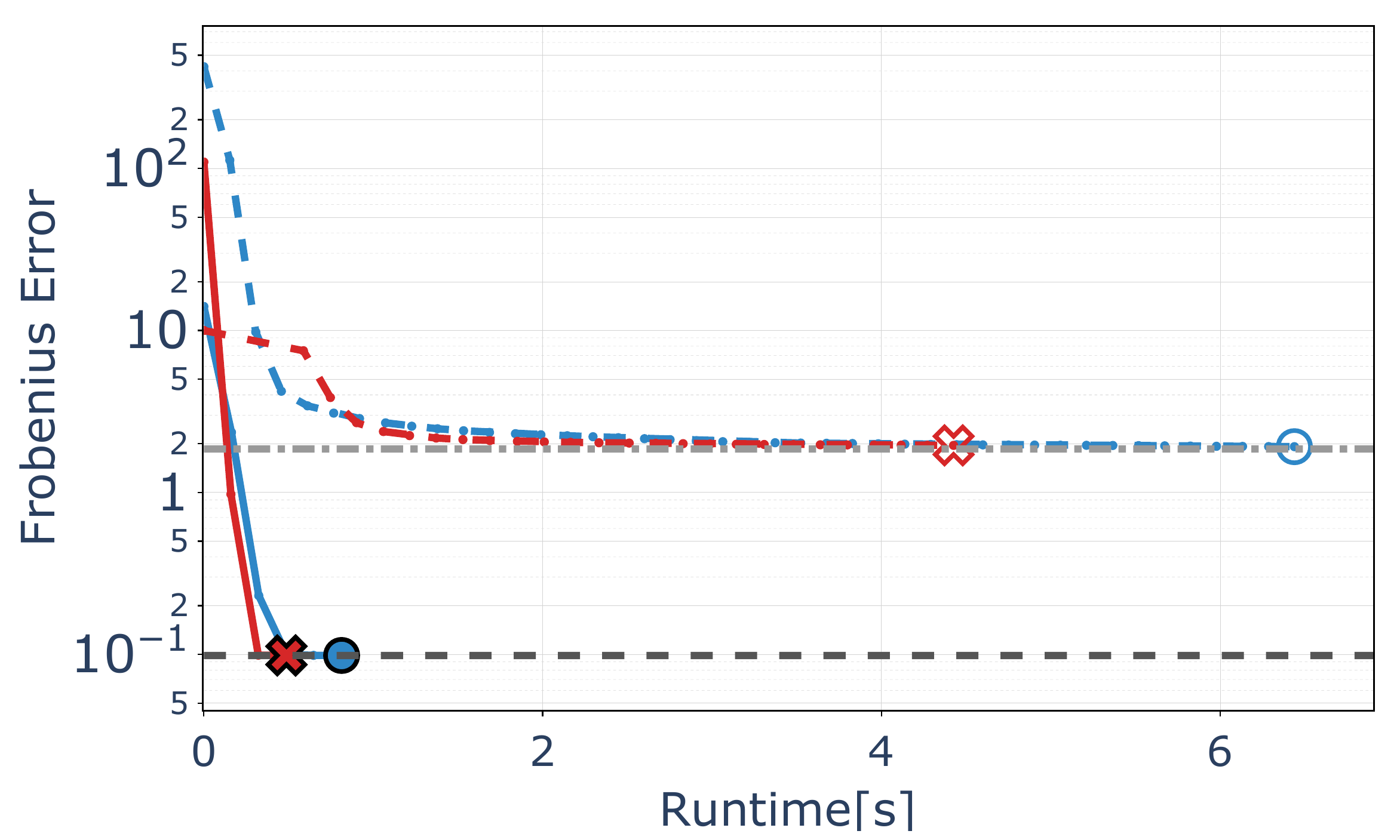}%
      \put(58,47){\InsetTimeTable{0.813}{\textbf{0.487}}{6.437}{\textbf{4.425}}}%
    \end{overpic}%
  }%

    &
  \PanelBox{%
    \centering
    {\footnotesize\bfseries \SO{3} Pantheon Ext.}\quad{\scriptsize $\numX\!=\!718,\;\numZ\!=\!49254$}\\[2pt]
    \begin{overpic}[width=\linewidth]{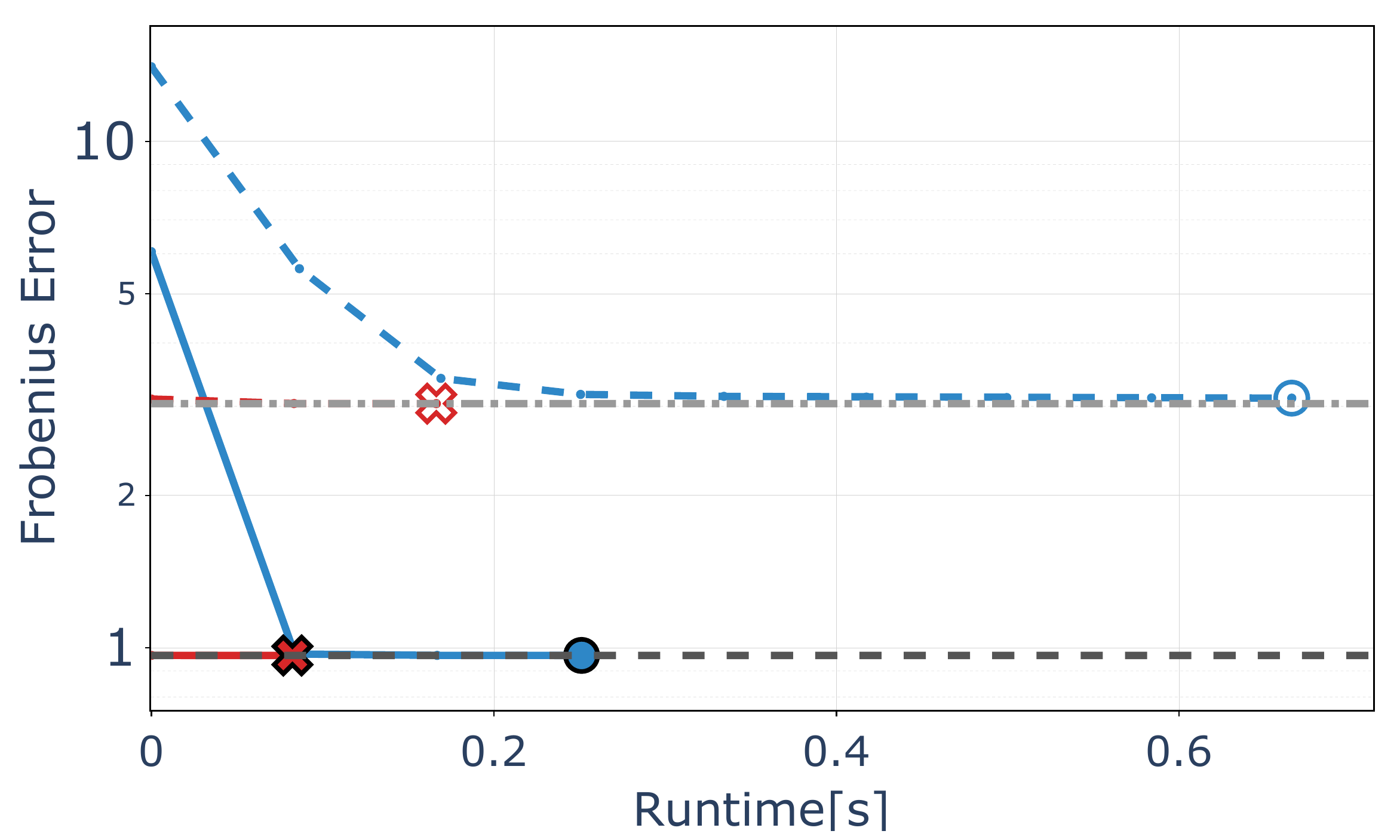}%
      \put(58,47){\InsetTimeTable{0.251}{\textbf{0.082}}{0.665}{\textbf{0.166}}}%
    \end{overpic}%
  }%

    \\ [-4pt]
    
    %
  \PanelBox{%
    \centering
    {\footnotesize\bfseries \Gal{3} Drone}\quad{\scriptsize $\numX\!=\!1000,\;\numZ\!=\!2000$}\\[2pt]
    \begin{overpic}[width=\linewidth]{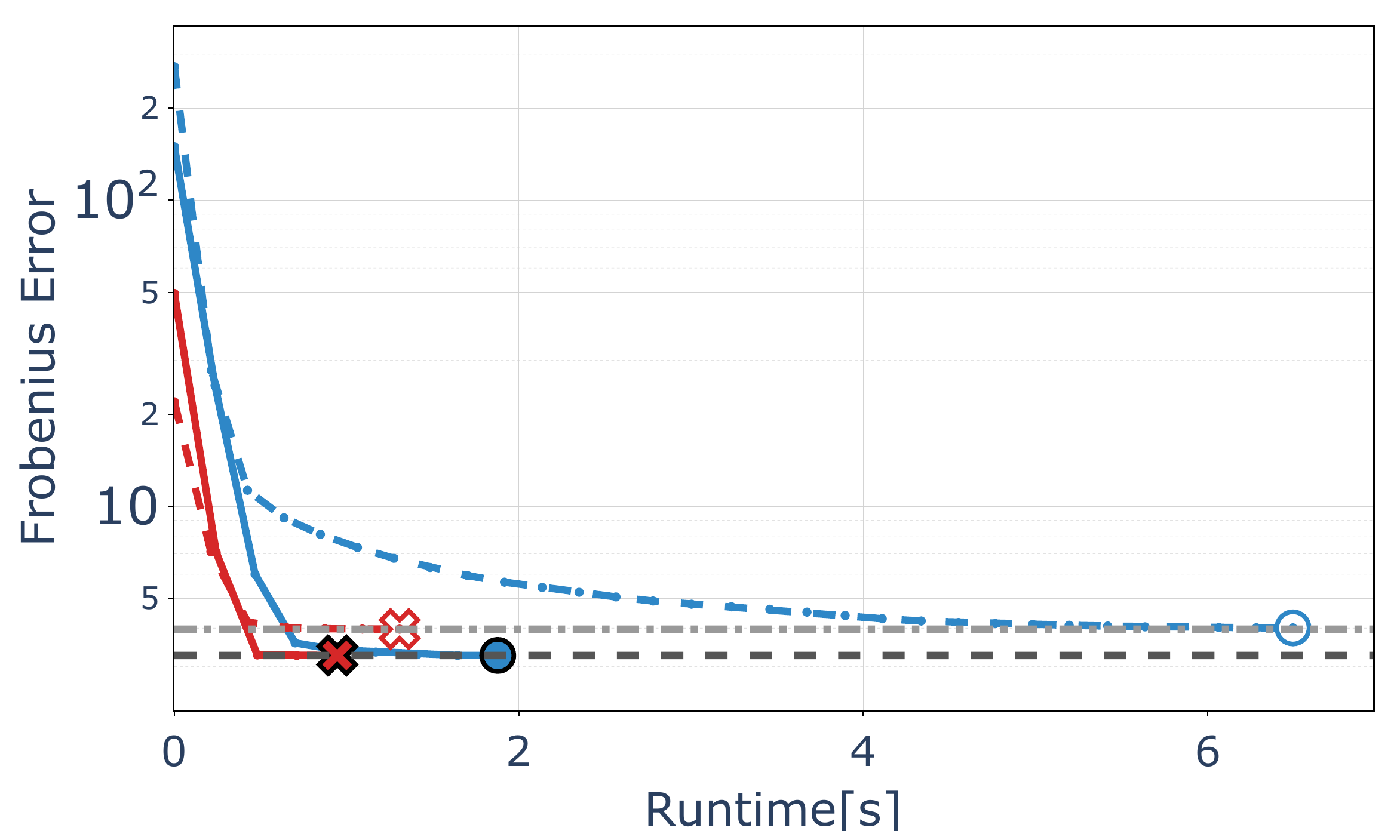}%
      \put(58,47){\InsetTimeTable{1.877}{\textbf{0.945}}{6.493}{\textbf{1.307}}}%
    \end{overpic}%
  }%

    &
    \PanelBox{%
    {\footnotesize\bfseries Dataset configuration difficulty map}\\[4pt]
    \begin{overpic}[width=\linewidth]{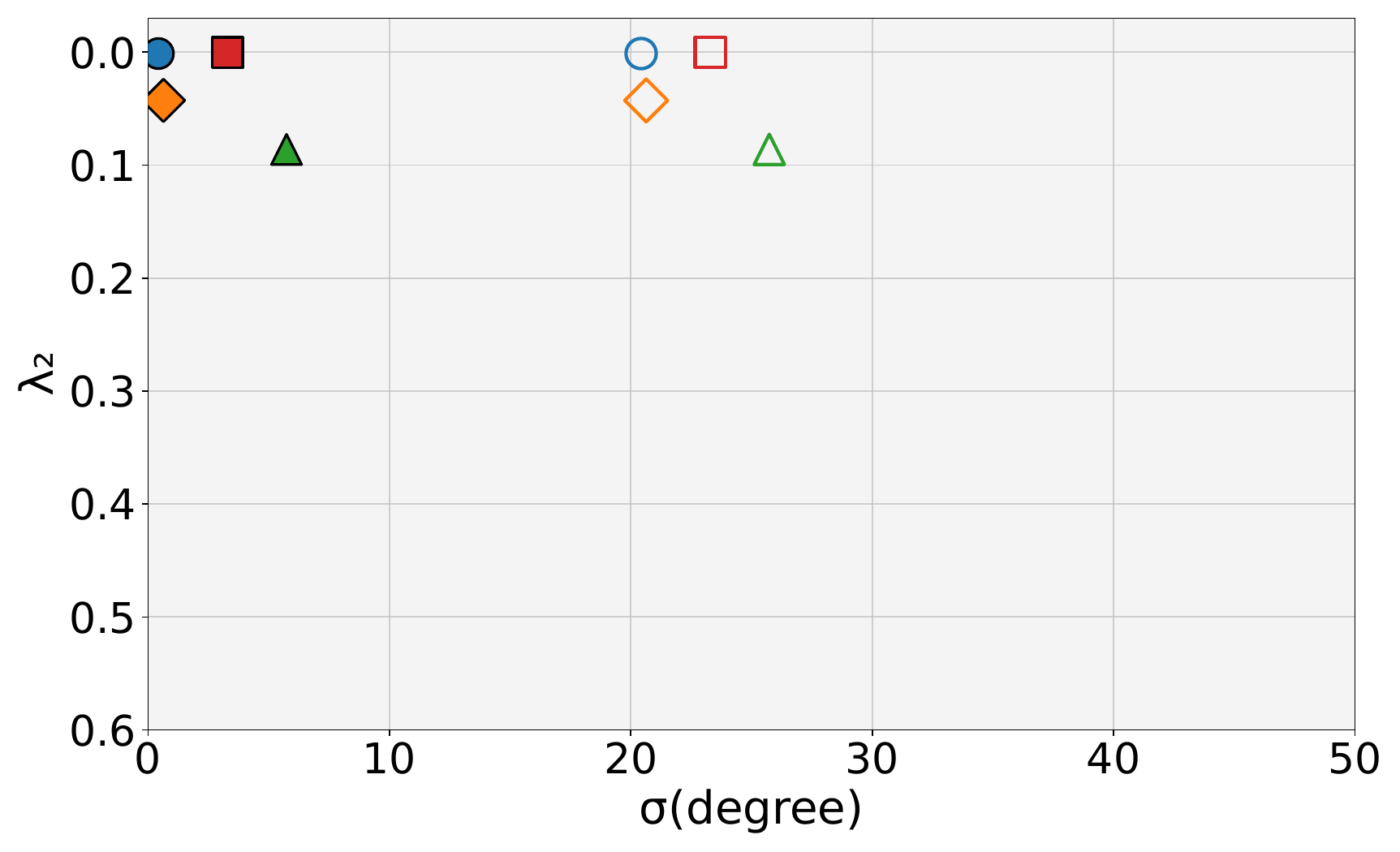}%
      \put(51,9){\includegraphics[width=0.45\linewidth]{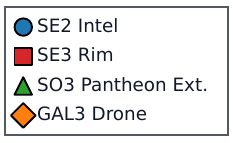}}%
    \end{overpic}%
    }
    \\ [-1pt]
    
    \multicolumn{3}{c}{\vspace{3pt}\includegraphics[width=0.20\textwidth]{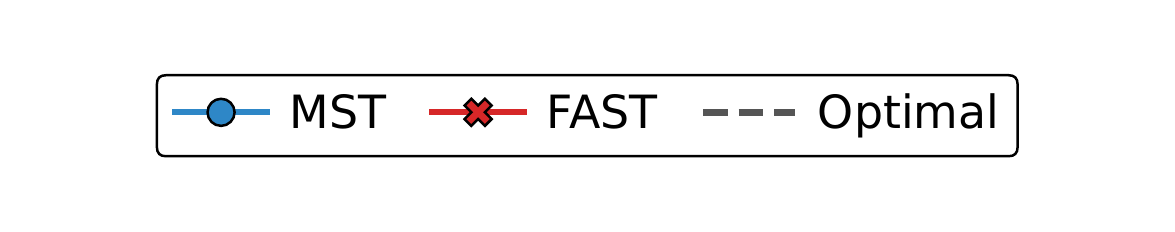}}
    \\
    
  \end{tabular}%
  }
  
  \caption{Cross-group benchmark: speed vs.\ Frobenius accuracy for representative
  datasets across multiple matrix Lie groups.}
  \label{fig:cross_group_frob}
\end{figure*}

Given the measurement graph $G$, we sample realizations of the ground-truth values
$X_i$ and measurements $\Zij$ associated with the vertices and edges of
$G$, covering the following groups: \SO{3}, \SE{3}, \SU{2}, \SL{4}, \Gal{3}, \Sim{3}.
In addition to the rewiring probability $p$, problem difficulty is controlled by the
measurement noise level $\sigma$, with higher $\sigma$ and lower $p$ yielding
more challenging instances. We generate test problems by jointly varying $\sigma$ and $p$
and sampling 10 random \ac{GS} instances for each setting of $(\sigma, p)$.

We compare against an initialization baseline
obtained by chaining relative measurements along the edges of a spanning tree,
preferentially selecting high-confidence measurements by weighting edges by
their inverse noise variance $\kappa_{ij}$; we refer to this as a maximum spanning tree
or \mst{}.
To quantify \fastsync{}'s and \mst{}'s performance \emph{as initialization methods}, we
measure their \emph{local success rate}: the fraction of instances where each method's
{initial} estimate leads to a \emph{globally optimal} \emph{final} estimate after
\emph{local} refinement. We compare final objective values against a high-quality
reference solution: in Figures~\ref{fig:heatmap_so3}--\ref{fig:heatmap_su2}, the
reference is provided by \sesync{}, which computes \emph{certifiably globally optimal}
solutions for \ac{GS} over $\SO{d}$, $\SE{d}$, or $\SU{d}$; in
Figures~\ref{fig:heatmap_sl4}--\ref{fig:heatmap_sim3}, we \emph{estimate} the global
optimum by initializing with ground truth and then performing local optimization. A run is
declared a success if the final objective value satisfies
$|J(X^*) - J(\hat{X})| \leq 10^{-3}(1 + |J(\hat{X})|)$, where $J$ denotes the Frobenius
surrogate objective~\eqref{eq:J_F} for \SO{d}, \SE{d}, and \SU{2}
and the analogous objective formed by summing  \eqref{eq:original_error} over all edges
for \SL{4}, \Gal{3}, and \Sim{3}.

We also directly compare final objective values from \fastsync{} and \mst{} using the
\emph{performance profiles} of Dolan and Moré~\cite{Dolan02mp_performanceProfile}. Given
a set of test instances and methods, the performance profile for each method reports, as a
function of $\tau \ge 1$, the fraction of problems where that method achieves a cost
within a factor of $\tau$ of the \emph{best-performing} method on that instance.
Finally, we quantify computational cost by reporting both (i) median initialization time
and (ii) time for subsequent local optimization using that initialization.

\textbf{Groups using \sesync{}}
From Figures~\ref{fig:heatmap_so3}--\ref{fig:heatmap_su2} we see that \mst{} exhibits
lower initialization cost but requires substantially more local optimization for
high-noise regimes, resulting in total execution times that scale poorly. Conversely,
\fastsync{} incurs upfront initialization costs but dramatically reduces
subsequent local optimization time. The local success rate heatmaps show \mst{} has a
lower ratio under high noise and rewiring probability, while \fastsync{} maintains a
higher ratio of success.
The performance profile comparison further shows that \fastsync{} outperforms \mst{} on
nearly all instances. \fastsync{} consistently provides
initializations within the basin of attraction of the global optimum for \SO{3},
\SE{3} and \SU{2}, enabling efficient convergence to high-quality solutions while remaining faster
than \mst{}.

\textbf{Groups using ground-truth initialized estimate}
Figures~\ref{fig:heatmap_sl4}--\ref{fig:heatmap_sim3} present results for
\SL{4}, \Gal{3}, and \Sim{3}. Across these, \fastsync{} consistently achieves
better or comparable refinement relative to \mst{}. Although \fastsync{} has higher
initialization cost, the local optimizer is closer to the basin of attraction,
reducing subsequent optimization time and yielding a lower total computational cost.

\begin{table}[t]
\caption{Comparison of \fastsync{} with and without \acf{ND},
 \revison{and against off-the-shelf Chordal initialization}}
\label{tab:merged-nd-ablation-chordal}
\centering
\scriptsize
\setlength{\tabcolsep}{4pt}
\begin{tabular}{llccccc}
\toprule
Statistic & Method & SO3 & SE3 & SIM3 & GAL3 & SL4 \\
\midrule
\multirow{3}{*}{Optimality}
  & \revison{Chordal}     & \revison{1.0} & \revison{1.0} & \revison{---} & \revison{---} & \revison{---} \\
  & Fast w/o ND  & 1.0          & 1.0          & 0.8  & 1.0  & 1.0  \\
  & Fast         & 1.0          & 1.0          & \textbf{0.9} & 1.0 & 1.0 \\
\midrule
\multirow{3}{*}{Speed (ms)}
  & \revison{Chordal}     & \revison{46.7} & \revison{79.4} & \revison{---} & \revison{---} & \revison{---} \\
  & Fast w/o ND  & 33.9         & 67.1         & 143  & 187  & 534  \\
  & Fast         & \textbf{28.6}& \textbf{50.0}& \textbf{72.2} & \textbf{113} & \textbf{333} \\
\midrule
\multirow{3}{*}{Iterations}
  & \revison{Chordal}     & \revison{26.5} & \revison{27.4} & \revison{---} & \revison{---} & \revison{---} \\
  & Fast w/o ND  & 26.5         & 29.8         & 30.1 & 19.2 & 18.7 \\
  & Fast         & 28.5         & 30.3         & 28.3 & 19.7 & \textbf{17.6} \\
\midrule
\multirow{3}{*}{DAG height}
  & \revison{Chordal}     & \revison{98.2} & \revison{98.1} & \revison{---} & \revison{---} & \revison{---} \\
  & Fast w/o ND  & 98.2         & 98.1         & 98.2 & 98.2 & 98.1 \\
  & Fast         & \textbf{72.6}& \textbf{72.6}& \textbf{72.5}& \textbf{72.5}& \textbf{72.5}\\
\bottomrule
\end{tabular}
\end{table}

\textbf{Ablation Study}
To isolate the effect of \Acf{ND}, we repeated the experiments
with and without \Ac{ND}. Rather than reproducing the full heatmap grids,
Table~\ref{tab:merged-nd-ablation-chordal} reports scalar summaries obtained by averaging over 10 trials
and then aggregating
over the same $(p,\sigma)$ cells. We report median local success over the high-noise
regime $\sigma \geq 25^\circ$, where initialization quality is most discriminative;
median total runtime over all cells; median LM iteration count over the same high-noise
cells; and the median height of the elimination DAG. Boldface denotes significance under
a paired one-sided sign test with ties discarded and $\alpha=0.05$.

Table~\ref{tab:merged-nd-ablation-chordal} shows that \Ac{ND} preserves or slightly improves solution
quality while substantially reducing runtime by producing shallower elimination
structures, with little effect on local optimizer convergence.

\textbf{Comparison With Chordal}
Finally, we compare \fastsync{} with the original Chordal method on \SO{3} and \SE{3}, as defined in~\cite{Martinec07cvpr_robust,Carlone15icra_initialization}. We observe in Table~\ref{tab:merged-nd-ablation-chordal} that \fastsync{} is approximately
$35\%$--$40\%$ faster than Chordal while achieving nearly the same solution quality. This
supports the interpretation of \fastsync{} as a generalized and computationally improved
extension of Chordal initialization.


\section{RESULTS ON REAL DATASETS}

Real measurement graphs often exhibit irregular structure, varying edge densities, and
noise characteristics that are difficult
to simulate faithfully. In this section, we evaluate how \fastsync{} and \mst{} compare
on real-world datasets. Since we do not have access to ground truth for these examples,
we measure (i) initial error, (ii) final refinement error, and (iii) computation time.

We evaluate on four datasets spanning different Lie groups, problem sizes, and graph
connectivity: \SE{2} Intel and \SE{3} Rim are
from~\cite{Carlone14tro_FromAngularManifolds}. \SO{3} Pantheon Ext is
from~\cite{Heinly15cvpr_Yahoo100M}. The \Gal{3} Drone dataset is \textit{derived} from a
drone sequence described in~\cite{Burri16ijrr_EuRoC}, but with synthetic measurements.
The number of nodes and relative measurements for each dataset is labeled with $N$ and
$M$ respectively.

For each dataset, we track the Frobenius error of both \fastsync{} and \mst{} as a
function of runtime, from initialization through local optimization. To test robustness,
we also evaluate on a corrupted variant of each dataset in which additional noise is
injected into the existing edge measurements. Each configuration was run 10 times on the
same dataset to obtain median runtime estimates. In each subplot, solid lines denote
performance on the original dataset and dashed lines denote performance on the corrupted
variant. Two optimal baselines are shown: the black dotted line indicates the optimal
error for the original dataset, and the gray dotted line indicates the optimal error
for the corrupted variant. The optimal errors for \SE{2}, \SE{3}, and \SO{3} were found
using \sesync{}, while for \Gal{3} Drone we approximate its
optimal error using its ground truth-initialized estimate.

The bottom right plot identifies each dataset within a difficulty landscape, plotting
algebraic connectivity ($\lambda_2$) against estimated noise level ($\sigma$). Filled
markers represent the original datasets and unfilled markers represent their
noise-corrupted variants. This provides context for interpreting the convergence plots:
datasets in the lower right corner (low connectivity, high noise) represent the most
challenging problems, while those in the upper left (high connectivity, low noise) are
the easiest.

Across all four datasets, \fastsync{} consistently achieves a lower initial Frobenius
error than \mst{} and reaches the optimal solution faster in most cases. This holds for
both the original datasets and their noise-corrupted variants, indicating that
\fastsync{}'s initialization advantage does not depend on favorable measurement
conditions. These results reinforce our synthetic findings: \fastsync{}'s estimates are
closer to the global optimum at the outset, reducing the burden on local optimization.
This advantage persists across diverse Lie groups, problem scales, and connectivity
structures, and remains robust under degraded measurements, demonstrating that
\fastsync{} is a reliable and effective initialization method for practical group
synchronization problems.

\section{CONCLUSIONS}

We present \fastsync{}, a simple, group-agnostic initialization method for
synchronization problems with applications in large estimation tasks in robotics and
vision.
We show that a simple Frobenius surrogate, paired with careful linear-algebraic
design—Kronecker reduction of the measurement matrix, gauge fixing with a
block-preserving \ac{ND} ordering, stable factorization on the reduced system, and a
structured back-substitution—yields high-quality initializations across matrix Lie
groups.
An open direction is to provide a theoretical foundation for the observed effectiveness
of ND-based gauge selection: while our results demonstrate its strong practical value,
establishing when and why it succeeds remains an important challenge for future work.

\addtolength{\textheight}{-0.6cm}

\bibliographystyle{IEEEtran}
\bibliography{references}

\begin{thebibliography}{10}
\providecommand{\url}[1]{#1}
\csname url@samestyle\endcsname
\providecommand{\newblock}{\relax}
\providecommand{\bibinfo}[2]{#2}
\providecommand{\BIBentrySTDinterwordspacing}{\spaceskip=0pt\relax}
\providecommand{\BIBentryALTinterwordstretchfactor}{4}
\providecommand{\BIBentryALTinterwordspacing}{\spaceskip=\fontdimen2\font plus
\BIBentryALTinterwordstretchfactor\fontdimen3\font minus
  \fontdimen4\font\relax}
\providecommand{\BIBforeignlanguage}[2]{{%
\expandafter\ifx\csname l@#1\endcsname\relax
\typeout{** WARNING: IEEEtran.bst: No hyphenation pattern has been}%
\typeout{** loaded for the language `#1'. Using the pattern for}%
\typeout{** the default language instead.}%
\else
\language=\csname l@#1\endcsname
\fi
#2}}
\providecommand{\BIBdecl}{\relax}
\BIBdecl

\bibitem{Govindu01cvpr_MotionEstimation}
V.~Govindu, ``{Combining two-view constraints for motion estimation},'' in
  \emph{Proceedings of the 2001 IEEE Computer Society Conference on Computer
  Vision and Pattern Recognition. CVPR 2001}, vol.~2, 2001.

\bibitem{Martinec07cvpr_robust}
D.~Martinec and T.~Pajdla, ``{Robust rotation and translation estimation in
  multiview reconstruction},'' in \emph{2007 IEEE Conference on Computer Vision
  and Pattern Recognition}.\hskip 1em plus 0.5em minus 0.4em\relax IEEE, 2007,
  pp. 1--8.

\bibitem{Lu97ar_GloballyConsistent}
F.~Lu and E.~Milios, ``{Globally Consistent Range Scan Alignment for
  Environment Mapping},'' \emph{Autonomous Robots}, vol.~4, no.~4, 1997.

\bibitem{Govindu04cvpr_LieAlgebraic}
V.~Govindu, ``{Lie-algebraic averaging for globally consistent motion
  estimation},'' in \emph{Proceedings of the 2004 IEEE Computer Society
  Conference on Computer Vision and Pattern Recognition.}, vol.~1, 2004.

\bibitem{Rosen19ijrr_SE_Sync}
D.~M. Rosen, L.~Carlone, A.~S. Bandeira, and J.~J. Leonard, ``{SE-Sync: A
  certifiably correct algorithm for synchronization over the special Euclidean
  group},'' \emph{The International Journal of Robotics Research}, vol.~38, no.
  2-3, pp. 95--125, 2019.

\bibitem{Dellaert20eccv_Shonan}
F.~Dellaert, D.~M. Rosen, J.~Wu, R.~Mahony, and L.~Carlone, ``{Shonan Rotation
  Averaging: Global Optimality by Surfing $SO(p)^n$},'' in \emph{ECCV 2020},
  2020, pp. 292--308.

\bibitem{Arrigoni20ijcv_Synchronization}
F.~Arrigoni and A.~Fusiello, ``{Synchronization Problems in Computer Vision
  with Closed-Form Solutions},'' \emph{International Journal of Computer
  Vision}, vol. 128, no.~1, pp. 26--52, 2020.

\bibitem{Strasdat10rss_ScaleDrift}
H.~Strasdat, J.~Montiel, and A.~J. Davison, ``{Scale drift-aware large scale
  monocular SLAM},'' \emph{Robotics: science and Systems VI}, vol.~2, no.~3,
  p.~7, 2010.

\bibitem{Maggio25arxiv_VGGT_SLAM}
D.~Maggio, H.~Lim, and L.~Carlone, ``{VGGT-SLAM: Dense RGB SLAM Optimized on
  the SL (4) Manifold},'' \emph{arXiv preprint arXiv:2505.12549}, 2025.

\bibitem{Barrau16tac_InvariantEKF}
A.~Barrau and S.~Bonnabel, ``{The invariant extended Kalman filter as a stable
  observer},'' \emph{IEEE Transactions on Automatic Control}, vol.~62, no.~4,
  pp. 1797--1812, 2016.

\bibitem{Kelly23arxiv_Galilean}
J.~Kelly, ``{Making Space for Time: The Special Galilean Group and Its
  Application to Some Robotics Problems},'' \emph{arXiv preprint
  arXiv:2409.14276}, 2024.

\bibitem{Wilson14eccv_1dsfm}
K.~Wilson and N.~Snavely, ``{Robust global translations with 1dsfm},'' in
  \emph{European conference on computer vision}.\hskip 1em plus 0.5em minus
  0.4em\relax Springer, 2014, pp. 61--75.

\bibitem{Singer11siam_Cryo}
A.~Singer and Y.~Shkolnisky, ``Three-dimensional structure determination from
  common lines in {cryo-EM} by eigenvectors and semidefinite programming,''
  \emph{SIAM journal on imaging sciences}, vol.~4, no.~2, pp. 543--572, 2011.

\bibitem{Wilson20cvpr_Distribution}
K.~Wilson and D.~Bindel, ``{On the distribution of minima in intrinsic-metric
  rotation averaging},'' in \emph{Proceedings of the IEEE/CVF Conference on
  Computer Vision and Pattern Recognition}, 2020.

\bibitem{Singer11acha_AngularSync}
A.~Singer, ``{Angular synchronization by eigenvectors and semidefinite
  programming},'' \emph{Applied and Computational Harmonic Analysis}, vol.~30,
  no.~1, pp. 20--36, 2011.

\bibitem{Hartley13ijcv_RA}
R.~Hartley, J.~Trumpf, Y.~Dai, and H.~Li, ``{Rotation Averaging},''
  \emph{International Journal of Computer Vision}, vol. 103, no.~3, pp.
  267--305, 2013.

\bibitem{Goemans95jacm_SDP}
M.~X. Goemans and D.~P. Williamson, ``{Improved approximation algorithms for
  maximum cut and satisfiability problems using semidefinite programming},''
  \emph{Journal of the ACM (JACM)}, vol.~42, no.~6, 1995.

\bibitem{Bandeira17mp_MaxLikelihood}
A.~S. Bandeira, N.~Boumal, and A.~Singer, ``{Tightness of the maximum
  likelihood semidefinite relaxation for angular synchronization},''
  \emph{Mathematical Programming}, vol. 163, no.~1, pp. 145--167, 2017.

\bibitem{Boumal14jmlr_Manopt}
N.~Boumal, B.~Mishra, P.-A. Absil, and R.~Sepulchre, ``{Manopt, a Matlab
  Toolbox for Optimization on Manifolds},'' \emph{Journal of Machine Learning
  Research}, vol.~15, no.~42, pp. 1455--1459, 2014.

\bibitem{Carlone15icra_initialization}
L.~Carlone, R.~Tron, K.~Daniilidis, and F.~Dellaert, ``{Initialization
  Techniques for 3D {SLAM}: A Survey on Rotation Estimation and Its use in Pose
  Graph Optimization},'' in \emph{{IEEE} International Conference on Robotics
  and Automation}.\hskip 1em plus 0.5em minus 0.4em\relax IEEE, 2015, pp.
  4597--4604.

\bibitem{Dellaert12techreport_GTSAM}
F.~Dellaert, ``{GTSAM: A Nonlinear Optimization Library for Robotics and
  Vision},'' Georgia Institute of Technology, Technical Report
  GT-RIM-CP\&R-2012-002, 2012.

\bibitem{Yannakakis81siam_NPComplete}
M.~Yannakakis, ``{Computing the Minimum Fill-In is {NP}-Complete},'' \emph{SIAM
  Journal on Algebraic and Discrete Methods}, vol.~2, no.~1, pp. 77--79, 1981.

\bibitem{Davis08toms_CHOLMOD}
Y.~Chen, T.~Davis, W.~Hager, and S.~Rajamanickam, ``{Algorithm 887: CHOLMOD,
  Supernodal Sparse {C}holesky Factorization and Update/Downdate},'' \emph{ACM
  Trans. Math. Softw.}, vol.~35, no.~3, pp. 22:1--22:14, Oct 2008.

\bibitem{George73siam_NestedDissection}
A.~George, ``{Nested Dissection of a Regular Finite Element Mesh},'' \emph{SIAM
  Journal on Numerical Analysis}, vol.~10, no.~2, pp. 345--363, April 1973.

\bibitem{Lipton79siam_GeneralizedNestedDissection}
R.~Lipton, D.~Rose, and R.~Tarjan, ``{Generalized Nested Dissection},''
  \emph{SIAM Journal on Applied Mathematics}, vol.~16, no.~2, pp. 346--358,
  1979.

\bibitem{Dolan02mp_performanceProfile}
E.~D. Dolan and J.~J. Mor{\'e}, ``Benchmarking optimization software with
  performance profiles,'' \emph{Mathematical programming}, vol.~91, no.~2, pp.
  201--213, 2002.

\bibitem{Carlone14tro_FromAngularManifolds}
L.~Carlone and A.~Censi, ``{From Angular Manifolds to the Integer Lattice:
  Guaranteed Orientation Estimation With Application to Pose Graph
  Optimization},'' \emph{IEEE Trans. Robotics}, vol.~30, no.~2, 2014.

\bibitem{Heinly15cvpr_Yahoo100M}
J.~Heinly, J.~L. Sch{\"o}nberger, E.~Dunn, and J.-M. Frahm, ``Reconstructing
  the world in six days (as captured by the {Yahoo} 100 million image
  dataset),'' in \emph{Proceedings of the IEEE Conference on Computer Vision
  and Pattern Recognition (CVPR)}, 2015.

\bibitem{Burri16ijrr_EuRoC}
M.~Burri, J.~Nikolic, P.~Gohl, T.~Schneider, J.~Rehder, S.~Omari, M.~W.
  Achtelik, and R.~Siegwart, ``{The EuRoC micro aerial vehicle datasets},''
  \emph{The International Journal of Robotics Research}, 2016.

\end{thebibliography}

\end{document}